%% file: main_tech_report.tex
\documentclass[oneside,a4paper,onecolumn,11pt]{article}

\usepackage{fullpage}
\usepackage{fancyhdr,fancyheadings,nopageno,lastpage} % setting header and footer
\usepackage[left=2cm,top=2cm,bottom=2cm,right=2cm,includehead,nomarginpar,headheight=16pt]{geometry}
\usepackage[ruled, linesnumbered, vlined, commentsnumbered]{algorithm2e}
\usepackage{graphicx,subfig} % figure related
\usepackage{amsfonts,amssymb,amsmath,amsthm,amsopn,mathrsfs,mathtools}	% math related
\usepackage{multirow,diagbox,booktabs,threeparttable}
\usepackage{color,xcolor} % Required for custom colors
\usepackage{ifsym}

\usepackage{textcomp}

\usepackage{ragged2e}
\usepackage{bbding}

\usepackage{enumerate}
\usepackage[shortlabels]{enumitem}
\usepackage{xcolor,colortbl}
\usepackage{pifont}
\usepackage{afterpage}

\usepackage{csquotes}
\usepackage{authblk}
\usepackage{footnote}
\usepackage{hyperref}
\usepackage{prettyref}
\usepackage{cite}
\usepackage{setspace}
\usepackage{color}
\usepackage{xcolor}  % Required for custom colors

\usepackage{geometry}
\usepackage{academicons}
\usepackage{fontawesome5}
\usepackage{arydshln}
\usepackage{tikz}
\usepackage{dashrule}
\usepackage{fontawesome5}

\usepackage{array}
\newcolumntype{C}[1]{>{\centering\arraybackslash}p{#1}}

\fancypagestyle{plain}{%
    \fancyfoot[C]{}
    \fancyfoot[R]{\faLeanpub\ \, \thepage\ / \pageref{LastPage}}
    \fancyfoot[L]{\faBraille\ \textsc{COLALab Report} \faSlackHash\ \footnotesize\textifsym{2024015}}
}

\hypersetup{
    colorlinks=true,
    linkcolor=ultramarine,
    filecolor=magenta,
    urlcolor=ultramarine,
    pdftitle={Overleaf Example},
    pdfpagemode=FullScreen,
}

\definecolor{red(munsell)}{rgb}{0.95, 0.0, 0.24}
\definecolor{navyblue}{RGB}{0, 0, 128}
\definecolor{myblue}{RGB}{34,31,217}
\definecolor{mycyan}{gray}{.7}
\definecolor{Gray}{gray}{0.9}
\definecolor{usccardinal}{rgb}{0.6, 0.0, 0.0}
\definecolor{ultramarine}{RGB}{0,32,96}
\definecolor{amber}{rgb}{1.0, 0.49, 0.0}

\newtheorem{definition}{Definition}

\def\our{\texttt{\textsc{MBL-CPDP}}}

\SetCommentSty{mycommfont}

\DeclareMathOperator*{\argmin}{argmin}

\newcommand{\pref}{\prettyref}
\newcommand{\bb}[1]{\multicolumn{1}{>{\columncolor{mycyan}}c}{\textbf{{#1}}}}

\newrefformat{fig}{Fig.~\ref{#1}}
\newrefformat{tab}{Table~\ref{#1}}
\newrefformat{sec}{Section~\ref{#1}}
\newrefformat{app}{Appendix~\ref{#1}}
\newrefformat{alg}{Algorithm~\ref{#1}}
\newrefformat{property}{Property~\ref{#1}}
\newrefformat{theorem}{Theorem~\ref{#1}}
\newrefformat{corollary}{Corollary~\ref{#1}}
\newrefformat{proposition}{Proposition~\ref{#1}}
\newrefformat{def}{Definition~\ref{#1}}
\newrefformat{eq}{equation~(\ref{#1})}

\newenvironment{code-example}
{
\vspace{0.15cm}
\noindent\begin{minipage}{\linewidth}
\begin{center}
\arrayrulecolor{black}
\color{black}
\begin{tabular}{|p{0.95\linewidth}|}
\hline%
\rowcolor{pink!20}%
}
{
\\\hline
\end{tabular}
\end{center}
\end{minipage}
\vspace{-0.2cm}
}

\begin{document}

%% title
\title{\vspace{-1ex}\LARGE\textbf{MBL-CPDP: A Multi-objective Bilevel Method for
Cross-Project Defect Prediction via Automated
Machine Learning}}

%% authors and affiliations
\author[1]{\normalsize Jiaxin Chen}
\author[1]{\normalsize Jinliang Ding}
\author[2]{\normalsize Kay Chen Tan}
\author[3]{\normalsize Jiancheng Qian}
\author[3]{\normalsize Ke Li}
\affil[1]{\normalsize State Key Laboratory of Sythetical Automation for Process Industries, Northeastern University of China, Shenyang 110819, China}
\affil[2]{\normalsize Department of Data Science and Artificial Intelligence, The Hong Kong Polytechnic University, Hong Kong SAR}
\affil[3]{\normalsize Department of Computer Science, University of Exeter, EX4 4RN, Exeter, UK}
%\affil[\Faxmachine\ ]{\normalsize \texttt{frank@hit.edu.cn}}

\date{}
\maketitle

\vspace{-3ex}
{\normalsize\textbf{Abstract:}}
Cross-project defect prediction (CPDP) leverages machine learning (ML) techniques to proactively identify software defects, especially where project-specific data is scarce. However, developing a robust ML pipeline with optimal hyperparameters that effectively use cross-project information and yield satisfactory performance remains challenging. In this paper, we resolve this bottleneck by formulating CPDP as a multi-objective bilevel optimization (MBLO) method, dubbed \our. It comprises two nested problems: the upper-level, a multi-objective combinatorial optimization problem, enhances robustness and efficiency in optimizing ML pipelines, while the lower-level problem is an expensive optimization problem that focuses on tuning their optimal hyperparameters. Due to the high-dimensional search space characterized by feature redundancy and inconsistent data distributions, the upper-level problem combines feature selection, transfer learning, and classification to leverage limited and heterogeneous historical data. Meanwhile, an ensemble learning method is proposed to capture differences in cross-project distribution and generalize across diverse datasets. Finally, a MBLO algorithm is presented to solve this problem while achieving high adaptability effectively.  
To evaluate the performance of \our, we compare it with five automated ML tools and $50$ CPDP techniques across $20$ projects. Extensive empirical results show that \our~outperforms the comparison methods, demonstrating its superior adaptability and comprehensive performance evaluation capability.


\input{introduction}

\input{preliminaries}

\input{approach}

\input{settings}

\input{results}

\input{related}

\input{threats}

\input{conclusion}

\section*{Acknowledgment}
K. Li was supported in part by the UKRI Future Leaders Fellowship under Grant MR/S017062/1 and MR/X011135/1; in part by NSFC under Grant 62376056 and 62076056; in part by the Royal Society under Grant IES/R2/212077; in part by the EPSRC under Grant 2404317; in part by the Kan Tong Po Fellowship (KTP\textbackslash R1\textbackslash 231017); and in part by the Amazon Research Award and Alan Turing Fellowship.

\bibliographystyle{IEEEtran}
\bibliography{IEEEabrv,mybib}

\end{document}

%% file: introduction.tex
%!TeX root=main.tex

\section{Introduction}
\label{sec:introduction}

Software defects (SDs) refer to any faults, flaws, or issues in code that can lead to buggy behavior, performance degradation, or failure to meet expected software requirements. Even minor defects can lead to significant financial loss and security breaches loss in some critical applications. For example, if it were measured as a country, then software vulnerabilities and accumulated deficiencies caused cybercrime—which is predicted to inflict damages totaling $\$ 9.5$ trillion USD globally in 2024, according to Cybersecurity Ventures—would be the world’s third-largest economy after the U.S. and China~\cite{biedron2024}.

SD prediction (SDP) is the process of developing predictive models that help in the early identification of SDs based on software metrics and defect data~\cite{QiuLJ18}. It enables
%Software defect prediction (SDP) is one of the most active research areas to detect SDs in software engineering (SE) and plays an important role in software quality assurance. It aims to identify potential SDs by analyzing historical data and characteristics of software projects, enabling 
proactive measures to enhance software quality and reduce testing costs. Early studies concentrated on within-project defect prediction, relying heavily on historical data from the project itself, which can be ineffective for new projects or those with scarce data~\cite{HosseiniTG19}. Cross-project defect prediction (CPDP) has emerged as an approach to address this problem, leveraging historical data from source projects to predict defects in the target project~\cite{HerboldTG18}. Its advantage lies in its reduced dependency on internal historical data, making it applicable in scenarios with limited testing history, especially for newly launched projects.

The common CPDP process involves the use of various machine learning (ML) techniques and methodologies~\cite{TongZLXLLW24}. The performance of most ML techniques depends on the setting of their hyperparameter, such as the learning rate and regularization strength in neural networks. In the past decade, considerable efforts have been invested in developing hyperparameter optimization (HPO) techniques to automate the laborious task of hyperparameter tuning of ML techniques. Some of these techniques have significantly improved the performance of CPDP by exploring the impact of HPO on single classifiers and/or transfer learners~\cite{QuCZJ18,Ozturk191,KwonRB23,WanZQLQ24,OmondiagbeLM24,ChenL21,WangL24,ChenL23,LiC23,XuLL24,LiCY24,ChenLTL22,FanLT20}. These studies, often considering only a few parameters, adopt single-level structures and adjust hyperparameters within a fixed model framework, which can result in insufficient exploration of the parameter space and potentially overlook better model configurations~\cite{LiX0WT20}. Therefore, employing an ingenious structure becomes crucial to effectively handling the complex parameter space and enhancing the performance of CPDP models. 

HPO can essentially boil down to two nested search problems (a.k.a. bilevel optimization problem (BLOP)): at the upper level, we try to seek a good ML technique, while at the lower level, we focus on identifying an optimal configuration for the technique~\cite{FranceschiFSGP18}. Adhering to this idea, Li et al.~\cite{LiXCT20} introduced \texttt{BiLO-CPDP}, a tool that first formulates CPDP tasks as a single-objective BLOP, where the upper-level problem searches for the optimal combination of transfer learning techniques and classifiers, and the lower level selects the hyperparameter for these models. This architecture demonstrated overwhelming superiority over its single-level optimization variant in all cases. Although \texttt{BiLO-CPDP} demonstrates excellent performance, it is formulated as a single-objective problem and overlooks the characteristics of the feature space and model capacity. 

Single-objective methods in CPDP often struggle with adapting to various application scenarios, fail to utilize data information, and offer limited diversity~\cite{LiSY18,NiCWSG19,CanforaLPOPP15}. For example, Canfora et al.~\cite{CanforaLPOPP13,CanforaLPOPP15} proposed a multi-objective CPDP strategy that optimizes detection effectiveness and minimizes lines of code costs, demonstrating greater cost-effectiveness compared to single-objective methods. In~\cite{RyuB16}, they introduce novel multi-objective learning techniques for CPDP that effectively handle class imbalance by employing the harmony search algorithm to optimize multiple objectives. Kanwar et al.~\cite{KanwarAS23} introduce a multi-objective random forest algorithm combined with a data resampling technique for CPDP to minimize false alarms and maximize detection probability while addressing class imbalance issues. These methods demonstrate that single-objective problem form easily leads to a bias toward specific types of predictions or models than multi-objective optimization problem (MOP) form, resulting in obtaining inferior models.

%they did not consider the internal nested hierarchical issues brought by the model parameters.

Multi-objective bilevel optimization (MBLO)—a highly active research field over the last decade—has emerged as a crucial topic in various domains such as engineering design, industrial processes, and decision-making processes~\cite{ChenDLTC24}. 
%MBLO inherently involves a complex optimization landscape, where both the upper and lower levels of the optimization process need to be simultaneously optimized, often with conflicting objectives. 
However, surprisingly, to the best of our knowledge, few works have considered MBLO in CPDP. This can be attributed to the following three imperative challenges.

\begin{itemize}

    \item In CPDP, using MBLO to search the combined space of ML pipelines and their hyperparameter is challenging: the space is highly dimensional, involves both categorical and continuous choices, and contains hierarchical dependencies (e.g., the hyperparameter of an ML technique is only meaningful if that technique is chosen). The high-dimensional search space makes optimization intricate and time-consuming.

    \item Another challenge posed by the high-dimensional search space in CPDP is the adaptability of optimization across different projects. Since the data distribution, characteristics, and defect models differ between projects, it is challenging to develop a one-size-fits-all ML technique that effectively generalizes across diverse datasets and addresses the unique needs of each project.
    
    \item Under the MBLO framework, the upper-level optimization must flexibly adjust its strategy to accommodate different projects, while the lower-level optimization seeks the optimal solution for each project. This nested optimization increases complexity, often leading to local optima and making it hard to find global solutions.

\end{itemize}

%Therefore, the main challenges in CPDP models can be summarized as: 1) The variability in cross-project data with its feature inconsistency and distributional discrepancies complicates CPDP, demanding advanced methods for effective defect prediction; 2)The CPDP models and their hyperparameter constitute a high-dimensional search space, making it complex and time-consuming to find the optimal combination of hyperparameter; 3) Given the significant differences between project distributions and characteristics, relying on a single model often fails to meet performance expectations and is prone to overfitting; 4) Due to the complex and diverse nature of CPDP, optimization algorithms may fall into local optima.

Given the aforementioned consideration, innovative approaches are required to leverage limited, heterogeneous historical data effectively and improve the performance of CPDP model. In fact, different projects often use distinct programming languages, frameworks, or development environments, which lead to inconsistent features for prediction. Moreover, the data from different projects may exhibit significant disparities in quality and scale, resulting in inconsistencies in data distribution. 
%These factors contribute to feature redundancy (i.e., irrelevant or duplicate features) and distribution discrepancies (i.e., differences in data distribution between source and target projects) within the high dimensional space of cross-project data.
%, ultimately affecting the model's predictive performance. 
%As identified by Yu et al.~\cite{YuJQ16}, instances and features are two main factors influencing CPDP performance.
To address these challenges, some studies integrate feature selection with transfer learning to enhance model performance and mitigate issues related to feature redundancy and distribution discrepancies. For example, the study in~\cite{HosseiniTM18} examines the use of Genetic Instance Selection (GIS) as a transfer learning approach for CPDP, focusing on optimizing the training dataset through feature selection and mitigating label noise, which significantly improves the effectiveness of GIS compared to other variants.
An empirical study presented in~\cite{WZGJ19} combined source selection, including feature selection and source project selection, with a transfer learning model to address the instability issues of Transfer Component Analysis+ (TCA+)~\cite{NamPK13} in CPDP. To tackle information redundancy, Lei et al.~\cite{LeiXH20} applied the maximum information coefficient for feature filtering based on project differences and integrated a transfer learning technique, which emphasizes multiple classification error samples. Subsequent work~\cite{LeiXWNSZ22} further focused on reducing differences between projects at the feature level through feature selection and introduced an advanced distance weight method, leading to a more accurate prediction model. Li et al.~\cite{LiZJXGR23} focused on selecting suitable source data for CPDP at the project level, using a data selection predictor to address data biases, parameter tuning, and class imbalance. Additionally, the feature set was refined through metric correlation analysis, with a correlation-based algorithm being used to select representative metrics.

%These studies mainly rely on specific methods, which may be difficult to generalize to different CPDP scenarios. Although innovative solutions are provided, their effectiveness is often closely related to specific datasets or backgrounds. 

Despite these advantages, complexity increases as feature selection and transfer learning techniques introduce additional configurable parameters. Furthermore, although many methods have been investigated in CPDP, no consensus has been reached on the best approach~\cite{HosseiniTG19}. 
This lack of consensus underscores the necessity for a more systematic approach. Drawing inspiration from Automated ML (AutoML), which automates tasks such as feature engineering, model selection, and HPO, making it easier to develop high-performing models with minimal human intervention. Automating feature selection, transfer learning, classification, and their corresponding hyperparameter may enhance the effectiveness of the model development pipeline. However, these techniques and their associated hyperparameter further expand the search space in CPDP. Moreover, it is difficult to develop a universal ML technique that adapts to different projects' distinct characteristics and effectively generalizes across diverse datasets. As a solution, ensemble learning can be utilized to take advantage of various ML models, each tailored to different characteristics, thereby enhancing the overall performance~\cite{HeZRS16}. 
%Current CPDP methods are primarily focused on optimizing single learners for optimal performance. However, ensembles of multiple promising learners are often known to yield better results than individual learners. 
%This approach has been successfully demonstrated in various competitions; for example, He et al.~\cite{HeZRS16} secured first place in the ILSRVC2015 competition by ensembling predictions from multiple learners, and similarly, ensemble methods frequently achieve top rankings in data science competitions like Kaggle~\cite{Hoch15,abs-2009-07701}. 

%Inspired by these successes, modern AutoML systems such as \texttt{Auto-sklearn}, \texttt{Auto-Pytorch}~\cite{ZimmerLH21}, and \texttt{VolcanoML}~\cite{LiSZJLDZY00021} have adopted post-hoc ensemble strategies that leverage all base learners throughout the optimization process, resulting in superior empirical performance compared to single best learners. Despite the effectiveness of these post-hoc ensemble designs in practice, their objectives do not fully align with those of CPDP methods. Therefore, the challenge of simultaneously guiding the search for both performance and diversity in a specific CPDP task remains an open question.

%This approach not only mitigates the risk of overfitting associated with high-dimensional spaces but also leverages the diversity among models to improve generalization across different projects.

%Bearing the above concerns in mind, 
In light of the above factors, we propose a tool combining a MBLO framework with AutoML techniques (dubbed \our). Our main contributions are outlined as follows:

\begin{enumerate}

	\item  To comprehensively explore trade-offs and account for the dependencies between ML pipelines and their hyperparameter, we formulate CPDP as a MBLO problem (MBLOP). The upper-level problem is designed as a multi-objective combinatorial optimization problem aimed at finding the optimal ML pipelines, while the lower-level problem is dedicated to optimizing hyperparameters for these ML pipelines.
 
	\item By utilizing limited and heterogeneous historical data, we automatically select the optimal combination of feature selection, transfer learning, and classification from a given portfolio to extract cross-project information, thereby addressing high-dimensional complexity and enhancing the model’s generalization, robustness, and overall prediction performance.
	
	\item To address differences in project distributions and characteristics and to generalize across diverse datasets, we propose an ensemble learning method that incorporates both ensemble techniques and local search methods. This approach not only enhances performance but also ensures diversity among the ensemble members, further strengthening the effectiveness of the ensemble.
	
	\item To effectively explore the search space while achieving high adaptability and flexibility to accommodate multiple objectives, a multi-objective tabu search method is proposed to optimize the upper-level MOP while the Tree-structured Parzen Estimator (TPE)~\cite{BergstraBBK11} is employed to optimize the lower-level expensive optimization problem.

   \item Through extensive experiments and comparisons with existing technologies, \our~has demonstrated significantly superior performance, highlighting the effectiveness and importance of the associated techniques. Additionally, we have thoroughly examined the advantages these techniques offer in CPDP.
	
\end{enumerate}

The remainder of this paper is organized as follows. \pref{sec:preliminaries} introduces some background knowledge pertinent to this paper. \pref{sec:approach} delineates the implementation of the tool \our~proposed by us. The experimental setup is introduced in \pref{sec:settings}, and the results are presented and analyzed in \pref{sec:results}. In addition, \pref{sec:related} and \pref{sec:threats} review the related works and discuss the threats to validity, respectively. Finally, \pref{sec:conclusion} concludes this paper and sheds some light on future directions.

%% file: preliminaries.tex
%!TeX root=main.tex

\section{Preliminaries}
\label{sec:preliminaries}

BLOP is a hierarchical decision-making process where one optimization problem is nested within another. This nested nature of decision-making is similar to the Stackelberg competition model in game theory~\cite{Stackelberg52,ChenLLT24}. In the upper-level optimization task, a solution, in addition to satisfying its own constraints, must also be an optimal solution to another optimization problem, called the lower-level optimization problem. BLOPs have a wide range of implications in a variety of fields, such as chemical engineering~\cite{AvraamidouP19a}, security~\cite{YangSXZ19}, mobile cloud computing~\cite{GuanSQLGJ23}.

MBLOP extends BLOP by involving multiple conflicting objectives in at least one level. In MBLOPs, a common real-world decision-making scenario occurs when the upper level has multiple objectives, while the lower level has only a single objective, such as in network architecture search~\cite{QianHWLGGZZW22}, poisoning attack~\cite{CarnereroCanoMSC20} and railway alignment optimization problem~\cite{YangYHLW23}. Solving these problems requires not only balancing multiple objectives but also addressing the interaction between the two levels, namely, identifying a set of upper-level Pareto-optimal solutions that are feasible at the lower level. The MBLOP considered in this paper can be mathematically formulated as follows:

\begin{equation}
	\begin{array}{l l}
		\underset{\mathbf{x}^\mathrm{U}\in\Omega^\mathrm{U},\mathbf{x}^\mathrm{L}\in\Omega^\mathrm{L}}{\min}\mathbf{F}\left(\mathbf{x}^\mathrm{U},\mathbf{x}^\mathrm{L}\right),\\
		\mathrm{subject\ to}\quad \mathbf{x}^\mathrm{L}\in \underset{\mathbf{x}^\mathrm{L}\in\Omega^\mathrm{L}}{{\argmin\,}}f^\mathrm{L}\left(\mathbf{x}^\mathrm{U},\mathbf{x}^\mathrm{L}\right), \\
	\end{array}
	\label{eq:bilevel}
\end{equation}
where decision vector $\mathbf{x}=\left(\mathbf{x}^\mathrm{U},\mathbf{x}^\mathrm{L}\right)^\top$ consists of an upper-level decision variable $\mathbf{x}^\mathrm{U}=\left(x_1^\mathrm{U},\ldots,x_{n_\mathrm{u}}^\mathrm{U}\right)^\top$ and a lower-level decision variable $\mathbf{x}^\mathrm{L}=\left(x_1^\mathrm{L},\ldots,x_{n_\ell}^\mathrm{L}\right)^\top$. Note that, in our context, the upper-level problem is a discrete combinatorial problem with $\Omega^\mathrm{U}$ representing a search space defined by categorical variables. On the other hand, the lower-level $\Omega^\mathrm{L}$ is a hybrid search space, including integers, real numbers, and categorical variables. $\mathbf{F}(\mathbf{x})=\left(F_1(\mathbf{x}),\ldots,F_{m_\mathrm{u}}(\mathbf{x})\right)^\top$ and $f^\mathrm{L}\left(\mathbf{x}^\mathrm{U},\mathbf{x}^\mathrm{L}\right)$ are the upper- and lower-level objective vector(s), respectively. Based on this problem formulation, we have four basic definitions. Evolutionary algorithms have been widely accepted as an effective method for multi-objective optimization problems~\cite{LiNGY22,LiZLZL09,LiKM11,LiDY18,GaoNL19,LiLL22,ShanL21,LiCFY19,LiLLY24,CaoKWL12,LiKD15,LiKWTM13,LiK14,RuanLDL20,LiLY23,LiCSY19,LiZKLW14,LiLDMY20,LiDZZ17,Li19,LiKCLZS12,WuKZLWL15,LiKZD15,WuLKZZ17}, with applications in adversarial robustness~\cite{Zhou0M22,Williams0M22,WilliamsLM23b,WilliamsL23b,Williams023,WilliamsLM23a,WilliamsLM23}, parameter control~\cite{LiFK11,LiKWCR12,LiWKC13,LiFKZ14,SunL20,CaoKWLLK15,LiXT19,WuLKZZ19}, software engineering~\cite{Liu0020,LiX0WT20,LiXCT20,ZhouHSL24}, automated machine learning~\cite{LyuLHWYL23,HuangL23,LyuYWHL23,LyuHYCYLWH23}, smart grid~\cite{XuLA21,XuLA021,XuLL24a,XuLA22}, networking~\cite{BillingsleyLMMG20,BillingsleyMLMG20,GuanSQGLJ23,BillingsleyLMMG21,LiC23,ChenL23}, large language model tuning~\cite{YangL23}.

\begin{definition}
	Given an upper-level variable $\mathbf{x}^\mathrm{U}\in\Omega^\mathrm{U}$, the \underline{lower-level optimal solution} is defined as:
	\begin{equation}
		\mathbf{x}^\mathrm{L_*}=\left\{\mathbf{x}^\mathrm{L}\in\Omega^\mathrm{L}\big|f^\mathrm{L}\left(\mathbf{x}^\mathrm{U},\mathbf{x}^\mathrm{L}\right)< f^\mathrm{L}\left(\mathbf{x}^\mathrm{U},\mathbf{x}^\mathrm{L^\prime}\right),\forall \mathbf{x}^\mathrm{L^\prime}\in\Omega^\mathrm{L}\right\}.
	\end{equation}
\end{definition}

\begin{definition}
	The set of feasible solutions over which the upper-level optimization routine can optimize is called the \underline{inducible region}:
	\begin{equation}
		\mathcal{IR}=\left\{\left(\mathbf{x}^\mathrm{U},\mathbf{x}^\mathrm{L}\right)\big|\mathbf{x}^\mathrm{U}\in\Omega^\mathrm{U},\mathbf{x}^\mathrm{L}\in\mathbf{x}^\mathrm{L_*}\right\}.
	\end{equation}
\end{definition}

\begin{definition}
	Given two solutions $\mathbf{x}^1,\mathbf{x}^2\in\mathcal{IR}$, $\mathbf{x}^1$ is said to \underline{dominate} $\mathbf{x}^2$, denoted as $\mathbf{x}^1\preceq\mathbf{x}^2$, iff $\forall i\in\{1,\ldots,m_\mathrm{u}\}$, $F_i\left(\mathbf{x}^1\right)\leq F_i\left(\mathbf{x}^2\right)$ and $\mathbf{F}\left(\mathbf{x}^1\right)\neq\mathbf{F}\left(\mathbf{x}^2\right)$.
\end{definition}

\begin{definition}
	A solution $\mathbf{x}^\ast\in\mathcal{IR}$ is said to be \underline{Pareto-optimal} iff $\nexists\mathbf{x}^\prime\in\mathcal{IR}$ such that $\mathbf{x}^\prime\preceq\mathbf{x}^\ast$.
\end{definition}

%In this paper, our MBLOP aims to find a set of Pareto-optimal solutions, demonstrating the optimal trade-offs among multiple conflicting objectives at the upper level. This process not only reveals the interactions between objectives but also provides a comprehensive solution set for the decision maker in a hierarchical decision environment. Readers interested in further information can refer to some excellent survey papers~\cite{SinhaMD18,Dempe20,JesAE23}.

%% file: approach.tex
%!TeX root=main.tex

\section{The \our~Method}
\label{sec:approach}

In this section, we introduce the implementation details of our proposed \our\ step by step. 

\subsection{Overview of \our}
\label{sec:overview_of_algorithm}

\begin{figure*}[t!]
	\centering
	\includegraphics[width=\linewidth]{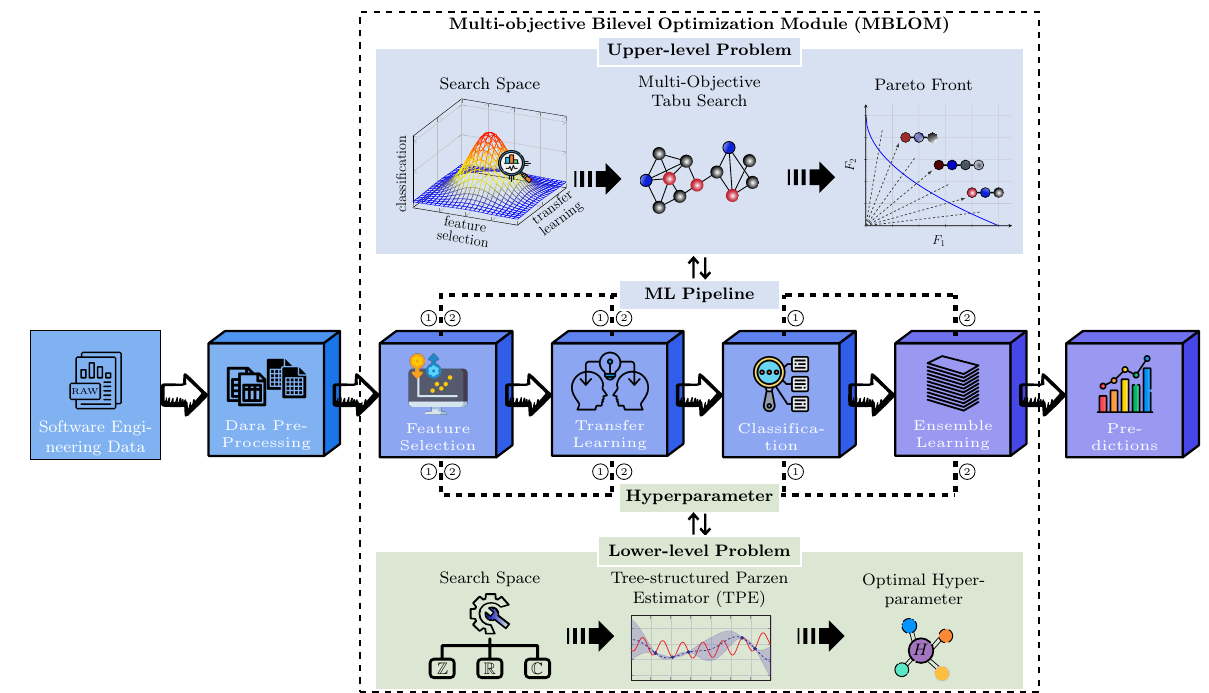}
	\caption{The overall architecture of \our\protect\footnotemark.}
	\label{fig:MBL-CPDP}
\end{figure*}
\afterpage{\footnotetext{Phase $1$, indicated by {\ding{192}}, includes feature selection, transfer learning, and classification. Phase $2$, denoted by {\ding{193}}, employs feature selection, transfer learning, and ensemble learning.}}

\pref{fig:MBL-CPDP} gives the overall architecture of \our, which consists of three core modules.
\begin{itemize}
    \item \underline{\textbf{Data pre-processing:}} In the context of CPDP, there are $N>1$ projects under consideration. One project is selected as the target domain to assess the model's generalization capability, while the remaining $N-1$ projects are used as the source domain for transfer learning. In practice, data in the target domain can be partitioned into training and testing sets according to the setup of the corresponding transfer learning algorithms. For instance, \cite{QiuLJ18} uses $90\%$ of data for training, while the remaining ones are used for testing purposes.
        %In contrast, algorithms without this requirement employ the entire dataset for testing, providing a comprehensive evaluation. 
	
    \item \underline{\textbf{MBLO module (MBLOM):}} In \our, we model the search for the optimal CPDP model as a MBLOP. In particular, to enhance the model's robustness across different projects, the upper-level optimization is formulated as a bi-objective optimization problem (see~\pref{sec:mbo_approach}). 
    $\blacktriangleright$ At the upper level, it searches for the optimal ML pipeline, which is a combination of feature selection, transfer learning, and classification algorithms (see~\pref{sec:component_structure}).  $\blacktriangleright$ At the lower level, it serves as a HPO routine that fine-tunes the hyperparameters associated with each pipeline in the upper level. $\blacktriangleright$ Following the best practices in the AutoML community~\cite{HKV2019}, we construct an ensemble learner as the model output. In particular, our ensemble learning method is designed to achieve a balance between predictive accuracy and model diversity (see~\pref{sec:ensemble_learning}). 
    $\blacktriangleright$ An efficient algorithm is proposed to tackle this problem, meticulously designed to achieve high adaptability and flexibility across different projects (see~\pref{sec:alg_details_Imple}).

%        The lower level focuses on hyperparameter fine-tuning of these chosen combinations to achieve the best performance. Moreover, a multi-objective form is employed to understand the effect of different metrics on CPDP and enhance robustness across various projects (detailed in~\pref{sec:mbo_approach}). Ensemble learning is known to improve model performance. While some ensemble-based AutoML approaches have shown effectiveness in practice, they do not always align with the objectives of CPDP. To address this, we introduce an ensemble learning method that balances performance and diversity, further discussed in~\pref{sec:ensemble_learning}. Finally, we introduce a multi-objective bilevel optimization algorithm for this problem, with the methodology and its implementation presented in~\pref{sec:alg_details_Imple}.
	
	\item \underline{\textbf{Prediction:}} This last module is used to evaluate the fitness value of the CPDP model found by the MBLOM module. In principle, any off-the-shelf performance metric in the ML community can serve this purpose. In \our, we adopt the area under the receiver operating characteristic curve (AUC)~\cite{ZhouYLCLZQX18} as the metric for fitness evaluation, given its robustness to different parameter settings and imbalanced data distribution, which is not uncommon in defect prediction.
\end{itemize}

\subsection{Hierarchical Structure of MBLOM}
\label{sec:mbo_approach}

In view of its nested structure, we present the module from upper- and lower-levels.

\subsubsection{Upper-level Multi-objective Problem}
\label{sec:ULMOP}
In CPDP, most optimization studies aim to determine the "best" solution corresponding to the minimum or maximum value of one objective function, as well as \texttt{Bilo-CPDP}~\cite{LiXCT20}. However, one objective function often fails to accurately represent the defect module's performance due to the high heterogeneity and diversity of software project data. Therefore, we design a MOP form at the upper level to improve the flexibility of the model for various projects. The design of this multi-objective framework not only offers high adaptability to existing optimization objectives, but also allows for introducing new optimization objectives as project demands and decision preferences change.

Without loss of generality, the multi-objective optimization in this paper adopts a straightforward conflicting objective, i.e., $F_1$ and $1-\sqrt{F_1}$, with $F_1$ defined as AUC metric, as mentioned in \pref{sec:overview_of_algorithm}. Reflecting on \pref{eq:bilevel}, the upper-level problem incorporates two decision variables: the ML pipeline $\mathbf{x}^\mathrm{U}$ (selected from a candidate set, as outlined in \pref{tab:fs} to \pref{tab:classifiers}) and its corresponding optimal hyperparameters $\mathbf{x}^\mathrm{L}$. Detailed components of MBLOM will be introduced in \pref{sec:component_structure}.
Notably, these learners are selected from previous CPDP studies and the advanced ML library scikit-learn~\cite{PedregosaVGMTGBPWDVPCBPD11}. Considering the flexibility of combining these learners and the ability of the ensemble learner to incorporate any number of classifiers, this results in a minimum of $1,056$ ML pipelines, with the potential to reach up to $4,325,310$ pipelines. This design ensures a wide range of methods to meet the diverse datasets and project requirements of CPDP. Moreover, software engineers could adjust algorithmic combinations according to their specific needs and preferences, increasing the flexibility and customizability of the model in responding to software engineering (SE) challenges. 
\input{tables/feature_selections}
\input{tables/transfer_learners}
\input{tables/classifiers}

%Finally, it is worth noting that the lower-level optimal $\mathbf{x}^\mathrm{L}$ for a given $\mathbf{x}^\mathrm{U}$ remains undetermined before the lower-level optimization. The computation of the upper-level function is embedded in and constrained by the results of lower-level optimization.

\subsubsection{Lower-level Single-objective Problem}
\label{sec:LLSOP}

In addition to selecting the most suitable ML pipeline, fine-grained tuning of hyperparameters, rather than solely relying on default settings or empirical configurations, is also critical to enhancing the model's performance~\cite{Tantithamthavorn19}. Therefore, we implement HPO at the lower level to identify the most effective hyperparameters. Compared to traditional manual tuning, the lower-level problem focuses on matching the optimal hyperparameter configurations with the algorithm combinations determined by the upper level. 

Recalling \pref{eq:bilevel}, for a given ML pipeline $\mathbf{x}^\mathrm{U}$, the lower-level optimization determines its corresponding optimal hyperparameter configuration $\mathbf{x}^\mathrm{L}$. HPO guides the optimization direction to obtain the greatest AUC metric. The computation of the upper-level function is embedded in and constrained by the results of lower-level optimization. The hyperparameter space comprises three types: integers [$\mathbb{Z}$], real numbers [$\mathbb{R}$], and categorical variables [$\mathbb{C}$], with each category's specific ranges and classifications detailed in \pref{tab:fs} to \pref{tab:classifiers}.

\subsection{Basic Components of MBLOM}
\label{sec:component_structure}

Although some prior works have considered the HPO for CPDP models, most of them mainly consider optimizing classification~\cite{QuCZJ18,Ozturk191,LiXCT20,KwonRB23}. In these methods, \texttt{Bilo-CPDP} is a more comprehensive approach, considering both transfer learning and classification~\cite{LiXCT20}. However, it ignores the characteristics of feature space and model capability. 

In CPDP problems, data heterogeneity across the project environment may be introduced due to the differences in project size, domain, language, and defect distribution~\cite{Saeed2023}. To capture the characteristics of feature space and construct an efficient knowledge discovery and mining model, \our~integrates feature selection, transfer learning, classification, and ensemble learning. These components, which are specified below, are designed to enhance model adaptability and accuracy.

 %Feature selection is an effective and efficient data preprocessing strategy for selecting relevant features, as it selects the most relevant features and reduces the data's dimensionality, improving model performance and interpretability~\cite{LiCWMTTL16}. Furthermore, ensemble learning methods exploit multiple ML algorithms to produce weak predictive results based on features extracted through a diversity of projections on data and fuse results with various voting mechanisms to achieve better performances than those obtained from any constituent algorithm alone~\cite{DongYCSM20}.

\begin{itemize}
	\item \underline{\textbf{Feature Selection:}}  CPDP datasets often contain numerous features. Ignoring the feature space characteristics may lead to inappropriate treatment of important and irrelevant features, resulting in a waste of computational resources. Feature selection reduces the input variable by using only relevant data and gets rid of noise in data. Selecting information-rich features not only reduces noise but also helps build models that can be effectively generalized across projects. 
	
	\item \underline{\textbf{Transfer Learning:}} It is not uncommon for CPDP datasets to contain imbalanced, high-dimensional, noisy data~\cite{Saeed2023}. Traditional ML methods may fail to capture multiple characteristics and the underlying structure of data. Transfer learning, which can transfer knowledge across different domains, has been shown to overcome the challenges of data scarcity and distribution for CPDP problems~\cite{LiX0WT20}. 
	
	\item \underline{\textbf{Classification:}} Classification is a predictive modeling problem where the class label is anticipated for a specific example of input data. Selecting appropriate classification methods and learning historical data are critical to improving predictive accuracy.

    \item \underline{\textbf{Ensemble Learning:}} Ensemble learning stands out as a powerful technique in ML, offering a robust approach to improving model performance and predictive accuracy. Combining the strengths of multiple individual models, ensemble methods can often outperform any single model, making them valuable in the ML toolkit (see \pref{sec:ensemble_learning}).
	
\end{itemize}

%Feature selection, transfer learning, and classification are combined at the upper level to maximize model generalization across different SE datasets. Using these components, we attempt to display a systematic optimization model to improve CPDP performance.

\subsection{Ensemble Learning}
\label{sec:ensemble_learning}
As discussed in \pref{sec:introduction}, relying on a single classifier makes it difficult to meet the expected performance, given that the distribution and characteristics of different projects may be significantly different. Ensemble learning, a powerful ML paradigm, is valued for its generalization capability and adaptability to data diversity~\cite{HeZRS16}. This paper proposes an ensemble learning method specifically designed for CPDP, which aims to maximize the ensemble effect while maintaining diversity among ensemble members. This process, also called phase $2$, includes the following five steps:

\begin{enumerate}[label=\arabic*.]
	\item \underline{\textbf{Generation:}} We employ the model exploration and HPO operations outlined in~\pref{sec:mbo_approach}, referred to as phase $1$, to generate some ML pipelines with good performance, which include feature selection, transfer learning, and classification. 
	
	\item \underline{\textbf{Candidate Pool Construction:}} This step involves: 1) Assessing the performance of ML pipelines that are generated in phase $1$; 2) To maximize the preservation of ML pipelines with good performance, greedy ensemble selection with replacement~\cite{TsoumakasPV09} is used to select the top $N^c$ ML pipelines for constructing candidate pool $\mathcal{P}^c$; 3) If the same classifier appears in $\mathcal{P}^c$, it indicates that this classifier is the only suitable option for this task, and therefore, we will omit the subsequent steps. Otherwise, we repeat the following three steps from the best to the worst ML pipelines until the predefined condition (detailed in~\pref{sec:parameters}) is met.
	
	\item \underline{\textbf{Diversity Assessment:}} Considering ensemble components are updated in each iteration, and the ensemble may include various configurations, it is difficult to model the diversity of the ensemble with configurations and multiple classifiers~\cite{ShenL0TZ022}. Thus, we adopt a pair-wise measurement method for diversity assessment. Specifically, if the number of classifiers in $\mathcal{P}^c$ is smaller than the ensemble selection size, we use them directly; otherwise, to balance effectiveness and diversity in the ensemble learner, we select several well-performing classifiers and then maximize the $Q$-statistic~\cite{Yule90} to select some classifiers with diversity.
 
% to balance the effectiveness and diversity in the ensemble learner, we first select some top performing classifiers. Then, $Q$-statistic~\cite{Yule90} is adopted to quantify the predictive consistency between classifiers and enhance diversity. By maximizing the $Q$-statistic principle for selection, we construct a diversified ensemble learner comprising $N^s$ classifiers, ensuring both top performance and diversity in the model collection.

	\item \underline{\textbf{Refinement and Aggregation:}} To further refine the ensemble learner, we propose a progressive classifier pruning mechanism that systematically excludes the less impactful classifiers. Considering the variety of classification methods in~\pref{tab:classifiers}, we employ stacking~\cite{Wolpert92} with logistic regression (LR) to address this heterogeneous ensemble.
	Particularly, when multiple classifiers exist, their results are used as features by the meta-learner to generate the final prediction. Subsequently, the pruning mechanism systematically removes classifiers with minor contributions to performance, after which the remaining classifiers are again aggregated for the next prediction round. This cycle of aggregation, followed by pruning, continues until only the best candidate remains. When only a single classifier is left, we bypass aggregation and directly use its prediction.
	
	\item \underline{\textbf{HPO:}} Through the steps above, we obtain a set of ensemble learners. Given that the optimal parameter configuration under ensemble learning may differ from that of phase $1$, we research the parameters of the ensemble learners, feature selection, and transfer learning techniques instead of directly adopting the parameter settings used before to ensure optimal model performance. The process of selecting the best parameters is the same as described in~\pref{sec:LLSOP}.
\end{enumerate}

\subsection{Algorithm Details and Implementation}
\label{sec:alg_details_Imple}
As discussed earlier, the overall optimization objective is to identify the optimal ML pipelines and determine the optimal hyperparameters for the selected scheme. A straightforward method would be obtained by testing all possible ML pipelines and their configurations, but this approach is computationally expensive and time-consuming. Therefore, we use evolutionary computation to find the optimal solution efficiently, avoiding exhaustive search. Next, we will detail the specific implementation of the MBLO algorithm.

\subsubsection{Upper-level Optimization Algorithm}
\label{ULOA}
The upper-level optimization problem is a multi-objective combinatorial optimization, which requires finding non-dominated solutions that meet multi-objective criteria from at least $1,056$ ML pipelines. The complex search space and the existence of many local optimal solutions increase the difficulty of finding optimal solutions. 

To overcome these problems, we follow Li's suggestion~\cite{LiXCT20} and adopt tabu search, a meta-heuristic search algorithm that avoids falling into local optima by maintaining a tabu list and using historical information to guide the search direction~\cite{GloverL97,TanKLY03}. 
%By traversing the search space, it effectively finds the global optimum, making it a good choice for combinatorial optimization.
However, when dealing with MOP, tabu search exhibits limitations in balancing trade-offs between multiple objectives and exploring a broader search space. To deal with these problems and fully leverage the potential of tabu search, we propose a multi-objective tabu search algorithm. This algorithm not only retains the advantages of tabu search but also effectively handles multi-objective combinatorial optimizations. 
%by further guiding the search direction, maintaining a non-dominated solution set, and introducing a multi-objective optimization mechanism, thereby enhancing search efficiency and solution quality in complex search spaces.
The overall pseudocode of MBLO algorithm is illustrated in \pref{alg:moba}: Initialization (lines $1-4$); phase $1$ (lines $5-12$); phase $2$ (lines $13-21$). 

In the initialization part, for each ML pipeline $\mathbf{x}_i^\mathrm{U}$, \pref{alg:ulea} is embedded to evaluate the upper-level multi-objective objective function. 
As shown in \pref{alg:ulea}, we calculate the optimized lower-level hyperparameters $\mathbf{x}_i^\mathrm{L_*}$, the lower-level objective function $f^\mathrm{L}\left(\mathbf{x}_i^\mathrm{U},\mathbf{x}^\mathrm{L_*}\right)$, and the upper-level multi-objective function $\mathbf{F}(\mathbf{x}_i^\mathrm{U},\mathbf{x}_i^\mathrm{L_*})$ (lines $1-3$). Afterwards, the tabu list and candidate pool are updated (lines $4-10$).

\begin{algorithm}[t!]
	\DontPrintSemicolon
	\footnotesize
	%\scriptsize
	\caption{\footnotesize Multi-objective Bilevel Algorithm.}
	\label{alg:moba}
	\KwIn{
		\\
		~~~~$\mathcal{S}, \mathcal{T}, \mathcal{C}$: portfolio of feature selection techniques $s$, transfer learning techniques $t$, and classifiers $c$, respectively\\
		~~~~$N^u$: upper-level population size\\
		~~~~$B_1,B_2$: phase $1$ and phase $2$ time budget\\
	}
	\KwOut{
		\\
		~~~~$\mathcal{P}$: nondominated ML pipelines and their optimal hyperparameters}

	$\mathcal{\ell}^t\leftarrow\emptyset, \mathcal{P}\leftarrow\emptyset, \mathcal{F}\leftarrow\emptyset, \mathcal{P}^c\leftarrow\emptyset, \mathcal{F}^c\leftarrow\emptyset$\\
	
	$\mathcal{P}^\mathrm{U}\leftarrow\left\{\mathbf{x}_1^\mathrm{U},\mathbf{x}_2^\mathrm{U},\ldots,\mathbf{x}_{N^u}^\mathrm{U}\right\},\mathbf{x}_i^\mathrm{U}\leftarrow(s,t,c), s\in\mathcal{S}, t\in\mathcal{T}, c\in\mathcal{C}$ \\

	\ForEach{$\mathbf{x}_i^\mathrm{U}\in\mathcal{P}^\mathrm{U}$}  
	{
		$\{\mathcal{\ell}^t, \mathcal{P},\mathcal{F},\mathcal{P}^c,\mathcal{F}^c\}\leftarrow\textsc{UpperLevel}(\mathbf{x}_i^\mathrm{U}, \mathcal{\ell}^t, \mathcal{P},\mathcal{F},\mathcal{P}^c,\mathcal{F}^c)$\\
	}
	
	\While{The computational budget $B_1$ is not exhausted}
	{
		$\mathcal{P}^{\mathrm{U}'}\leftarrow$Randomly or use SBX and PM to generate $N_u$ ML pipelines using \pref{eq:cm1} and \pref{eq:cm2}\\
		\ForEach{$\mathbf{x}_i^\mathrm{U'}\in\mathcal{P}^{\mathrm{U}'}$}  
		{  
			$\delta\leftarrow$Retrieve the neighbors of $\mathbf{x}_i^\mathrm{U'}$\\
			\ForEach{$\mathbf{x}^n\in\delta \wedge \mathbf{x}^n \notin \mathcal{\ell}^t$}
			{ 
				$\left\{\mathcal{\ell}^t, \mathcal{P},\mathcal{F},\mathcal{P}^c,\mathcal{F}^c\right\}\leftarrow\textsc{UpperLevel}(\mathbf{x}^n,\mathcal{\ell}^t,\mathcal{P},\mathcal{F},\mathcal{P}^c,\mathcal{F}^c)$\\
			}
			$\left\{\mathcal{P},\mathcal{F}\right\}\leftarrow$ Use nondominated sorting and crowding distance to select $N^u$ solutions of $\mathcal{P}$ and their upper-level function values $\mathcal{F}$\\
			$\mathcal{\ell}^t\leftarrow$ Remove the ML pipeline(s) not in $\mathcal{P}$ from $\mathcal{\ell}^t$
		}
		
	}
	
	\While{The computational budget $B_2$ is not exhausted}
	{
		\ForEach{$\mathbf{x}_i^\mathrm{U''}\in\mathcal{P}^c$}  
		{
			$\mathcal{P}^e\leftarrow \mathcal{P}^c, (s,t,c) \leftarrow \mathbf{x}_i^\mathrm{U''}$\\
			\While{the classifier in $\mathcal{P}^e$ is not None}
			{
				$(\mathcal{C}^e, f^e)\leftarrow\textsc{ConstructEnsemble}(\mathcal{P}^e)$\\
				$ \mathbf{x}_i^\mathrm{U''}\leftarrow (s,t,\mathcal{C}^e)$\\
				$\{\mathcal{\ell}^t, \mathcal{P},\mathcal{F},\mathcal{P}^c,\mathcal{F}^c\}\leftarrow\textsc{UpperLevel}(\mathbf{x}_i^\mathrm{U''}, \mathcal{\ell}^t, \mathcal{P},\mathcal{F},\mathcal{P}^c,\mathcal{F}^c)$\\
				$\mathcal{P}^e\leftarrow$remove the classifier with minor contribution to $f^e$\\
				$\left\{\mathcal{P},\mathcal{F}\right\}\leftarrow$ Use nondominated sorting and crowding distance to select $N^u$ solutions of $\mathcal{P}$ and their upper-level function values $\mathcal{F}$\\
			}
		}
	}
	\Return $\mathcal{P}$
	
\end{algorithm}

\begin{algorithm}[t!]
	\DontPrintSemicolon
	\footnotesize
	%\scriptsize
	\caption{\footnotesize $\textsc{UpperLevel}$: Upper-level Evaluation.}
	\label{alg:ulea}
	\KwIn{
		\\
		~~~~$\mathbf{x}_i^\mathrm{U}, \mathcal{\ell}^t$: ML pipeline, tabu list\\
		~~~~$\mathcal{P},\mathcal{F}$: ML pipelines and their hyperparameters, upper-level objective function values\\
		~~~~$\mathcal{P}^c, \mathcal{F}^c$: ML pipelines and upper-level function values of the candidate pool\\
		~~~~$N^c$: candidate pool size}
	\KwOut{
		\\
		~~~~$\mathcal{\ell}^t, \mathcal{P}, \mathcal{F}, \mathcal{P}^c, \mathcal{F}^c$: same as those of the \textbf{Input}\\
	}

	$\left\{\mathbf{x}_i^\mathrm{L_*}, f^\mathrm{L}\left(\mathbf{x}_i^\mathrm{U},\mathbf{x}^\mathrm{L_*}\right)\right\}\leftarrow\textsc{LowerLevel}(\mathbf{x}_i^\mathrm{U})$\\
	$\mathbf{F}(\mathbf{x}_i^\mathrm{U},\mathbf{x}_i^\mathrm{L_*})\leftarrow$ Evaluate the upper-level objective functions\\
	$\mathcal{P}\leftarrow\mathcal{P} \bigcup (\mathbf{x}_i^\mathrm{U},\mathbf{x}_i^\mathrm{L_*}),\mathcal{F}\leftarrow\mathcal{F} \bigcup \mathbf{F}(\mathbf{x}_i^\mathrm{U},\mathbf{x}_i^\mathrm{L_*})$\\
	
	\tcc*[l]{Construct the Candidate Pool}       		$\mathcal{P}^c\leftarrow\mathcal{P}^c \bigcup \mathbf{x}_i^\mathrm{U}, \mathcal{F}^c\leftarrow\mathcal{F}^c\bigcup \mathbf{F}(\mathbf{x}_i^\mathrm{U},\mathbf{x}_i^\mathrm{L_*})$\\
	\uIf{$|\mathcal{P}^c|>N^c$}
	{
			$\left\{\mathcal{P}^c,\mathcal{F}^c\right\}\leftarrow$ Use nondominated sorting and crowding distance to select $N^c$ solutions of $\mathcal{P}^c$ and their upper-level function values $\mathcal{F}^c$\\
	}

	\If{$\mathbf{x}_i^\mathrm{U}\notin\mathcal{\ell}^t$}
	{
		$\mathcal{\ell}^t\leftarrow\mathcal{\ell}^t \bigcup \left\{\mathbf{x}_i^\mathrm{U}\right\}$
	}
	
	\Return $\mathcal{\ell}^t, \mathcal{P}, \mathcal{F}, \mathcal{P}^c, \mathcal{F}^c$
	
\end{algorithm}

In phase $1$, to effectively explore and exploit the search space, we combine simulated binary crossover (SBX)~\cite{DebA95}, polynomial mutation (PM)~\cite{DebG1996}, and random search to generate the offspring $\mathbf{x}_i^\mathrm{U'}$. The formula is as follows:

\begin{equation}
	\alpha_i = 0.5\left[\left(1 + \beta_q\right)\mathbf{x}_i^\mathrm{U}+\left(1 - \beta_q\right)\mathbf{x}_j^\mathrm{U}\right].
	\label{eq:cm1}
\end{equation}

\begin{equation}
\mathbf{x}_i^\mathrm{U'} =
	\begin{cases}
		\alpha_i + \delta_q \left(u^\mathrm{U} - \mathbf{\ell}^\mathrm{U}\right) & \text{if } r_1 > \frac{1}{n_\mathrm{u}}, \\
		\mathbf{\ell}^\mathrm{U} + r_2\left(u^\mathrm{U} - \mathbf{\ell}^\mathrm{U}\right) & \text{otherwise}.
	\end{cases}
\label{eq:cm2}
\end{equation}
where $\beta_q$ and $\delta_q$ are calculated based on the distribution indices $\eta_c$ and $\eta_m$ of SBX and PM, respectively, controlling the intensity of crossover and mutation. 
$\mathbf{x}_i^\mathrm{U}$ and $\mathbf{x}_j^\mathrm{U}$ are the upper-level variables randomly selected from that of $\mathcal{P}$. $u^\mathrm{U}$, $\mathbf{\ell}^\mathrm{U}$, and $n_u$ are the upper and lower bounds, and dimension of the upper-level variables, respectively. \(r_1\) and \(r_2\) are random numbers in the interval [0, 1]. Next, tabu search carries out a neighborhood search where the neighborhood of $\mathbf{x}_i^\mathrm{U'}$ is restricted by the search history of previously visited solutions, and the function values are calculated (lines $8-10$). Facing the challenges of high dimensionality and complex search spaces in CPDP tasks, it is crucial to accurately identify high-quality solutions and maintain a balance among multiple objectives. To address these challenges, we employ non-dominated sorting~\cite{DebAPM02} to select solutions to ensure the approximation of the Pareto front. Additionally, crowding distance~\cite{DebAPM02} is used to maintain diversity among the retained non-dominated solutions, preventing premature convergence to local optima (line $11$). Finally, the tabu list is updated (line $12$).

In phase $2$, some ensemble learners are constructed from the candidate pool through a dynamic, iterative process (line $17$ of \pref{alg:moba}). The corresponding pseudocode is provided in \pref{alg:ce}, and the ensemble learner construction begins with the best-performing pipeline from the candidate pool. Due to the recombination of these learners, a preliminary performance evaluation is conducted for the new ML pipelines (lines $3-6$). Three scenarios are considered depending on the number of classifier types in the candidate pool, and the final outputs are returned (lines $7-24$ of \pref{alg:ce}).  Next, the selected learners are re-evaluated at the upper level (lines $18-19$ of \pref{alg:moba}). Classifiers with minor contributions are then removed from the ensemble (line $20$). Non-dominated sorting is used to select individuals from the population (line $21$). The optimal ML pipeline is dynamically selected through this iterative process, and the final result is output (line $22$).

\begin{algorithm}[t!]
	\DontPrintSemicolon
	\footnotesize
	%\scriptsize
	\caption{\footnotesize $\textsc{ConstructEnsemble}$: Ensemble Learner.}
	\label{alg:ce}
	\KwIn{
		\\
		~~~~$\mathcal{P}^e$: ML pipelines of the candidate pool\\
		~~~~$N^s$: ensemble selection size
	}
	\KwOut{
		\\
		~~~~$\mathcal{C}^e, f^e$: ensemble learner, lower-level function value
	}
	
	$\mathcal{C}'\leftarrow$ Identify non-duplicate classification methods from $\mathcal{P}^e$\\
	
	$(s,t,c) \leftarrow \mathbf{x}_i^\mathrm{U}, f^e\leftarrow\emptyset, \mathcal{C}''\leftarrow\emptyset,\mathcal{C}^e\leftarrow\emptyset$\\
	\ForEach{classification method $c'\in\mathcal{P}^c$}  
	{
		$\mathbf{x}_i^\mathrm{U'}\leftarrow(s, t, c')$\\
		$\left\{\mathbf{x}_i^\mathrm{L_*}, f^\mathrm{L}\left(\mathbf{x}_i^\mathrm{U'},\mathbf{x}^\mathrm{L_*}\right)\right\}\leftarrow\textsc{LowerLevel}(\mathbf{x}_i^\mathrm{U'})$\\
		$f^e\leftarrow f^e\bigcup f^\mathrm{L}\left(\mathbf{x}_i^\mathrm{U'},\mathbf{x}^\mathrm{L_*}\right)$
	}
	
	\uIf{$|\mathcal{C}'|>N^s$}
	{
		
		$\mathcal{C}''\leftarrow\left\{c\in C'\big|c~\text{is among the top} \left\lfloor\frac{N^s}{2}\right\rfloor \text{values of}~f\right\}$\\
		\While{$|\mathcal{C}''|< N^s$ and $\exists c\in \mathcal{C}',c \notin \mathcal{C}''$}
		{
			$q\leftarrow0$\\
			\ForEach{unique pair$(c^1,c^2)$ in pipelines of $\mathcal{C}'\setminus\mathcal{C}^e$}
			{ 
				$q'\leftarrow$ Calculate the $Q$-statistic value of $(c^1,c^2)$\\
				\If{$q'>q$}
				{ 
					$q\leftarrow q'$\\
					\uIf{$f^e(c^1)>f^e(c^2)$}
					{$\mathcal{C}''\leftarrow\mathcal{C}''\bigcup c^1,\mathcal{C}'\leftarrow\mathcal{C}' \setminus c^1$}
					\Else{$\mathcal{C}''\leftarrow\mathcal{C}''\bigcup c^2,\mathcal{C}'\leftarrow\mathcal{C}' \setminus c^2$}
				}
			}
		}
		$\mathcal{C}^e\leftarrow$ Build the ensemble learner using $\mathcal{C}''$ according to stacking
	}
	\uElseIf{$1<|\mathcal{C}'|\leq N^s$}{
		$\mathcal{C}^e\leftarrow$ Build the ensemble learner using $\mathcal{C}'$ according to stacking
	}
	\Else{$\mathcal{C}^e\leftarrow \mathcal{C}'$}

	\Return $\mathcal{C}^e, f^e$
	
\end{algorithm}

\subsubsection{Lower-level Optimization Algorithm}
\label{LLOA}
In~\our, the lower-level optimization objective is to select optimal hyperparameters for the ML pipeline. This task encompasses three types of hyperparameters, each with its own search range, creating a highly heterogeneous search space. The mixed-type search space poses a significant challenge in identifying the optimal hyperparameters. Moreover, model evaluation in CPDP tasks typically involves complex data processing and model training processes, which require substantial computational resources, making the HPO an expensive optimization problem.

To address this challenge, we adopt TPE~\cite{BergstraBBK11} for optimization. TPE guides the search process by constructing a probabilistic model to handle hybrid spaces effectively. The flexibility of TPE adapts to the HPO requirements of different learners and provides an effective hyperparameter selection scheme for multiple learners in CPDP. In addition, TPE reduces the overall computational cost of HPO by minimizing the reliance on expensive function evaluations.

As shown in~\pref{alg:llea}, TPE initially determines the configuration space $\Theta_{\mathbf{x}_i^\mathrm{U}}$ based on $\mathbf{x}_i^\mathrm{U}$ (line $1$). It then samples a set of hyperparameters $\mathcal{H}$ within $\Theta_{\mathbf{x}_i^\mathrm{U}}$ using space-filling and evaluates the corresponding AUC values $\mathbf{f}^\mathrm{L}$ as the performance measure (lines $2-3$). In the main loop, the core of TPE is constructing a cost-effective surrogate model using $\mathcal{H}$ and $\mathbf{f}^\mathrm{L}$ based on Bayesian optimization principles (line $5$). This surrogate model reduces the reliance on expensive evaluations by estimating hyperparameters with potentially higher performance through an acquisition function (line $6$). Next, the CPDP model calculates the lower-level objective function using these hyperparameters and adds the new solutions and corresponding values to $\mathcal{H}$ and $\mathbf{f}^\mathrm{L}$, respectively (lines $7-8$). Through iterations, TPE gradually narrows the search space and refines the selection of hyperparameters. Finally, the algorithm outputs the optimized hyperparameters and their corresponding lower-level objective values, providing a set of finely tuned model hyperparameters for the CPDP task (line $9$).

\begin{algorithm}[t!]
	\DontPrintSemicolon
	\footnotesize
	%\scriptsize
	\caption{\footnotesize $\textsc{LowerLevel}$: Lower-level Algorithm.}
	\label{alg:llea}
	\KwIn{
		\\
		~~~~$\mathbf{x}_i^\mathrm{U}$: ML pipeline\\
		~~~~$B^\mathrm{L}$: lower-level time budget
	}
	\KwOut{
		\\
		~~~~$\mathbf{x}^\mathrm{L_*}$: optimized hyperparameters\\
		~~~~$f^\mathrm{L}\left(\mathbf{x}_i^\mathrm{U},\mathbf{x}^\mathrm{L_*}\right)$: lower-level function value
		
		}
	
	$\Theta_{\mathbf{x}_i^\mathrm{U}}\leftarrow$ Determine the configuration space based on $\mathbf{x}_i^\mathrm{U}$\\
	$\mathcal{H}\leftarrow$ Sample a set of hyperparameters from a predefined configuration space $\Theta_{\mathbf{x}_i^\mathrm{U}}$ employing space-filling\\
	$\mathbf{f}^\mathrm{L}\leftarrow$ Evaluate their lower-level objective functions\\
	
	\While{The lower-level computational budget is not exhausted}
	{
		$S\leftarrow$ Build a surrogate model with the Tree Parzen using $\mathcal{H}$ and $\mathbf{f}^\mathrm{L}$\\
		$\mathbf{x}^\mathrm{L'}\leftarrow$ Optimal configuration derived from the predicted lower-level function value of the acquisition function over $S$\\
		$f^\mathrm{L}\left(\mathbf{x}_i^\mathrm{U},\mathbf{x}^\mathrm{L'}\right)\leftarrow$ Train the CPDP model physically to evaluate the lower-level objective function\\
		$\mathcal{H}\leftarrow\mathcal{H}\bigcup\mathbf{x}^\mathrm{L'}, \mathbf{f}^\mathrm{L}\leftarrow\mathbf{f}^\mathrm{L}\bigcup f^\mathrm{L}\left(\mathbf{x}_i^\mathrm{U},\mathbf{x}^\mathrm{L'}\right)$
		
	}
	
	\Return $\mathbf{x}^\mathrm{L_*}\leftarrow\underset{\mathbf{x}^\mathrm{L_*}\in\mathcal{H}}{\argmin}~\mathbf{f}^\mathrm{L}, f^\mathrm{L}\left(\mathbf{x}_i^\mathrm{U},\mathbf{x}^\mathrm{L_*}\right)\leftarrow\underset{\mathbf{x}^\mathrm{L_*}\in\mathcal{H}}{\min}~\mathbf{f}^\mathrm{L}$
	
\end{algorithm}

%% file: tables/feature_selections.tex
\begin{table}[htbp]
\scriptsize
  \centering
  \caption{Summary of the feature selection methods (where [$\mathbb{Z}$] denotes integers, [$\mathbb{R}$] real numbers, and [$\mathbb{C}$] categorical variables)}
    \begin{tabular}{C{1.9cm}C{1.5cm}||C{1.3cm}C{1.9cm}}
    \hline
    Parameter & Range & Parameter & Range \\
    \hline
    \multicolumn{2}{c||}{Hybrid Filter (HF)~\cite{LiCWMTTL17}} & \multicolumn{2}{c}{FeSCH~\cite{NiLCGCH17}} \\
    strategy [$\mathbb{C}$] & Var, MI, CHI & nt [$\mathbb{Z}$] & [1, $N_s$] \\
    threshold [$\mathbb{R}$] & $[0.6, 0.9]$ & strategy [$\mathbb{C}$] & SFD, LDF, FCR \\
    \hline
    \multicolumn{2}{c||}{Random Forest Variable}& \multicolumn{2}{c}{LASSO~\cite{MuthukrishnanR16}} \\
    \cline{3-4}
    \multicolumn{2}{c||}{Importance (RFVI)~\cite{GenuerPT10}} & C [$\mathbb{R}$] & $[0.1, 1]$ \\
    n\_est [$\mathbb{Z}$] & $[10, 200]$ & Penalty [$\mathbb{C}$] & L1, L2 \\
    \cline{3-4}
    max\_depth [$\mathbb{Z}$] & $[10, 200]$ & \multicolumn{2}{c}{PCAmining~\cite{NagappanBZ06}} \\
    I\_threshold [$\mathbb{C}$] & mean,  & dim [$\mathbb{Z}$] & $[5, max(N_s, N_t)]$ \\
    \cline{3-4}
    & median & \multicolumn{2}{c}{None} \\
    \hline
    \end{tabular}%
  \label{tab:fs}%
\end{table}%

%% file: tables/transfer_learners.tex
\begin{table}[htbp]
\scriptsize
  \centering
  \caption{Summary of the transfer learning methods (where [$\mathbb{Z}$] denotes integers, [$\mathbb{R}$] real numbers, and [$\mathbb{C}$] categorical variables)}
    \begin{tabular}{C{1.7cm}C{1.7cm}||C{1.5cm}C{1.7cm}}
    \hline
    Parameter & Range & Parameter & Range \\
    \hline
    \multicolumn{2}{c||}{MCWs~\cite{QiuLJ18}} & \multicolumn{2}{c}{FSS\_bagging~\cite{HePMY13}}  \\
    k [$\mathbb{Z}$] & $[2, N_s]$ & topN [$\mathbb{Z}$] & $[1, 15]$ \\
    sigma [$\mathbb{R}$] & $[0.01, 10]$ & threshold [$\mathbb{R}$] & $[0.3, 0.7]$ \\
    lambda [$\mathbb{R}$] & $[10E-7, 100]$ & ratio [$\mathbb{R}$] & $[0.1, 0.5]$ \\
    \hline
    \multicolumn{2}{c||}{TD~\cite{Herbold13}}  & \multicolumn{2}{c}{VCB~\cite{RyuCB16}}  \\
    strategy [$\mathbb{C}$] & NN, EM & m [$\mathbb{Z}$] & $[2, 30]$ \\
    k [$\mathbb{Z}$] & $[1, N_s]$ & lambda [$\mathbb{R}$] & $[0.5, 1.5]$ \\
    \hline
    \multicolumn{2}{c||}{Nearest-neighbor filter (NNfilter)~\cite{TurhanMBS09}} & \multicolumn{2}{c}{CDE\_SMOTE~\cite{LimsetthoBKHM18}} \\
    k [$\mathbb{Z}$] & $[1, 100]$ & k [$\mathbb{Z}$] & $[1, 100]$ \\
    metric [$\mathbb{C}$] & Euc, Man, Che, Min,Mah & metric [$\mathbb{C}$] & Euc, Man, Che, Min, Mah \\
    \hline

    \multicolumn{2}{c||}{UM~\cite{0001MKZ14}}  & \multicolumn{2}{c}{HISNN~\cite{RyuJB15}}  \\
    p [$\mathbb{R}$] & $[0.01, 0.1]$ & minham [$\mathbb{Z}$] & $[1, N_s]$ \\
    \cline{3-4}
    qua\_T [$\mathbb{C}$] & cli, cohen & \multicolumn{2}{c}{None} \\
    \hline
    
    \multicolumn{2}{c||}{GIS~\cite{HosseiniTM18}} & & \\
    prob [$\mathbb{R}$] & $[0.02, 0.1]$ & \multicolumn{2}{c}{CLIFE\_MORPH~\cite{PetersMGZ13}} \\
    chrm\_size [$\mathbb{R}$] & $[0.02, 0.1]$ & n [$\mathbb{Z}$] & $[1, 100]$  \\
    pop\_size [$\mathbb{Z}$] & $[2,30]$ & alpha [$\mathbb{R}$] & $[0.05, 0.2]$ \\
    num\_parts [$\mathbb{Z}$] & $[2,6]$ & beta [$\mathbb{R}$] & $[0.2, 0.4]$ \\
    %\cline{3-4}
    %\cmidrule{3-4}    
    num\_gens [$\mathbb{Z}$] & $[5, 20]$ & per [$\mathbb{R}$] & $[0.6, 0.9]$ \\
    mcount [$\mathbb{Z}$] & $[3, 10]$ &  &  \\
    \hline
    \end{tabular}%
  \label{tab:tl}%
\end{table}%

%% file: tables/classifiers.tex
\begin{table*}[htbp]
\scriptsize
  \centering
  \caption{Summary of the classification methods (where [$\mathbb{Z}$] denotes integers, [$\mathbb{R}$] real numbers, and [$\mathbb{C}$] categorical variables)}
    \begin{tabular}{cc||cc||cc||cc}
    \hline
    Parameter & Range & Parameter & Range & Parameter & Range & Parameter & Range \\
    \hline
    \multicolumn{2}{c||}{Extra Trees Classifier (EXs)} & \multicolumn{2}{c||}{Extra Tree Classifier (Extree)} & \multicolumn{2}{c||}{Decision Tree (DT)} & \multicolumn{2}{c}{Random Forest (RF)} \\
    max\_e [$\mathbb{Z}$] & $[10, 100]$ & max\_e [$\mathbb{Z}$] & $[10, 100]$ & max\_e [$\mathbb{Z}$] & $[10, 100]$ & m\_stim [$\mathbb{Z}$] & $[10, 100]$ \\
    criterion [$\mathbb{C}$] & gini, entropy & criterion [$\mathbb{C}$] & gini, entropy & criterion [$\mathbb{C}$] & gini, entropy & criterion [$\mathbb{C}$] & gini, entropy \\
    min\_s\_l [$\mathbb{Z}$] & $[1,20]$ & min\_s\_l [$\mathbb{Z}$] & $[1, 20]$ & min\_s\_l [$\mathbb{Z}$] & $[1, 20]$ & min\_s\_l [$\mathbb{Z}$] & $[1, 20]$ \\
    Splitter [$\mathbb{C}$] & random, best & splitter [$\mathbb{C}$] & random, best & splitter [$\mathbb{C}$] & auto, sqrt, log2 & splitter [$\mathbb{C}$] & auto, sqrt, log2 \\
    min\_a\_p [$\mathbb{Z}$] & $[2, N_s/10]$ & min\_a\_p [$\mathbb{Z}$] & $[2, N_s/10]$ & min\_a\_p [$\mathbb{Z}$] & $[2, N_s/10]$ & min\_a\_p [$\mathbb{Z}$] & $[2, N_s/10]$ \\
    \hline
    \multicolumn{2}{c||}{Logistic Regression (LR)} & \multicolumn{2}{c||}{Ridge} & \multicolumn{2}{c||}{Bagging} & \multicolumn{2}{c}{Naïve Bayes (NB)} \\
    penalty [$\mathbb{C}$] & L1, L2 & alpha [$\mathbb{R}$] & $[10E-5,1000]$ & n\_est [$\mathbb{Z}$] & $[10, 200]$ & NBType [$\mathbb{C}$] & gauss, multi, comp \\
    fit\_int [$\mathbb{C}$] & true, false & fit\_int [$\mathbb{C}$] & true, false & max\_s [$\mathbb{R}$] & $[0.7, 1.0]$ & alpha [$\mathbb{R}$] & $[0, 10]$ \\
    tol [$\mathbb{R}$] & $[10E-6, 0.1]$ & tol [$\mathbb{R}$] & $[10E-6, 0.1]$ & max\_f [$\mathbb{R}$] & $[0.7, 1.0]$ & norm [$\mathbb{C}$] & true, false \\
    \hline
    \multicolumn{2}{c||}{K-Nearest Neighbor (KNN)} & \multicolumn{2}{c||}{Radius Neighbors Classifier (RNC)} & \multicolumn{2}{c||}{adaBoost} & \multicolumn{2}{c}{Nearest Centroid Classifier (NCC)} \\
    n\_neigh [$\mathbb{Z}$] & $[1, 50]$ & radius [$\mathbb{R}$] & $[0, 10000]$ & n\_est [$\mathbb{Z}$] & $[10, 1000]$ & metric [$\mathbb{C}$] & Euc, Man, Che, Min, Mah \\
    p [$\mathbb{Z}$] & $[1, 5]$ & weight [$\mathbb{C}$] & uni, dist & rate [$\mathbb{R}$] & $[0.01, 10]$ & shrink\_t [$\mathbb{R}$] & $[0, 10]$ \\
    \hline
    \multicolumn{2}{c||}{Support Vector Machine (SVM)} & \multicolumn{2}{c||}{Passive Aggressive Classifier (PAC)} & \multicolumn{2}{c||}{Perceptron} & \multicolumn{2}{c}{Multilayer Perceptron (MLP)} \\
    kernel [$\mathbb{C}$] & rbf, lin, poly, sig & C [$\mathbb{R}$] & $[0.001, 100]$ & penalty [$\mathbb{C}$] & L1, L2 & active [$\mathbb{C}$] & iden, log, tanh, relu \\
    C [$\mathbb{R}$] & $[0.001, 10]$ & fit\_int [$\mathbb{C}$] & true, false & alpha [$\mathbb{R}$] & $[10E-6, 0.1]$ & hid\_l\_s [$\mathbb{Z}$] & $[50, 200]$ \\
    degree [$\mathbb{Z}$] & $[1, 5]$ & tol [$\mathbb{R}$] & $[10E-6, 0.1]$ & fit\_int [$\mathbb{C}$] & true, false & solver [$\mathbb{C}$] & llbfgs, sgd, adam \\
    coef0 [$\mathbb{R}$] & $[0, 10]$ & loss [$\mathbb{C}$] & hinge, s\_hinge & tol [$\mathbb{R}$] & $[10E-6, 0.1]$ & iter [$\mathbb{Z}$] & $[100, 250]$ \\
    gamma [$\mathbb{R}$] & $[0.01, 100]$ &       &       &       &       &       &  \\
    \hline
    \end{tabular}%
  \label{tab:classifiers}%
\end{table*}%

%% file: settings.tex
%!TeX root=main.tex

\section{Experimental Setup}
\label{sec:settings}

The experimental setup of our empirical study is introduced as follows.

\subsection{Dataset}
\label{sec:dataset}

To enhance reproducibility, practicality, and credibility, we consider the following \textit{inclusion} and \textit{exclusion} criteria to constitute the dataset used in our empirical study.
\begin{itemize}
    \item We have three \textit{inclusion} criteria. $\blacktriangleright$ We select the projects hosted in public repositories. This not only makes the information readily available to researchers and the public but also facilitates the verification of our findings. In particular, we prioritize non-academic software to align our study closely with real-world software development practices. $\blacktriangleright$ To generalize our results across various contexts, we incorporate a diverse range of projects across different domains. $\blacktriangleright$ To ensure our findings are robust and credible, we focus on projects well recognized within the CPDP community.

    \item We apply two \textit{exclusion} criteria to refine the dataset and avoid potential biases. $\blacktriangleright$ We adhere to a single-version policy, retaining only the latest version of each project. This is because projects often have multiple versions with software evolution. Therefore, retaining only the latest version mitigates version duplication. $\blacktriangleright$ To maintain data integrity, we exclude repeated projects and those with missing data, because such datasets can compromise the validity.
\end{itemize}

In practice, by applying the \textit{inclusion} criteria, we initially selected five publicly available datasets: \texttt{JURECZKO}, \texttt{NASA}, \texttt{SOFTLAB}, \texttt{AEEEM}, and \texttt{ReLink}, all of which have been extensively reviewed in CPDP surveys~\cite{KhatriS22,HosseiniTG19,HerboldTG18}. Thereafter, by applying the \textit{exclusion} criteria, we remove \texttt{SOFTLAB} and \texttt{NASA}, because of their data quality concerns reported in~\cite{ShepperdSSM13}. All in all, our final datasets comprise $20$ open-source projects with $10,952$ instances. Their key characteristics are summarized as follows.
\begin{itemize}
    \item \texttt{AEEEM}~\cite{DAmbrosLR10}: This dataset includes five open-source projects with $5,371$ instances in total. Each instance is characterized by $61$ metrics, covering static and process metrics, such as the entropy of code changes and source code churn.

    \item \texttt{JURECZKO}~\cite{JureczkoM10}: This dataset originally consists $92$ software releases from various sources. By applying the \textit{inclusion} criteria, we only consider the latest versions of $12$ open-source projects therein, with $4,932$ instances in total.
  
    \item \texttt{ReLink}~\cite{WuZKC11}: This dataset includes three open-source projects with $649$ instances. Specifically, each instance is characterized by $26$ static metrics. Note that the defect labels are manually verified after they are generated from the source code management system commit comments.
\end{itemize}

\subsection{Peer Algorithms}
\label{sec:peer_algorithms}

To validate the effectiveness our proposed \our, we designed the following \textit{inclusion} and \textit{exclusion} criteria to select our peer algorithms for comparison.
\begin{itemize}
    \item Our \textit{inclusion} criterion is based on a search for open-source projects related to AutoML in GitHub. In particular, following the standards recommended in~\cite{CalefatoQLK23}, we use the keyword combination of \lq\texttt{AutoML}\rq\ and \lq\texttt{AutoAI}\rq. Finally, we obtained $4,700$ relevant projects.

    \item We consider five \textit{exclusion} criteria to refine our selection. $\blacktriangleright$ We prioritize the projects developed in Python, given the widespread use of Python and its strong community support. This narrows the number of projects down to $\approx 1,500$. $\blacktriangleright$ To exclude small projects that might lack stability and community support, we focus on projects with more than $200$ stars and being actively updated with the last year. This further narrows the number of projects down to $\approx 71$. We also find a project that lacks a relevant README document, and we remove it from further consideration. $\blacktriangleright$ To ensure relevance to our CPDP context, we exclude the projects that focus only on computer vision and those developed for pedagogic or competition purposes. This narrows the number of projects down to $\approx 24$. $\blacktriangleright$ We also exclude those small projects under large commercial solutions or repositories because they actually do not contain AutoML capabilities.
\end{itemize}

Based on the above criteria, we finally have $14$ projects at hand. After a thorough review of their documentations, we finally choose five AutoML tools for comparisons: \texttt{AutoKeras}~\cite{JinSH19}, \texttt{MLjar}~\cite{mljar}, \texttt{SapientML}~\cite{SahaUM0LHYKP22}, \texttt{Auto-sklearn}~\cite{FeurerKESBH15}, and \texttt{Bilo-CPDP}~\cite{LiXCT20}.

\subsection{Parameter Settings}
\label{sec:parameters}

Considering the high computational cost of bilevel optimization, we set the population size of the upper-level optimization to $10$ to maintain a reasonable search speed and response time under limited computing resources. As in~\cite{LiDZK15}, we set the crossover probability $p_c=1.0$ and the distribution index $\eta_c=30$ for the SBX crossover, while the mutation probability $p_m=\frac{1}{n}$ and the distribution index $\eta_m=20$ for the polynomial mutation, where $n$ is the number of decision variables. As for the ensemble method, we set the ensemble selection size $N^\mathrm{s}=3$, and accordingly, and the ensemble pool size $N^\mathrm{e}=2N^\mathrm{s}$, following the principle of \lq\textit{the law of diminishing returns in ensemble construction}\rq~\cite{BonabC19}.

In our experiments, we follow the protocol used in~\cite{FeurerKESBH15} and~\cite{LiXCT20} that restrict the computational budget to $3,600$ seconds in total. Taking account of ensemble learning, we allocate $3200$ seconds for phase $1$ while the remaining $400$ seconds are allocated for phase $2$. Each lower-level search is allocated with $20$ seconds. Due to the stochastic nature of the optimization algorithms, all experiments are independently repeated $31$ times with different random seeds.

\subsection{Performance Evaluation}
\label{sec:performance_evaluation}

\subsubsection{Performance metrics}
\label{sec:metrics}

We use the following five metrics to evaluate the performance of different methods quantitatively.
\begin{itemize}
    \item \textbf{AUC}~\cite{ZhouYLCLZQX18}: It can measure the ability of the classifier under different threshold settings. $\mathrm{AUC}$ is not affected by the selection of thresholds, particularly suitable for handling imbalanced datasets.

    \item \textbf{Accuracy (ACC)}~\cite{Metz78}: $\mathrm{ACC}$ reflects the overall accuracy of a classifier:
        \begin{equation}
            \mathrm{ACC}=\frac{\mathrm{TP}+\mathrm{TN}}{\mathrm{TP}+\mathrm{FP}+\mathrm{FN}+\mathrm{TN}},
            \label{eq:acc}
        \end{equation}
        where $\mathrm{TP}$ and $\mathrm{FP}$ indicate the number of times a benign instance is correctly or incorrectly identified as defective, respectively. In contrast, $\mathrm{FN}$ and $\mathrm{TN}$ refer to the number of times a defective instance is wrongly or correctly identified as benign, respectively. Note that $\mathrm{ACC}$ is suitable for balanced class distributions.

    \item \textbf{Recall}~\cite{HanK2000}: $\mathrm{Recall}$ is used to assess the ability of the classifier to identify positive samples:
        \begin{equation}
            \mathrm{Recall}=\frac{\mathrm{TP}}{\mathrm{TP}+\mathrm{FN}}.
        \end{equation}
        Note that $\mathrm{Recall}$ is an important metric when considering minority (positive) classifications, where accurately detecting rare or infrequent events is critical.

    \item \textbf{F1-Score (F1)}~\cite{HanK2000}: $\mathrm{F1}$ is the harmonic mean of $\mathrm{Precision}$ and $\mathrm{Recall}$:
        \begin{equation}
            \mathrm{F1}=2\times\frac{\mathrm{Precision}\times\mathrm{Recall}}{\mathrm{Precision}+\mathrm{Recall}},
        \end{equation}

    where $\mathrm{Precision}$ measures the accuracy of positive predictions and is defined as:
        \begin{equation}
            \mathrm{Precision}=\frac{\mathrm{TP}}{\mathrm{TP}+\mathrm{FP}}.
        \end{equation}

    \item \textbf{Matthews Correlation Coefficient (MCC)}~\cite{BaldiBCAN00}: $\mathrm{MCC}$ is a holistic metric that accounts for all four classification results defined in~\pref{eq:acc}, i.e., $\mathrm{TP}$, $\mathrm{FP}$, $\mathrm{TN}$, and $\mathrm{FN}$):
        \begin{equation}
            \mathrm{MCC}=\frac{\mathrm{TP}\times\mathrm{TN}-\mathrm{FP}\times\mathrm{FN}}{\sqrt{(\mathrm{TP}+\mathrm{FP})(\mathrm{TP}+\mathrm{FN})(\mathrm{TN}+\mathrm{FP})(\mathrm{TN}+\mathrm{FN})}}.
        \end{equation}
\end{itemize}
The value ranges of $\mathrm{AUC}$, $\mathrm{F1}$, $\mathrm{ACC}$, and $\mathrm{Recall}$ are all within $[0,1]$, while the value of $\mathrm{MCC}$ is within the range between $-1$ and $1$. A larger value of these metrics indicates a better performance. 

\subsubsection{Statistical tests}
\label{sec:statistics}

Due to the stochastic nature of the algorithms considered in our empirical study, we use the following three statistical tests to interpret the statistical significance of our comparison results.
\begin{itemize}
	\item \textbf{Wilcoxon rank sum test}~\cite{Haynes2013}: This is a non-parametric test that assesses differences between two datasets without relying on a normal distribution assumption. It is commonly used with a significance level of $p=0.05$~\cite{ArcuriB11} to determine if there is a statistically significant median difference between the results.
	
	\item \textbf{Scott-Knott test}~\cite{MittasA13}: This is a clustering-based method that accounts for mean rankings and variance analysis. Its basic idea is to partition data into homogeneous subsets with low intergroup variance and high intergroup variance. This test can identify statistically significant differences in performance.

	\item \textbf{$A_{12}$ effect size}~\cite{VarghaD00}: To assess the significance of differences between two independent samples, we use $A_{12}$ effect size to rank results. Specifically, $A_{12} = 0.5$ indicates equivalent performance between groups; $A_{12} > 0.5$ suggests the first group outperforms the second more than $50$\% of the comparisons; $0.56 \leq A_{12} < 0.64$ signifies a small effect size, $0.64 \leq A_{12} <0.71$ indicates a medium effect size and $A_{12} \geq 0.71$ denotes a large effect size, thus identifying differences of statistical significance.
	
\end{itemize}

%% file: results.tex
\section{Results and Discussions}
\label{sec:results}

In this section, we evaluate and analyze the effectiveness of \our~through answering the following research questions (RQs).

\begin{itemize}
    \item \underline{\textbf{RQ1:}} How is the performance of \our~compared with some state-of-the-art AutoML tools?
       
    \item \underline{\textbf{RQ2:}} Could MBL-CPDP achieve better performance than the existing CPDP techniques?

    \item \underline{\textbf{RQ3:}} What are the benefits of MBLO framework?

    \item \underline{\textbf{RQ4:}} What is the impact of various techniques on the overall performance?
\end{itemize}

\subsection{Comparison with the AutoML Tools}
\label{sec:res_automl}

To address \textbf{RQ1}, we conduct a statistical analysis to compare the performance of \our~against other AutoML tools across various datasets.

The statistical comparison results of five performance metrics, based on the Wilcoxon rank sum test, among six AutoML tools are given in \pref{tab:automl_table}. These metrics provide a comprehensive evaluation of the performance of each tool. The Wilcoxon rank sum test results reveal that \our~significantly outperforms the other five AutoML in performance metrics, indicating its superior performance in addressing CPDP tasks.
 
\input{tables/automl_table}

\begin{figure*}[t!]
	\centering
	\includegraphics[width=\linewidth]{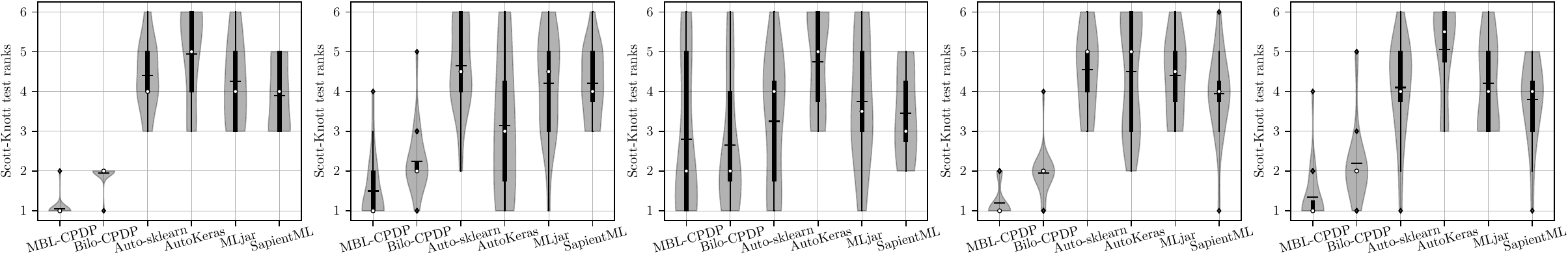}
	\caption{Violin plots and box plots of Scott-Knott test ranks achieved by six AutoML tools on AUC, ACC, Recall, F1, and MCC.}
	\label{fig:AutoML-All-ScottKnott}
\end{figure*}

\begin{figure}[t!]
	\centering
	\includegraphics[width=0.5\linewidth]{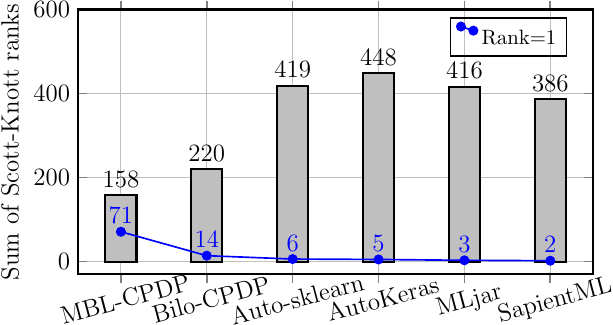}
	\caption{Total Scott-Knott test ranks achieved by six AutoML tools (A smaller sum of ranks in the bar chart indicates superior performance. The line graph shows the frequency of top ranks, with higher values denoting better performance.).}
	\label{fig:AutoML-Total-ScottKnott}
\end{figure}

To facilitate a better ranking among these AutoML tools, we apply the Scott-Knott test to classify them into different groups according to their performance on each dataset. Due to the large number of comparison results, it will be messy if we present all ranking results ($20\times5\times6=600$ in total) obtained by the Scott-Knott test collectively. Instead, to have a better interpretation of the comparison among different datasets, we divide the Scott-Knott test results according to different metrics, pull the test results of the same metric together, and show their distribution and variance as violin plots and box plots in \pref{fig:AutoML-All-ScottKnott}. In addition, to facilitate an overall comparison, we further summarize the Scott-Knott test results obtained across all datasets and performance metrics and show them as the bar charts in~\pref{fig:AutoML-Total-ScottKnott}. The Scott-Knott test ranks the AutoML tools based on their mean performance scores. Since each metric evaluates the performance according to different standards, there are slight differences in the performance evaluation. However, in most cases, \our~has shown excellent performance in all metrics and is ranked first place with regard to the other five AutoML tools. Based on the Scott-Knott test, we can conclude that \our~outperforms all other AutoML tools, as it is ranked highest and placed in a distinct group.

\begin{figure*}[t!]
	\centering
	\includegraphics[width=\linewidth]{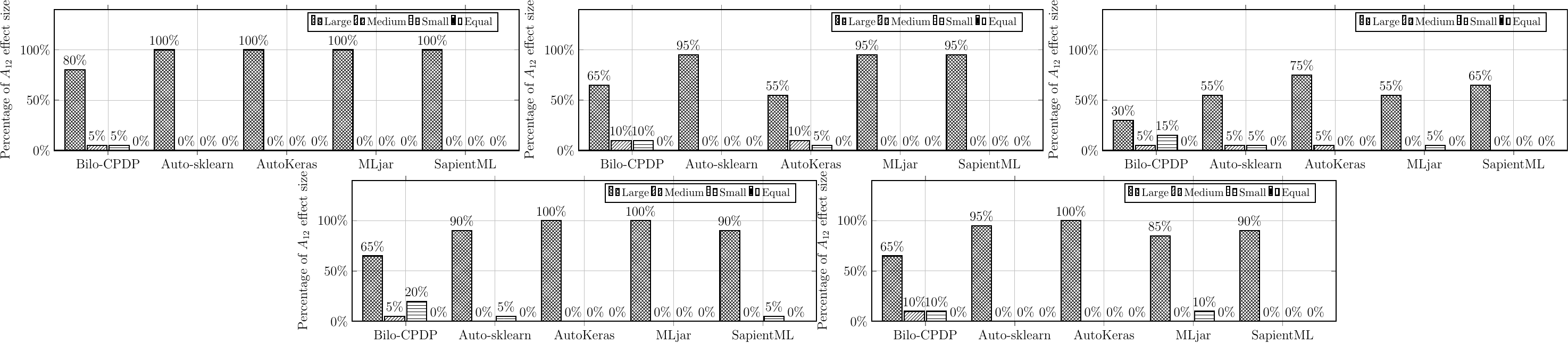}
	\caption{Percentage of the large, medium, small, and equal $A_{12}$ effect size, respectively, when comparing \our~with other five AutoML tools on AUC, ACC, Recall, F1, and MCC.}
	\label{fig:AutoML-All-A12}
\end{figure*}

\begin{figure}[t!]
	\centering
	\includegraphics[width=\linewidth]{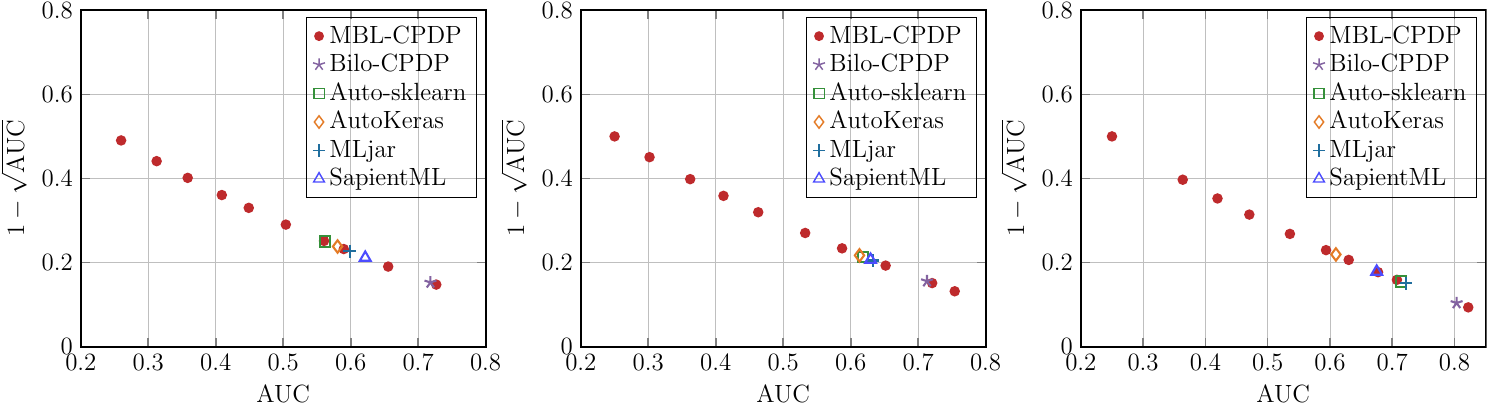}
	\caption{The final solutions obtained by the six AutoML tools for three projects: EQ, LC, and Safe, respectively.}
	\label{fig:PF}
\end{figure}

Furthermore, we use \our~to compare performance differences with other AutoML tools, employing the $A_{12}$ effect size. Since the $A_{12}$ effect size calculation is conducted pairwise, there are $20\times5\times5=500$ piecemeal $A_{12}$ comparison results. We again pull the results together and categorize them by different metrics. We calculate the percentages of equivalent, small, medium, and large effect sizes for each of the other five tools. From the statistical results shown in~\pref{fig:AutoML-All-A12}, \our~has shown dominantly better results compared to \texttt{Auto-sklearn}, \texttt{MLjar}, \texttt{SapientML} on AUC, F1, Recall and MCC where the large effect sizes are all over 90\%. In contrast, the effect sizes with regard to \texttt{Bilo-CPDP} and \texttt{AutoKeras} are relatively comparable. Additionally, it is worth noting that the performance of different methods may vary due to the different evaluation criteria of various metrics for different aspects. Therefore, relying on a single evaluation metric is insufficient to evaluate the performance of an AutoML tool comprehensively. Finally, to have a better visual presentation of the solutions, the final solutions obtained by the six AutoML tools in EQ, LC, and Safe are presented in \pref{fig:PF}. From these plots, it is clear to see that \our~can obtain a set of diverse solutions with good performance.

\vspace{1em}
\noindent
\framebox{\parbox{\dimexpr\linewidth-2\fboxsep-2\fboxrule}{
		\textbf{\underline{Response to RQ1:}} \textit{From the empirical results discussed in this subsection, we confirm the effectiveness of \our. Specifically, when addressing the CPDP tasks involved in this paper, \our~exhibits the best performance, followed by \texttt{Bilo-CPDP} in second place, with \texttt{MLjar} and \texttt{Auto-sklearn} next in the ranking. \texttt{AutoKeras} shows the worst performance; however, it demonstrates moderate performance regarding Recall. In addition, it is recommended to use different metrics to evaluate the performance of the AutoML tool.}}}

\subsection{Comparison with CPDP Techniques}
\label{sec:res_exi_cpdp}
To answer \textbf{RQ2}, we conduct a comprehensive experiment using all the feature selection techniques, transfer learning techniques, and classifiers listed in \pref{tab:fs} to \pref{tab:classifiers}, forming a total of $1056$ CPDP techniques for comparison. It is worth noting that since feature selection techniques and transfer learning techniques can be set to None, these CPDP techniques include those composed only of feature selection techniques and classifiers, as well as those composed of transfer learning techniques and classifiers. Some of these techniques have been studied and applied in the literature. For hyperparameter settings, we adopt the same HPO method in this paper. Due to the numerous CPDP techniques, it is impractical to list all the results. Therefore, we only discuss the results of the ML pipelines that showed the best performance in the $31\times20=620$ experiments. 

\begin{figure*}[t!]
	\centering
	\includegraphics[width=\linewidth]{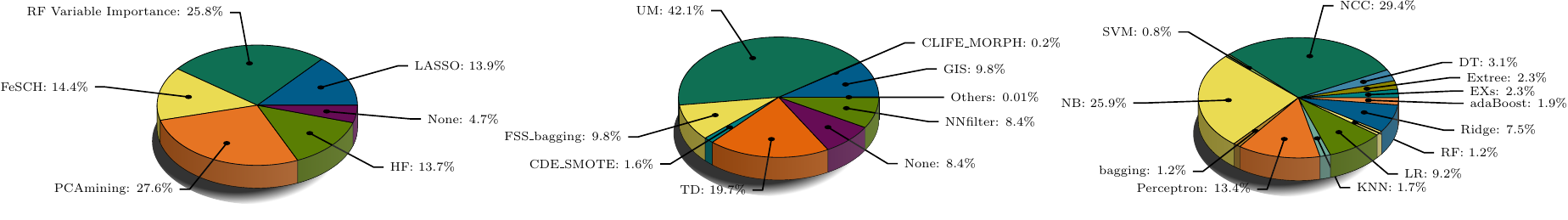}
	\caption{Percentage of the best-performing feature selection techniques, transfer learning techniques, and classifiers (a higher ratio indicates better overall performance).}
	\label{fig:Pie}
\end{figure*}

To better visualize the experimental results, we combine the pie chart to show the occurrence proportion of the feature selection techniques, transfer learning techniques, and classifiers with the best performance in $20$ projects. As shown in~\pref{fig:Pie}, the results indicate a relatively even distribution among the transfer learning methods. Principal Component Analysis mining (PCAmining) demonstrates the best performance, followed by Random Forest (RF). Universal Model (UM) shows a dominant advantage in transfer learning techniques with a $42.1\%$ ratio, followed by Training Data Selection (TD) at $19.7\%$. Regarding classifiers, Nearest Centroid Classifier (NCC) and Naïve Bayes (NB) rank in the first tier with ratios of $29.4\%$ and $25.9\%$, respectively.

\begin{figure}[t!]
	\centering
	\includegraphics[width=0.5\linewidth]{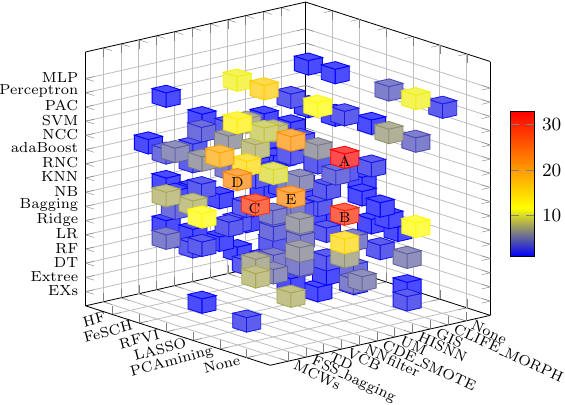}
	\caption{3D heat map of ML pipelines, where red indicates the ML pipeline with the best performance.}
	\label{fig:heatmap}
\end{figure}

Next, we investigate the effects of ML pipelines, and the number of occurrences of each pipeline across $620$ experimental results is shown in~\pref{fig:heatmap} in the form of a heat map. The color of the heat map transitions from blue to red, with colors closer to red indicating superior performance. Five points are marked in~\pref{fig:heatmap}, from A to E, representing the ML pipelines with the largest occurrences, among which A has the largest occurrences. These points correspond to the pipelines \{PCAmining, UM, NCC\}, \{PCAmining, UM, NB\}, \{RF Variable Importance (RFVI), Nearest-neighbor filter (NNfilter), NB\}, \{Hybrid Filter (HF), UM, NB\}, and \{RFVI, UM, NB\}, respectively. From these results, we conclude that PCAmining and RFVI are effective and adaptable in feature selection methods. UM exhibits the best performance among transfer learning methods, while NCC shows the best efficacy in classification methods. However, NB demonstrates better adaptability, allowing it to pair well with various ML pipelines.

\begin{figure*}[t!]
	\centering
    \includegraphics[width=0.9\linewidth,height=0.7\columnwidth]{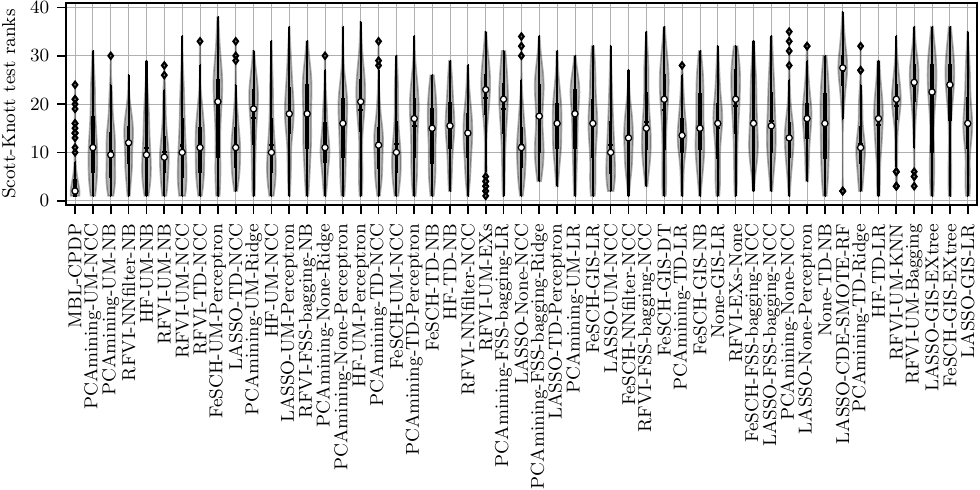} %{figs/CPDP-techniques-ScottKnott-All.pdf}
	\caption{Scott-Knott test ranks between MBL-CPDP (the leftmost one) and the $50$ peer CPDP techniques (the smaller test rank, the better).}
	\label{fig:CPDP-techniques-ScottKnott-AUC}
\end{figure*}

\begin{figure*}[t!]
	\centering
	\includegraphics[width=0.9\linewidth,height=0.7\columnwidth]{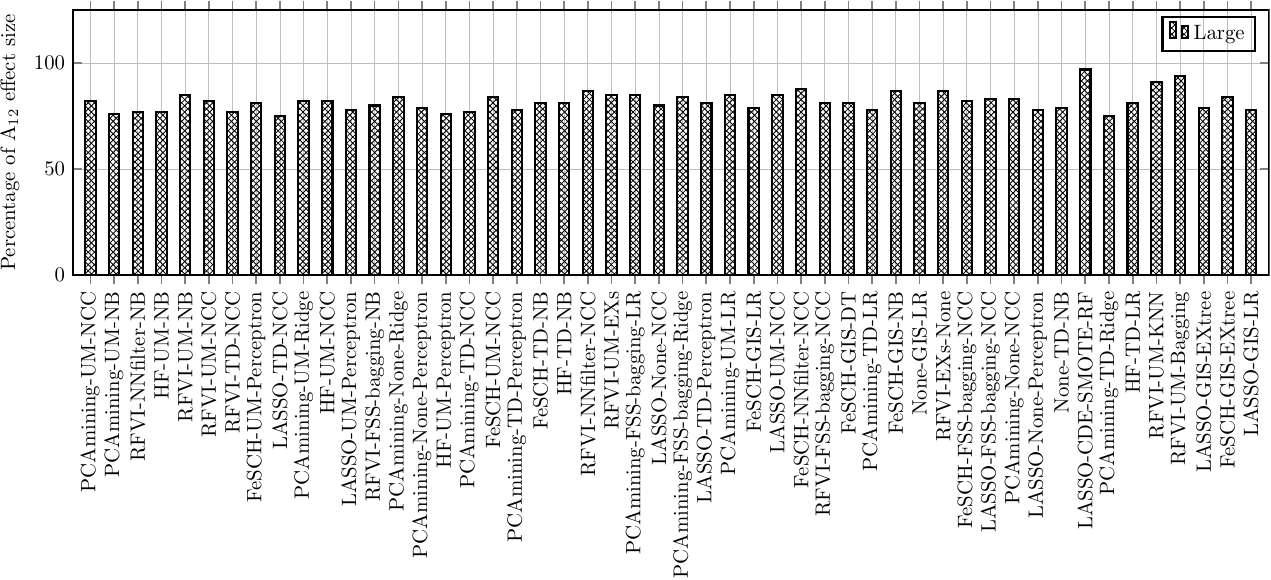}
	\caption{Large $A_{12}$ effect size comparisons between MBL-CPDP and the $50$ peer CPDP techniques over $31$ runs ($A_{12} > 50\%$ suggests MBL-CPDP outperforms the compared techniques, while $A_{12} > 71\%$ denotes a large effect size).}
	\label{fig:CPDP-techniques-A12}
\end{figure*}

Furthermore, to evaluate the performance of \our~against the top $50$ most frequently occurring CPDP techniques, comprehensive comparisons are conducted. Due to page limits, we present the Scott-Knott results and the large $A_{12}$ effect size results of for five metrics in \pref{fig:CPDP-techniques-ScottKnott-AUC} and \pref{fig:CPDP-techniques-A12}, respectively. The experimental results indicate that \our~demonstrates overwhelmingly superior performance compared to the other $50$ CPDP techniques. The combination algorithms denoted from points A to E exhibit similar performance, with RFVI-UM-NB ranking second with a slight advantage. 

\vspace{1em}
\noindent
\framebox{\parbox{\dimexpr\linewidth-2\fboxsep-2\fboxrule}{
		\textbf{\underline{Response to RQ2:}} \textit{Among these CPDP techniques, PCAmining and RFVI are identified as the most effective feature selection techniques, while UM and NCC stand out as the best-performing transfer learning technique and classifier, respectively. Notably, the ML pipeline of RFVI, UM, and NB exhibits the highest performance. Furthermore, \our~demonstrates significantly superior performance compared to $50$ CPDP techniques.}}}

\subsection{Investigation of the Benefit of Multi-Objective Bilevel Optimization Framework}
\label{sec:benefit_mbo}

To explore the potential advantages of multi-objective optimization, we develop a variant that maintains the bilevel optimization structure but converts the upper-level MOP into a single-objective format, which we refer to as \texttt{Single-CPDP}. Additionally, to ensure fairness in our study, we remove the ensemble learning method in another variant, referred to as \texttt{MBL-CPDP-WE}. Below, we analyze the experimental results of these two methods from the Scott-Knott test and the $A_{12}$ effect size.   

%three statistical perspectives: the Wilcoxon rank sum test, the Scott-Knott test, and the $A_{12}$ effect size.   
%\pref{tab:mul_single}
%\input{tables/mul_single_table}

\begin{figure}[t!]
	\centering
	\includegraphics[width=0.9\linewidth]{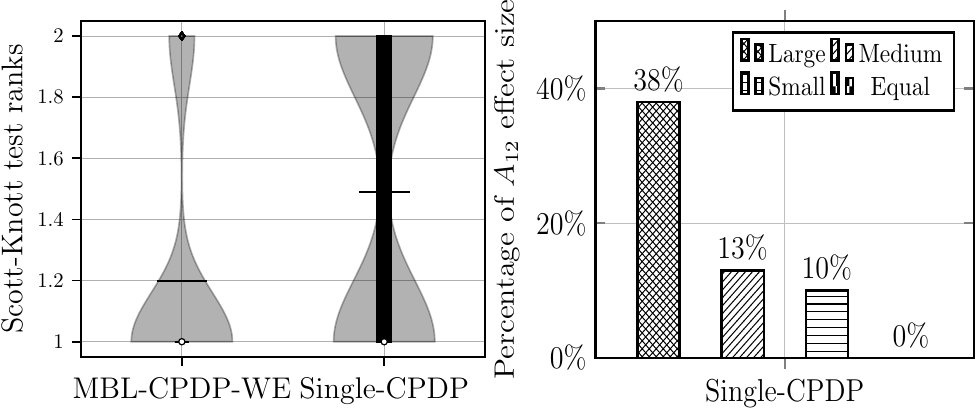}
	\caption{Scott-Knott test ranks and $A_{12}$ effect size comparisons between multi-objective bilevel framework and its single-objective bilevel framework variant.}
	\label{fig:Mul}
\end{figure}

Given the large volume of comparison results, with $200$ evaluations from the Scott-Knott test and another $100$ from the $A_{12}$ effect size ($20$ methods compared across five metrics each), listing all outcomes individually would lead to a cluttered presentation. To streamline interpretation and enhance clarity, we organize these results into single visual representations for each test. The results are shown in \pref{fig:Mul}. Violin plots illustrate the distribution of rankings from the Scott-Knott test, while bar charts depict the $A_{12}$ effect sizes. From the result of the Scott-Knott test, we can see that \texttt{MBL-CPDP-WE} consistently places in a higher performance category, signifying its superior compared to \texttt{Single-CPDP}, which categorizes into a lower performance tier. Furthermore, the $A_{12}$ effect size further confirms the high adaptability and flexibility of the multi-objective form.

\begin{figure*}[t!]
	\centering
	\includegraphics[width=0.9\linewidth]{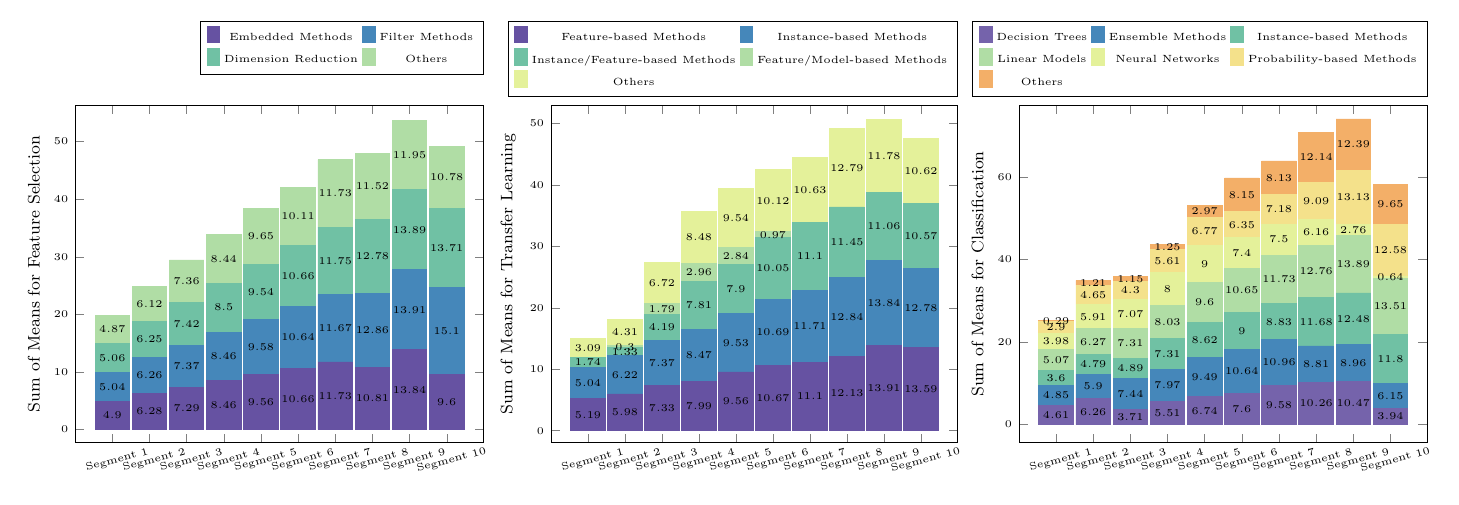}
	\caption{Stacked bar chart of the sum of AUC means for different types of feature selection, transfer learning, and classification methods across $20$ projects (a bigger value indicates better overall performance).}
	\label{fig:stacked_bar}
\end{figure*}

To analyze the distribution characteristics of different types of methods within multi-objective populations, we divide the learners in \pref{tab:fs} to \pref{tab:classifiers} into different categories. Then we divide the obtained approximate Pareto fronts based on upper-level population size into ten equal segments, from Segment $1$ to Segment $10$ based on AUC values. The sum of mean AUC values across $20$ projects is calculated for each method category in each segment. Each project is repeated $31$ times, encompassing feature selection, transfer learning, and classification components, generating $18,600$ results. For clear data presentation, three stacked bar charts are used to display the overall performance of these components. As shown in \pref{fig:stacked_bar}, filter methods in feature selection generally outperform dimension reduction and embedded methods. In transfer learning, feature-based methods consistently excel across all segments, particularly Segment $10$. Linear models achieve the best results for classification, followed by probability-based and instance-based methods, with neural networks having the worst performance. Overall, filter methods of feature selection, feature-based transfer learning, and linear model classification methods exhibit the best performance in their respective categories.

\begin{figure}[t!]
	\centering
	\includegraphics[width=\linewidth]{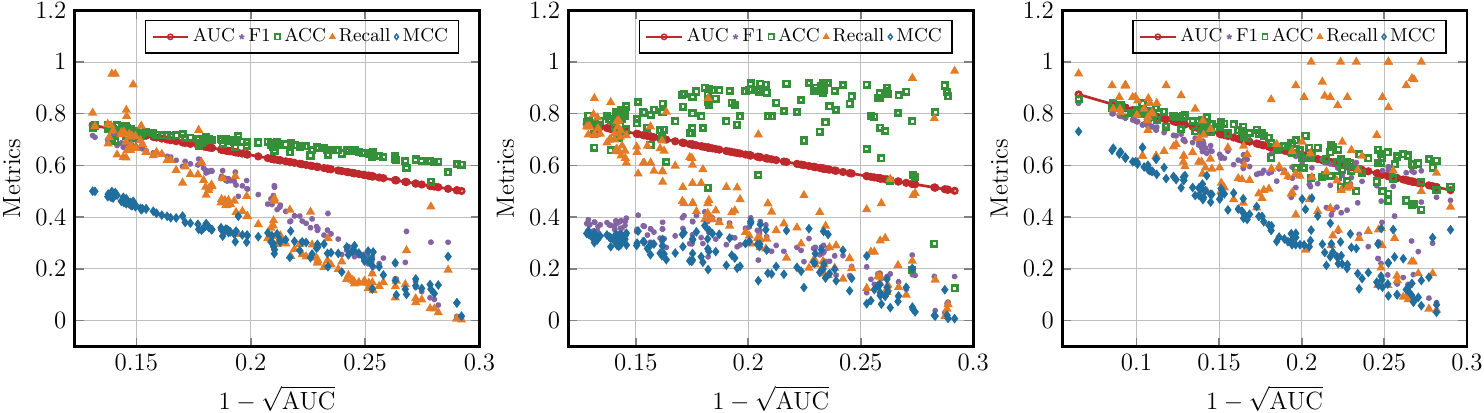}
	\caption{Performance evaluation of MBL-CPDP across multiple metrics on EQ, LC, and Safe projects.}
	\label{fig:FiveMetrics-AllPopulation}
\end{figure}

In the previous section, we conclude from \pref{fig:PF} that using AUC as an evaluation metric, the solution set obtained by the \our~tool outperforms those from other tools. To further validate the advantages of a multi-objective optimization approach, \pref{fig:FiveMetrics-AllPopulation} displays the evaluation of the \our~tool on the EQ, LC, and Safe projects using five different metrics. For ease of observation of the sizes and distribution of the metrics, the x-axis represents $1-\sqrt{\text{AUC}}$, and the y-axis represents the values of the five metrics. Given that metrics such as AUC, F1, ACC, Recall, and MCC are all better when higher, we particularly focus on whether solutions that perform well in AUC also show consistency in other metrics. The results indicate that this consistency does not always hold. Analyzing data from $20$ projects, we observe that while F1 and MCC trends resemble those of AUC, this is not always the case with ACC and Recall. For example, in the EQ dataset, other metrics generally also show an upward trend as AUC increases. However, in the LC dataset, a decrease in AUC is accompanied by an increase in ACC; in the Safe dataset, a decrease in AUC leads to fluctuations in Recall. These findings suggest that different solutions may perform variably across different metrics, thereby emphasizing the importance of exploring diverse solutions within a multi-objective optimization framework. Relying on a single metric in practical applications may obscure other important performance dimensions, limiting the scope and comprehensiveness of model applications. Our research highlights the significance of multi-objective optimization when dealing with complex datasets, particularly its capability to provide a comprehensive evaluation across multiple metrics. This approach identifies solutions that excel in specific metrics and reveals potential trade-offs between different performance indicators, offering decision-makers insights for balancing multiple objectives.

\begin{figure}[t!]
	\centering
	\includegraphics[width=0.7\linewidth]{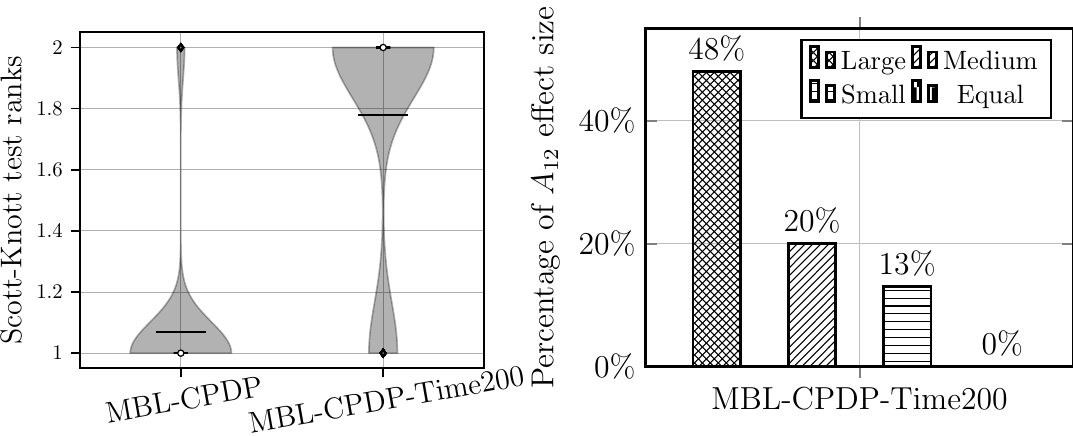}
	\caption{Scott-Knott test ranks and $A_{12}$ effect size comparisons between MBL-CPDP and its variant concentrating more computing resources on the lower-level optimization.}
	\label{fig:Time200}
\end{figure}

Finally, the unique bilevel programming formulated in \our~allows for flexible control over the time budget allocated to the two levels. Therefore, it is interesting to investigate which level is more important. Given that time budget allocations for two levels serve as a measure for computing resources and can thereby influence performance. In this study, we explore an alternative budget allocation strategy within the same total time budget. Specifically, we allocate more time budget to the lower-level HPO, setting the lower-level time budget to 200, denoted as \texttt{MBL-CPDP-Time200}.
Consequently, the upper-level time budget is reduced. The corresponding Scott-Knott test ranks and  $A_{12}$ effect size comparisons between our approach and its variant \texttt{MBL-CPDP-Time200} are presented in \pref{fig:Time200}. Results indicate that our approach outperforms \texttt{MBL-CPDP-Time200}, suggesting that allocating more time budget to the lower-level optimization may degrade performance within the same total time budget. The performance discrepancy may be attributed to the time-consuming property and inefficiency of the CPDP model training process. Once a combination of feature selection technique, transfer learning technique, and classifier is selected at the upper-level routine, there is a tendency for exhaustive exploration of hyperparameters at the lower-level routine, resulting in significant budget consumption. In contrast, constraining the budget at the lower level while ensuring sufficient exploration of the combinatorial space is crucial for optimizing performance. Thus, prioritizing effective exploration within the allocated budget takes priority over exhaustive hyperparameter search under resource limitations. Thus, prioritizing effective exploration within the allocated budget takes priority over exhaustive hyperparameter search under resource limitations.

\vspace{1em}
\noindent
\framebox{\parbox{\dimexpr\linewidth-2\fboxsep-2\fboxrule}{
		\textbf{\underline{Response to RQ3:}} \textit{The multi-objective bilevel framework is essential in our proposed tool. It demonstrates high adaptability and flexibility by easily accommodating existing and new optimization objectives. In this tool, filter-based feature selection, feature-based transfer learning, and linear model classification methods show the best performance in their respective categories and are recommended for priority consideration in similar applications. Furthermore, given a limited budget, prioritizing resource allocation to the upper-level optimization facilitates the exploration of ML pipelines without fully optimizing hyperparameters, which may enhance overall performance.}}}

\subsection{Ablation Studies}
\label{sec:ablation}

\subsubsection{Investigation of the Effect of Ensemble Learning}
\label{sec:effect_el}

The empirical results above clearly demonstrate that our proposed AutoML tool outperforms current state-of-the-art AutoML tools and CPDP techniques. Referring to \pref{fig:MBL-CPDP}, ensemble learning is a crucial part in \our. To answer RQ4, we compare \our~with \texttt{MBL-CPDP-WE} to investigate the effect of ensemble learning.

\begin{figure}[t!]
	\centering
	\includegraphics[width=0.7\linewidth]{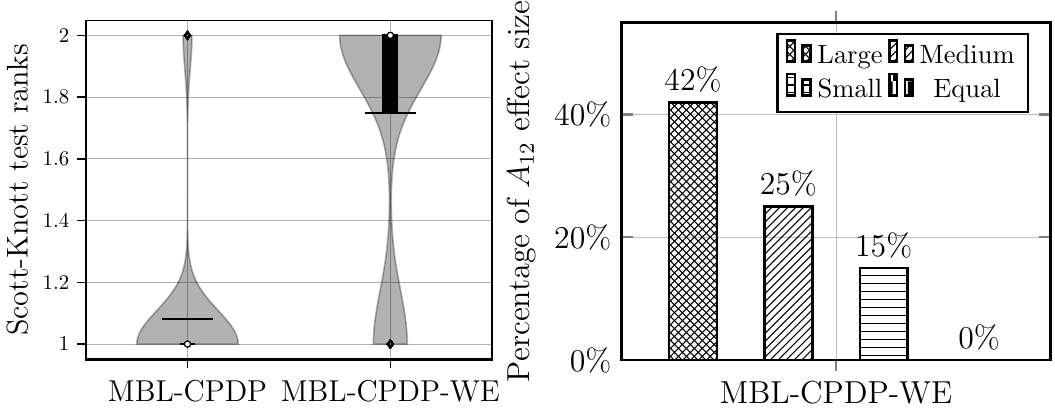}
	\caption{Scott-Knott test ranks and $A_{12}$ effect size comparisons between MBL-CPDP and its variant without using ensemble learning.}
	\label{fig:Ensemble}
\end{figure}

In the same manner as before, we consolidate the results of $200$ Scott-Knott tests and $100$ $A_{12}$ effect size comparisons into a single violin plot and a bar chart, respectively. These visualizations collectively provide a comprehensive overview of the comparative performance across different methods. As shown in \pref{fig:Ensemble}, the analysis clearly segments \our~and \texttt{MBL-CPDP-WE} into distinct clusters, with \our~consistently appearing in the top cluster, indicating its superior performance. Additionally, the $A_{12}$ statistics confirm a large effect, quantified at $42\%$, when comparing two tools, suggesting that removing the ensemble learning method substantially degrades the performance of the tool. Overall, significant differences categorized into large, medium, and small effects account for $82\%$ of the comparisons.

\subsubsection{Investigation of the Effect of The Key Components}
\label{sec:effect_three}

This part further analyzes the impact of different techniques. As illustrated in \pref{fig:MBL-CPDP}, feature selection, transfer learning, and classification are crucial components of our approach. To address RQ4, we conduct an ablation study to evaluate the impact of these components. Specifically, we compare the performance of \texttt{MBL-CPDP-WE} with its variants: one without feature selection (dubbed \texttt{WEF}), one without transfer learning (dubbed \texttt{WET}), and one without feature selection and transfer learning (dubbed \texttt{WEFT}).

\begin{figure}[t!]
	\centering
	\includegraphics[width=0.7\linewidth]{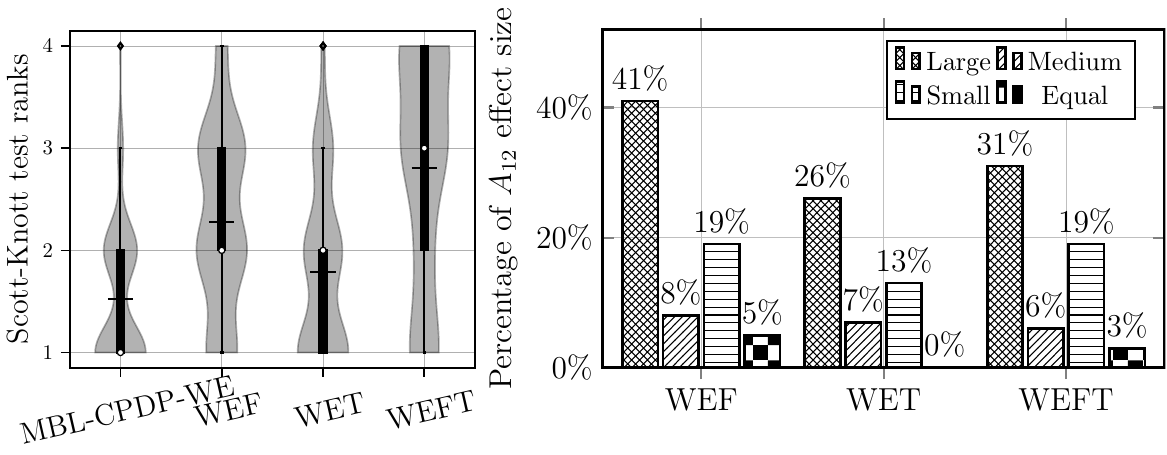}
	\caption{Scott-Knott test ranks and $A_{12}$ effect size comparisons between MBL-CPDP-WE and its variants, one without feature selection and the other without transfer learning.}
	\label{fig:FS_DS}
\end{figure}

Following the same approach as previously, we aggregate the results of Scott-Knott tests ($20\times5\times3$ in total) and $A_{12}$ effect size comparisons ($20\times5\times2$ in total) into a single violin plot and bar chart, respectively. These visualizations are displayed in \pref{fig:FS_DS}. From these results, we can see that \texttt{MBL-CPDP-WE} has shown to be consistently better than the variant without feature selection and the variant without transfer learning. In particular, we see that the performance of \texttt{MBL-CPDP-WE} has been classified into the best group in comparison to the violin plot. The $A_{12}$ effect size indicates that \texttt{MBL-CPDP-WE} also achieves notable performance improvements. 

Furthermore, it is clear from \pref{fig:FS_DS} that \texttt{WET} ranks second, \texttt{WEF} ranks third, while \texttt{WEFT} performs the worst. These results indicate that the removal of both transfer learning and feature selection methods leads to a decrease in performance, thereby substantiating the significant role of transfer learning in enhancing the generalization ability of CPDP, while feature selection contributes to improving adaptability. Particularly, the performance ranking reveals that the variant without feature selection (\texttt{WEF}) performs worse than the variant without transfer learning (\texttt{WET}), and the variant without feature selection and transfer learning exhibits the poorest performance. Therefore, we can conclude that in our experiments, transfer learning and feature selection are crucial for CPDP, yet the impact of feature selection slightly exceeds that of transfer learning.

\vspace{1em}
\noindent
\framebox{\parbox{\dimexpr\linewidth-2\fboxsep-2\fboxrule}{
		\textbf{\underline{Response to RQ4:}} \textit{We have the following takeaways from our experiments. 1) Ensemble learning has been shown to influence overall performance, significantly enhancing generalization performance. 2) Both feature selection and transfer learning play crucial roles in improving the performance of the model. However, feature selection demonstrates a more significant effect than transfer learning.}}}

%% file: tables/automl_table.tex
\newcommand{\customsize}{\fontsize{6.1}{6.2}\selectfont}

\begin{table*}[htbp]
	%\scriptsize
	%\tiny
	\customsize
	\centering
	\caption{Median and interquartile range performance of six AutoML tools over $31$ runs across five metrics (gray=better)}
	\begin{tabular}{cccccccc}
		\toprule
		&       & MBL-CPDP & Bilo-CPDP & Auto-sklearn & AutoKeras & MLjar & SapientML \\
		\midrule
		\multirow{5}[2]{*}{EQ} & AUC   & \bb{\textbf{7.3011E-1(1.26E-2)}} & 7.1827E-1(2.31E-2)\dag & 5.6208E-1(3.88E-3)\dag & 5.8056E-1(3.49E-2)\dag & 5.9833E-1(0.00E+0)\dag & 6.2159E-1(0.00E+0)\dag \\
		& ACC   & \bb{\textbf{7.3488E-1(1.41E-2)}} & 7.2037E-1(2.35E-2)$\approx$ & 6.4815E-1(3.09E-3)\dag & 6.5123E-1(3.09E-2)\dag & 6.7284E-1(0.00E+0)\dag & 6.9136E-1(0.00E+0)\dag \\
		& Recall & \bb{\textbf{7.1318E-1(6.63E-2)}} & 6.5891E-1(4.38E-2)\dag & 1.3953E-1(7.75E-3)\dag & 2.4806E-1(6.20E-2)\dag & 2.3256E-1(0.00E+0)\dag & 2.7907E-1(0.00E+0)\dag \\
  	& F1    & 
       \bb{\textbf{6.8055E-1(1.75E-2)}} & 6.6929E-1(3.16E-2)\dag & 2.4000E-1(1.17E-2)\dag & 3.5754E-1(8.20E-2)\dag & 3.6145E-1(0.00E+0)\dag & 4.1860E-1(0.00E+0)\dag \\
		& MCC   & \bb{\textbf{4.6069E-1(2.84E-2)}} & 4.3814E-1(4.32E-2)\dag & 2.4685E-1(9.81E-3)\dag & 2.3709E-1(9.06E-2)\dag & 3.0268E-1(0.00E+0)\dag & 3.5086E-1(0.00E+0)\dag \\
		\hline     
		\multirow{5}[2]{*}{JDT} & AUC   & \bb{\textbf{7.5882E-1(3.72E-3)}} & 7.4549E-1(4.52E-3)\dag & 6.8472E-1(1.48E-2)\dag & 6.1837E-1(2.25E-2)\dag & 7.0110E-1(0.00E+0)\dag & 6.6763E-1(0.00E+0)\dag \\
		& ACC   & \bb{\textbf{8.1303E-1(1.46E-2)}} & 7.2718E-1(5.72E-3)\dag & 8.4052E-1(4.51E-3)\dag & 7.8435E-1(1.96E-2)\dag & 8.3049E-1(0.00E+0)\dag & 8.3149E-1(0.00E+0)\dag \\
		& Recall & \bb{\textbf{6.6583E-1(3.20E-2)}} & 7.7670E-1(2.31E-2)\dag & 4.1748E-1(3.88E-2)\dag & 3.3010E-1(7.77E-2)\dag & 4.8058E-1(0.00E+0)\dag & 3.8835E-1(0.00E+0)\dag \\
  	& F1    & \bb{\textbf{5.9785E-1(1.00E-2)}} & 
        5.4054E-1(9.12E-3)\dag & 5.2121E-1(2.52E-2)\dag & 3.9067E-1(4.59E-2)\dag & 5.3951E-1(0.00E+0)\dag & 4.8780E-1(0.00E+0)\dag \\
		& MCC   & \bb{\textbf{4.8351E-1(1.43E-2)}} & 4.0811E-1(1.08E-2)\dag & 4.5325E-1(1.99E-2)\dag & 2.7083E-1(4.54E-2)\dag & 4.4254E-1(0.00E+0)\dag & 4.1420E-1(0.00E+0)\dag \\
		\hline     
		\multirow{5}[2]{*}{LC} & AUC   & \bb{\textbf{7.4165E-1(1.82E-2)}} & 7.1314E-1(3.03E-2)\dag & 6.1862E-1(2.06E-2)\dag & 6.1303E-1(3.28E-2)\dag & 6.3249E-1(0.00E+0)\dag & 6.2946E-1(0.00E+0)\dag \\
		& ACC   & 7.7106E-1(4.96E-2) & 7.9884E-1(3.20E-2)\ddag & \bb{\textbf{9.1896E-1(5.79E-3)}\ddag} & 8.2200E-1(2.17E-2)\dag & 9.0593E-1(0.00E+0)\dag & 9.1317E-1(0.00E+0)\dag \\
		& Recall & \bb{\textbf{7.1875E-1(5.70E-2)}} & 6.0938E-1(1.13E-1)\dag & 2.5000E-1(4.69E-2)\dag & 3.5938E-1(7.81E-2)\dag & 2.9688E-1(0.00E+0)\dag & 2.8125E-1(0.00E+0)\dag \\
  	& F1    & \bb{\textbf{3.7012E-1(3.28E-2)}} & 
        3.6098E-1(1.40E-2)$\approx$ & 3.6559E-1(3.38E-2)$\approx$ & 2.6891E-1(3.58E-2)\dag & 3.6893E-1(0.00E+0)$\approx$ & 3.7500E-1(0.00E+0)$\approx$ \\
		& MCC   & \bb{\textbf{3.2285E-1(2.53E-2)}} & 3.0218E-1(3.31E-2)\dag & 3.6583E-1(3.72E-2)\dag & 1.8176E-1(4.61E-2)\dag & 3.3287E-1(0.00E+0)\dag & 3.5717E-1(0.00E+0)\dag \\
		\hline     
		\multirow{5}[2]{*}{ML} & AUC   & \bb{\textbf{6.6464E-1(1.23E-2)}} & 6.5053E-1(4.39E-3)\dag & 5.7248E-1(5.26E-3)\dag & 5.5504E-1(1.68E-2)\dag & 6.0396E-1(0.00E+0)\dag & 6.0260E-1(0.00E+0)\dag \\
		& ACC   & 7.1267E-1(9.01E-2) & 7.8303E-1(1.19E-1)$\approx$ & \bb{\textbf{8.6412E-1(3.49E-3)}\ddag} & 8.4211E-1(8.06E-3)\ddag & 8.4748E-1(0.00E+0)\ddag & 8.5714E-1(0.00E+0)\ddag \\
		& Recall & \bb{\textbf{6.1633E-1(1.29E-1)}} & 4.6939E-1(1.84E-1)\dag & 1.7551E-1(1.43E-2)\dag & 1.6735E-1(5.51E-2)\dag & 2.7347E-1(0.00E+0)\dag & 2.5714E-1(0.00E+0)\dag \\
		& F1    & 3.6025E-1(2.53E-2) & \bb{\textbf{3.6249E-1(2.62E-2)}$\approx$} & 2.5507E-1(1.30E-2)\dag & 2.1563E-1(3.97E-2)\dag & 3.2057E-1(0.00E+0)\dag & 3.2143E-1(0.00E+0)\dag \\
        & MCC   & 2.5180E-1(2.29E-2) & 2.4916E-1(2.88E-2)$\approx$ & 2.2470E-1(1.55E-2)\dag & 1.4542E-1(4.25E-2)\dag & 2.4210E-1(0.00E+0)$\approx$ & \bb{\textbf{2.5723E-1(0.00E+0)}\dag} \\
		\hline     
		\multirow{5}[2]{*}{PDE} & AUC   & \bb{\textbf{6.8588E-1(3.15E-3)}} & 6.8587E-1(3.82E-3)$\approx$ & 6.1985E-1(9.41E-3)\dag & 5.7818E-1(2.79E-2)\dag & 6.1934E-1(0.00E+0)\dag & 6.5526E-1(0.00E+0)\dag \\
		& ACC   & 7.2345E-1(4.60E-2) & 7.1810E-1(6.41E-2)\dag & 8.3567E-1(6.01E-3)\ddag & 7.6954E-1(2.94E-2)\ddag & \bb{\textbf{8.4502E-1(0.00E+0)}\ddag} & 7.8958E-1(0.00E+0)\ddag \\
		& Recall & 6.2679E-1(7.27E-2) & \bb{\textbf{6.4115E-1(1.05E-1)}$\approx$} & 3.3014E-1(2.39E-2)\dag & 3.2057E-1(7.66E-2)\dag & 3.0622E-1(0.00E+0)\dag & 4.6890E-1(0.00E+0)\dag \\		
        & F1    & \bb{\textbf{3.8897E-1(1.71E-2)}} & 3.8841E-1(1.59E-2)$\approx$ & 3.5204E-1(1.82E-2)\dag & 2.7505E-1(3.96E-2)\dag & 3.5556E-1(0.00E+0)\dag & 3.8356E-1(0.00E+0)\dag \\
		& MCC   & \bb{\textbf{2.7615E-1(1.69E-2)}} & 2.7590E-1(1.23E-2)$\approx$ & 2.6135E-1(2.02E-2)\dag & 1.4058E-1(4.72E-2)\dag & 2.7468E-1(0.00E+0)$\approx$ & 2.6818E-1(0.00E+0)\dag \\
		\hline     
		\multirow{5}[2]{*}{Tomcat} & AUC   & \bb{\textbf{7.7628E-1(1.05E-2)}} & 7.6689E-1(5.12E-3)\dag & 7.2300E-1(1.27E-2)\dag & 6.3423E-1(7.36E-2)\dag & 6.7625E-1(0.00E+0)\dag & 7.2392E-1(0.00E+0)\dag \\
		& ACC   & \bb{\textbf{8.1067E-1(3.44E-2)}} & 7.4527E-1(4.02E-2)\dag & 7.4355E-1(1.33E-2)\dag & 4.5009E-1(1.70E-1)\dag & 7.1084E-1(0.00E+0)\dag & 7.4355E-1(0.00E+0)\dag \\
		& Recall & 7.4026E-1(5.19E-2) & 7.9221E-1(7.40E-2)\ddag & 7.0130E-1(3.25E-2)\dag & \bb{\textbf{8.5714E-1(2.01E-1)}\ddag} & 6.3636E-1(0.00E+0)\dag & 7.0130E-1(0.00E+0)\dag \\
		& F1    & \bb{\textbf{3.4132E-1(2.60E-2)}} & 2.9187E-1(1.88E-2)\dag & 2.6804E-1(7.94E-3)\dag & 1.7435E-1(4.42E-2)\dag & 2.2581E-1(0.00E+0)\dag & 2.6601E-1(0.00E+0)\dag \\
    	& MCC   & \bb{\textbf{3.3201E-1(1.65E-2)}} & 2.9057E-1(1.46E-2)\dag & 2.4874E-1(1.46E-2)\dag & 1.5334E-1(8.49E-2)\dag & 1.9005E-1(0.00E+0)\dag & 2.4727E-1(0.00E+0)\dag \\
		\hline     
		\multirow{5}[2]{*}{ant} & AUC   & \bb{\textbf{8.0558E-1(1.59E-3)}} & 7.8513E-1(1.66E-2)\dag & 6.4438E-1(1.45E-2)\dag & 6.2944E-1(1.11E-1)\dag & 6.0113E-1(0.00E+0)\dag & 6.3193E-1(0.00E+0)\dag \\
		& ACC   & 7.7861E-1(1.03E-2) & \bb{\textbf{7.8236E-1(1.41E-4)}\ddag} & 7.7017E-1(1.59E-2)\dag & 5.2345E-1(1.67E-1)\dag & 7.7861E-1(0.00E+0)$\approx$ & 7.4765E-1(0.00E+0)\dag \\
		& Recall & \bb{\textbf{8.4337E-1(1.20E-2)}} & 7.8916E-1(4.04E-2)\dag & 4.6988E-1(2.11E-2)\dag & 8.8554E-1(2.08E-1)$\approx$ & 3.4337E-1(0.00E+0)\dag & 4.6386E-1(0.00E+0)\dag \\
		& F1    & \bb{\textbf{5.4191E-1(6.10E-3)}} & 5.3036E-1(8.10E-3)\dag & 3.8342E-1(2.16E-2)\dag & 3.3619E-1(7.81E-2)\dag & 3.2571E-1(0.00E+0)\dag & 3.6407E-1(0.00E+0)\dag \\
        & MCC   & \bb{\textbf{4.7055E-1(5.75E-3)}} & 4.4801E-1(1.82E-2)\dag & 2.5405E-1(3.20E-2)\dag & 2.1564E-1(1.49E-1)\dag & 1.9406E-1(0.00E+0)\dag & 2.2367E-1(0.00E+0)\dag \\
		\hline     
		\multirow{5}[2]{*}{camel} & AUC   & \bb{\textbf{6.5497E-1(3.49E-3)}} & 6.1771E-1(0.00E+0)\dag & 5.3760E-1(9.45E-3)\dag & 5.9695E-1(3.41E-2)\dag & 5.3030E-1(0.00E+0)\dag & 5.5812E-1(0.00E+0)\dag \\
		& ACC   & 6.4537E-1(9.07E-2) & 7.8195E-1(4.59E-2)\ddag & 7.5240E-1(8.79E-3)\ddag & 5.7748E-1(1.24E-1)\dag & \bb{\textbf{7.9712E-1(0.00E+0)}\ddag} & 7.8115E-1(0.00E+0)\ddag \\
		& Recall & \bb{\textbf{6.5904E-1(1.39E-1)}} & 3.8298E-1(4.76E-2)\dag & 2.3404E-1(4.26E-2)\dag & 6.2766E-1(1.70E-1)$\approx$ & 1.4894E-1(0.00E+0)\dag & 2.3936E-1(0.00E+0)\dag \\
		& F1    & \bb{\textbf{3.6321E-1(1.37E-2)}} & 3.4532E-1(4.03E-3)\dag & 2.1945E-1(2.04E-2)\dag & 3.0621E-1(2.54E-2)\dag & 1.8065E-1(0.00E+0)\dag & 2.4725E-1(0.00E+0)\dag \\
        & MCC   & \bb{\textbf{2.2825E-1(1.45E-2)}} & 2.1754E-1(1.98E-2)\dag & 7.4035E-2(1.59E-2)\dag & 1.4066E-1(5.48E-2)\dag & 7.2985E-2(0.00E+0)\dag & 1.1947E-1(0.00E+0)\dag \\
		\hline     
		\multirow{5}[2]{*}{ivy} & AUC   & \bb{\textbf{8.0689E-1(3.22E-3)}} & 7.9857E-1(3.38E-2)\dag & 6.9614E-1(2.50E-2)\dag & 6.5578E-1(1.14E-1)\dag & 7.0521E-1(0.00E+0)\dag & 7.0283E-1(0.00E+0)\dag \\
		& ACC   & 7.3166E-1(1.78E-2) & 7.4004E-1(1.05E-1)\ddag & \bb{\textbf{8.4067E-1(8.39E-3)}\ddag} & 5.2411E-1(4.05E-1)\dag & 8.3438E-1(0.00E+0)\ddag & 8.0922E-1(0.00E+0)\ddag \\
		& Recall & 9.0000E-1(2.50E-2) & 8.2500E-1(1.53E-1)\dag & 5.2500E-1(6.25E-2)\dag & \bb{\textbf{9.5000E-1(2.25E-1)}$\approx$} & 5.5000E-1(0.00E+0)\dag & 5.7500E-1(0.00E+0)\dag \\
		& F1    & 3.6000E-1(8.76E-3) & \bb{\textbf{3.6181E-1(7.33E-2)}$\approx$} & 3.5398E-1(2.72E-2)$\approx$ & 2.2619E-1(8.39E-2)\dag & 3.5772E-1(0.00E+0)$\approx$ & 3.3577E-1(0.00E+0)\dag \\
        & MCC   & 3.6177E-1(5.65E-3) & \bb{\textbf{3.6368E-1(6.07E-2)}$\approx$} & 2.9249E-1(3.27E-2)\dag & 2.0541E-1(1.17E-1)\dag & 3.0006E-1(0.00E+0)\dag & 2.7936E-1(0.00E+0)\dag \\
		\hline     
		\multirow{5}[2]{*}{jEdit} & AUC   & \bb{\textbf{8.6451E-1(2.35E-2)}} & 7.5485E-1(1.16E-1)\dag & 6.9865E-1(2.21E-2)\dag & 5.9253E-1(6.92E-2)\dag & 6.6876E-1(0.00E+0)\dag & 6.5676E-1(0.00E+0)\dag \\
		& ACC   & 7.8330E-1(2.52E-2) & \bb{\textbf{8.3569E-1(6.58E-2)}$\approx$} & 7.6855E-1(3.98E-2)\dag & 5.4240E-1(1.09E-1)\dag & 7.0053E-1(0.00E+0)\dag & 7.6590E-1(0.00E+0)\dag \\
		& Recall & \bb{\textbf{9.5455E-1(3.64E-2)}} & 7.8182E-1(2.73E-1)\dag & 6.3636E-1(0.00E+0)\dag & 6.3636E-1(4.55E-2)\dag & 6.3636E-1(0.00E+0)\dag & 5.4545E-1(0.00E+0)\dag \\		
        & F1    & 7.6608E-2(7.47E-3) & \bb{\textbf{8.2840E-2(3.01E-2)}$\approx$} & 4.9645E-2(7.86E-3)$\approx$ & 2.6866E-2(8.61E-3)\dag & 3.9660E-2(0.00E+0)\dag & 4.3321E-2(0.00E+0)\dag \\
		& MCC   & \bb{\textbf{1.6883E-1(1.70E-2)}} & 1.4200E-1(4.74E-2)\dag & 9.0876E-2(1.52E-2)\dag & 3.7771E-2(3.56E-2)\dag & 7.2107E-2(0.00E+0)\dag & 7.2538E-2(0.00E+0)\dag \\
		\hline     
		\multirow{5}[2]{*}{log4j} & AUC   & \bb{\textbf{8.4588E-1(1.03E-2)}} & 7.8708E-1(8.88E-2)\dag & 5.6085E-1(5.29E-3)\dag & 6.7127E-1(1.80E-1)\dag & 5.4762E-1(0.00E+0)\dag & 5.7143E-1(0.00E+0)\dag \\
		& ACC   & \bb{\textbf{8.4043E-1(4.76E-2)}} & 7.7305E-1(1.13E-1)\dag & 4.1135E-1(7.09E-3)\dag & 5.8511E-1(2.94E-1)\dag & 3.9362E-1(0.00E+0)\dag & 4.2553E-1(0.00E+0)\dag \\
		& Recall & \bb{\textbf{8.3519E-1(1.40E-1)}} & 7.4074E-1(2.14E-1)\dag & 1.2169E-1(1.06E-2)\dag & 4.1270E-1(5.13E-1)\dag & 9.5238E-2(0.00E+0)\dag & 1.4286E-1(0.00E+0)\dag \\
		& F1    & \bb{\textbf{8.7436E-1(5.14E-2)}} & 8.0606E-1(1.20E-1)\dag & 2.1698E-1(1.68E-2)\dag & 5.7143E-1(4.73E-1)\dag & 1.7391E-1(0.00E+0)\dag & 2.5000E-1(0.00E+0)\dag \\
        & MCC   & \bb{\textbf{6.7053E-1(4.87E-2)}} & 5.4099E-1(1.64E-1)\dag & 2.0904E-1(9.90E-3)\dag & 3.5794E-1(2.59E-1)\dag & 1.8317E-1(0.00E+0)\dag & 2.2826E-1(0.00E+0)\dag \\
		\hline 
		\multirow{5}[2]{*}{lucene} & AUC   & \bb{\textbf{7.4047E-1(2.48E-3)}} & 7.1244E-1(2.88E-3)\dag & 5.8303E-1(4.72E-3)\dag & 6.0340E-1(6.29E-2)\dag & 6.0977E-1(0.00E+0)\dag & 5.9181E-1(0.00E+0)\dag \\
		& ACC   & \bb{\textbf{7.4627E-1(2.79E-2)}} & 7.4067E-1(0.00E+0)$\approx$ & 6.6418E-1(4.66E-3)\dag & 6.3806E-1(8.68E-2)\dag & 6.7164E-1(0.00E+0)\dag & 6.6604E-1(0.00E+0)\dag \\
		& Recall & \bb{\textbf{7.4384E-1(1.13E-1)}} & 6.1084E-1(1.48E-2)\dag & 2.5123E-1(9.85E-3)\dag & 5.6650E-1(5.20E-1)$\approx$ & 3.5468E-1(0.00E+0)\dag & 2.8571E-1(0.00E+0)\dag \\		
        & F1    & \bb{\textbf{6.8481E-1(1.07E-2)}} & 6.3517E-1(5.66E-3)\dag & 3.6299E-1(1.03E-2)\dag & 5.6250E-1(1.83E-1)\dag & 4.5000E-1(0.00E+0)\dag & 3.9322E-1(0.00E+0)\dag \\
		& MCC   & \bb{\textbf{4.7439E-1(1.87E-2)}} & 4.3762E-1(2.35E-3)\dag & 2.2850E-1(1.35E-2)\dag & 2.5902E-1(1.21E-1)\dag & 2.5781E-1(0.00E+0)\dag & 2.3620E-1(0.00E+0)\dag \\
		\hline     
		\multirow{5}[2]{*}{poi} & AUC   & \bb{\textbf{8.2990E-1(9.93E-4)}} & 8.1279E-1(2.64E-2)\dag & 6.1149E-1(2.32E-2)\dag & 6.1197E-1(1.23E-1)\dag & 5.9327E-1(0.00E+0)\dag & 6.0816E-1(0.00E+0)\dag \\
		& ACC   & \bb{\textbf{8.2863E-1(1.88E-3)}} & 8.1299E-1(2.89E-2)\dag & 5.9699E-1(2.35E-2)\dag & 6.3089E-1(1.38E-1)\dag & 5.8004E-1(0.00E+0)\dag & 5.9322E-1(0.00E+0)\dag \\
		& Recall & \bb{\textbf{8.2206E-1(1.78E-2)}} & 8.0427E-1(4.63E-2)\dag & 3.5587E-1(3.20E-2)\dag & 7.3665E-1(5.18E-1)$\approx$ & 3.6655E-1(0.00E+0)\dag & 3.5231E-1(0.00E+0)\dag \\
		& F1    & \bb{\textbf{8.3453E-1(2.71E-3)}} & 8.2208E-1(3.05E-2)\dag & 4.8426E-1(3.17E-2)\dag & 7.1954E-1(2.49E-1)\dag & 4.8019E-1(0.00E+0)\dag & 4.7826E-1(0.00E+0)\dag \\
        & MCC   & \bb{\textbf{6.5871E-1(3.29E-3)}} & 6.2515E-1(5.03E-2)\dag & 2.5439E-1(4.93E-2)\dag & 3.0450E-1(2.66E-1)\dag & 2.0767E-1(0.00E+0)\dag & 2.4919E-1(0.00E+0)\dag \\
		\hline     
		\multirow{5}[2]{*}{synapse} & AUC   & \bb{\textbf{7.2071E-1(7.35E-3)}} & 6.9996E-1(1.69E-2)\dag & 5.6402E-1(2.10E-2)\dag & 5.4610E-1(6.27E-2)\dag & 5.6478E-1(0.00E+0)\dag & 5.9652E-1(0.00E+0)\dag \\
		& ACC   & \bb{\textbf{7.2491E-1(3.53E-2)}} & 7.1111E-1(5.58E-2)$\approx$ & 6.9517E-1(2.04E-2)\dag & 5.0186E-1(2.40E-1)\dag & 6.4684E-1(0.00E+0)\dag & 7.0260E-1(0.00E+0)\dag \\
		& Recall & 7.0930E-1(1.05E-1) & 6.3953E-1(1.10E-1)$\approx$ & 2.2093E-1(3.49E-2)\dag & \bb{\textbf{8.9535E-1(4.53E-1)}$\approx$} & 3.3721E-1(0.00E+0)\dag & 3.0233E-1(0.00E+0)\dag \\
		& F1    & \bb{\textbf{6.1988E-1(1.01E-2)}} & 6.0094E-1(1.46E-2)\dag & 3.0645E-1(4.25E-2)\dag & 4.9704E-1(6.65E-2)\dag & 3.7908E-1(0.00E+0)\dag & 3.9394E-1(0.00E+0)\dag \\
        & MCC   & \bb{\textbf{4.1778E-1(2.61E-2)}} & 3.8026E-1(4.68E-2)\dag & 1.8762E-1(6.47E-2)\dag & 1.3351E-1(1.63E-1)\dag & 1.3971E-1(0.00E+0)\dag & 2.3910E-1(0.00E+0)\dag \\
		\hline     
		\multirow{5}[2]{*}{velocity} & AUC   & \bb{\textbf{7.1774E-1(7.52E-3)}} & 6.9830E-1(1.06E-2)\dag & 5.4529E-1(2.24E-2)\dag & 5.3941E-1(7.30E-2)\dag & 5.5149E-1(1.84E-3)\dag & 5.5906E-1(0.00E+0)\dag \\
		& ACC   & 7.0881E-1(2.87E-2) & \bb{\textbf{7.4713E-1(4.23E-2)}\ddag} & 6.7816E-1(1.72E-2)\dag & 6.0920E-1(2.70E-1)\dag & 7.0115E-1(0.00E+0)$\approx$ & 6.7050E-1(0.00E+0)\dag \\
		& Recall & \bb{\textbf{7.1795E-1(8.97E-2)}} & 5.7692E-1(7.69E-2)\dag & 2.0513E-1(5.77E-2)\dag & 5.6410E-1(5.83E-1)\dag & 1.7949E-1(6.41E-3)\dag & 2.8205E-1(0.00E+0)\dag \\
  	& F1    & \bb{\textbf{6.0335E-1(8.12E-3)}} & 
        5.7692E-1(1.01E-2)\dag & 2.8099E-1(5.10E-2)\dag & 4.3529E-1(1.60E-1)\dag & 2.6415E-1(6.81E-3)\dag & 3.3846E-1(0.00E+0)\dag \\
		& MCC   & \bb{\textbf{4.0130E-1(1.12E-2)}} & 3.9660E-1(4.18E-2)\dag & 1.2410E-1(5.31E-2)\dag & 1.2072E-1(1.42E-1)\dag & 1.5233E-1(3.01E-3)\dag & 1.3537E-1(0.00E+0)\dag \\
		\hline     
		\multirow{5}[2]{*}{xalan} & AUC   & \bb{\textbf{8.6577E-1(3.06E-2)}} & 7.5781E-1(4.74E-3)\dag & 5.6566E-1(8.06E-3)\dag & 6.0570E-1(5.40E-2)\dag & 5.5547E-1(0.00E+0)\dag & 6.0278E-1(0.00E+0)\dag \\
		& ACC   & \bb{\textbf{8.8212E-1(2.58E-2)}} & 6.8509E-1(1.34E-2)\dag & 3.5176E-1(1.51E-2)\dag & 4.3719E-1(8.84E-2)\dag & 3.5008E-1(0.00E+0)\dag & 4.2295E-1(0.00E+0)\dag \\
		& Recall & \bb{\textbf{9.1403E-1(2.30E-2)}} & 6.1359E-1(3.12E-2)\dag & 1.4365E-1(2.17E-2)\dag & 2.7506E-1(1.07E-1)\dag & 1.4811E-1(0.00E+0)\dag & 2.4610E-1(0.00E+0)\dag \\
		& F1    & \bb{\textbf{9.1985E-1(1.64E-2)}} & 7.4560E-1(1.72E-2)\dag & 2.5000E-1(3.26E-2)\dag & 4.2096E-1(1.31E-1)\dag & 2.5528E-1(0.00E+0)\dag & 3.9080E-1(0.00E+0)\dag \\
        & MCC   & \bb{\textbf{7.0150E-1(7.16E-2)}} & 4.4546E-1(8.03E-3)\dag & 1.8014E-1(1.81E-2)\dag & 2.1450E-1(8.04E-2)\dag & 1.4710E-1(0.00E+0)\dag & 2.2397E-1(0.00E+0)\dag \\
		\hline     
		\multirow{5}[1]{*}{xerces} & AUC   & \bb{\textbf{7.7253E-1(3.24E-2)}} & 6.7063E-1(3.76E-2)\dag & 5.7833E-1(5.27E-3)\dag & 5.7269E-1(8.26E-2)\dag & 5.7673E-1(0.00E+0)\dag & 5.7192E-1(0.00E+0)\dag \\
		& ACC   & \bb{\textbf{7.8666E-1(4.72E-2)}} & 6.2146E-1(7.75E-2)\dag & 4.6498E-1(3.73E-3)\dag & 6.0358E-1(1.10E-1)\dag & 4.5902E-1(0.00E+0)\dag & 4.6051E-1(0.00E+0)\dag \\
		& Recall & \bb{\textbf{8.2941E-1(5.50E-2)}} & 4.7140E-1(2.21E-1)\dag & 2.0595E-1(6.86E-3)\dag & 6.4531E-1(3.24E-1)\dag & 1.8764E-1(0.00E+0)\dag & 2.0366E-1(0.00E+0)\dag \\
  	& F1    & \bb{\textbf{8.3507E-1(3.89E-2)}} & 
        6.1862E-1(1.37E-1)\dag & 3.3520E-1(8.68E-3)\dag & 6.9742E-1(1.91E-1)\dag & 3.1120E-1(0.00E+0)\dag & 3.2963E-1(0.00E+0)\dag \\
		& MCC   & \bb{\textbf{5.4648E-1(7.37E-2)}} & 3.5037E-1(5.47E-2)\dag & 2.0751E-1(1.42E-2)\dag & 1.9861E-1(1.23E-1)\dag & 2.1460E-1(0.00E+0)\dag & 1.9016E-1(0.00E+0)\dag \\
		\hline 
		\multirow{5}[1]{*}{Apache} & AUC   & 7.4745E-1(6.03E-3) & \bb{\textbf{7.5298E-1(0.00E+0)}\ddag} & 6.7953E-1(0.00E+0)\dag & 6.5306E-1(4.10E-2)\dag & 6.8389E-1(0.00E+0)\dag & 6.9005E-1(0.00E+0)\dag \\
		& ACC   & 7.4742E-1(6.19E-3) & \bb{\textbf{7.5258E-1(0.00E+0)}\ddag} & 6.8041E-1(0.00E+0)\dag & 6.5464E-1(4.12E-2)\dag & 6.8557E-1(0.00E+0)\dag & 6.9072E-1(0.00E+0)\dag \\
		& Recall & 7.4490E-1(5.15E-2) & 7.1429E-1(8.16E-2)$\approx$ & 7.6531E-1(0.00E+0)$\approx$ & 6.5306E-1(1.38E-1)\dag & 8.4694E-1(0.00E+0)\ddag & \bb{\textbf{7.5510E-1(0.00E+0)}$\approx$} \\
		& F1    & \bb{\textbf{7.4872E-1(1.13E-2)}} & 7.4468E-1(7.70E-3)$\approx$ & 7.0755E-1(1.66E-3)\dag & 6.5285E-1(7.90E-2)\dag & 7.3128E-1(0.00E+0)\dag & 7.1154E-1(0.00E+0)\dag \\
        & MCC   & 4.9487E-1(1.20E-2) & 5.0725E-1(0.00E+0)\dag & 3.6468E-1(0.00E+0)\dag & 3.0929E-1(8.38E-2)\dag & 3.8956E-1(0.00E+0)\dag & 3.8354E-1(0.00E+0)\dag \\
		\hline
		\multirow{5}[2]{*}{Safe} & AUC   & \bb{\textbf{8.0949E-1(2.48E-2)}} & 8.0348E-1(1.13E-2)$\approx$ & 7.1390E-1(3.41E-2)\dag & 6.0963E-1(4.95E-2)\dag & 7.2193E-1(0.00E+0)\dag & 6.7513E-1(0.00E+0)\dag \\
		& ACC   & 8.0357E-1(2.14E-2) & \bb{\textbf{8.2353E-1(3.10E-2)}\ddag} & 7.5000E-1(2.68E-2)\dag & 6.4286E-1(5.36E-2)\dag & 7.5000E-1(0.00E+0)\dag & 7.3214E-1(0.00E+0)\dag \\
		& Recall & \bb{\textbf{8.4091E-1(1.11E-1)}} & 6.8182E-1(1.31E-1)\dag & 5.4545E-1(6.82E-2)\dag & 4.5455E-1(1.36E-1)\dag & 5.9091E-1(0.00E+0)\dag & 4.0909E-1(0.00E+0)\dag \\		
        & F1    & \bb{\textbf{7.6964E-1(3.09E-2)}} & 7.5676E-1(1.30E-2)\dag & 6.3158E-1(5.47E-2)\dag & 5.0000E-1(9.65E-2)\dag & 6.5000E-1(0.00E+0)\dag & 5.4545E-1(0.00E+0)\dag \\
		& MCC   & 6.2437E-1(4.42E-2) & \bb{\textbf{6.4288E-1(5.87E-2)}$\approx$} & 4.6250E-1(6.19E-2)\dag & 2.2928E-1(1.11E-1)\dag & 4.6415E-1(0.00E+0)\dag & 4.3057E-1(0.00E+0)\dag \\
		\hline    
		\multirow{5}[2]{*}{Zxing} & AUC   & \bb{\textbf{6.5419E-1(2.97E-3)}} & 6.4046E-1(8.08E-3)\dag & 5.4984E-1(4.24E-3)\dag & 5.3841E-1(2.36E-2)\dag & 5.4960E-1(0.00E+0)\dag & 5.7111E-1(0.00E+0)\dag \\
		& ACC   & 6.3659E-1(7.52E-3) & 6.5363E-1(7.09E-2)\ddag & \bb{\textbf{7.0175E-1(2.51E-3)}\ddag} & 6.5664E-1(2.26E-2)\ddag & 6.3910E-1(0.00E+0)\ddag & 6.8672E-1(0.00E+0)\ddag \\
		& Recall & \bb{\textbf{7.0339E-1(3.39E-2)}} & 5.9153E-1(1.80E-1)\dag & 1.7797E-1(8.47E-3)\dag & 2.3729E-1(4.66E-2)\dag & 3.3051E-1(0.00E+0)\dag & 2.8814E-1(0.00E+0)\dag \\		
        & F1    & \bb{\textbf{5.3115E-1(6.34E-3)}} & 5.0833E-1(2.15E-2)\dag & 2.6087E-1(1.07E-2)\dag & 2.9348E-1(5.64E-2)\dag & 3.5135E-1(0.00E+0)\dag & 3.5233E-1(0.00E+0)\dag \\
		& MCC   & \bb{\textbf{2.8157E-1(6.33E-3)}} & 2.6412E-1(1.63E-2)\dag & 1.4670E-1(1.09E-2)\dag & 9.5775E-2(5.23E-2)\dag & 1.0312E-1(0.00E+0)\dag & 1.6614E-1(0.00E+0)\dag \\
		%\hline
		\midrule
		\multicolumn{3}{c}{\dag/$\approx$/\ddag}
		& 70/20/10
		& 90/4/6
		& 90/6/4
		& 88/6/6
		& 93/2/5\\
		%\hline
		\bottomrule     
		\end{tabular}%
	\label{tab:automl_table}%
	\begin{tablenotes} 
		\item\dag\, denotes the performance of \our\ is significantly better than the other peers according to the Wilcoxon rank sum test at a $0.05$ significance level while \ddag\, denotes the opposite case, and $\approx$ denotes that there is no significant difference between the two tools.
	\end{tablenotes} 
\end{table*}%

%% file: related.tex
%!TeX root=main.tex

\section{Related Works}
\label{sec:related}

This section reviews previous work on HPO and AutoML that are relevant to this work.

%In SDP, automatic parameter optimization has been proven to have a significant impact on model performance. In [4,5], they point out that automatic parameter optimization is particularly important for parametric sensitive classification techniques, and therefore suggest that researchers try automatic parameter optimization in experiments instead of relying on default parameter settings. Similarly, Gupta et al.[6] found that hyperparameter tuning has a significant impact on the performance of deep neural network models in fault prediction, where the correct parameter settings can significantly improve the performance of classifiers.
%In CPDP, it is not uncommon to use HPO to find optimal hyperparameters for various tasks, such as software effort estimation~\cite{SongMY13}, search algorithms~\cite{ArcuriF13}, and SD counts~\cite{NevendraS22}. 
In CPDP, it is not uncommon to use HPO to find optimal hyperparameters for various tasks, and several studies have shown that HPO impacts model performance significantly. For example, Qu et al.~\cite{QuCZJ18} found that HPO is crucial for several CPDP methods, especially those using Support Vector Machines (SVM) and K-nearest neighbors (KNN) classifiers. The study in~\cite{Ozturk19} showed that HPO significantly impacts the success of both CPDP and within-project defect prediction, with CPDP generally outperforming within-project defect prediction after HPO. In~\cite{Ozturk191}, the impact of HPO in ensemble learning applied to CPDP and the importance of HPO for prediction performance was discussed. Li et al.~\cite{LiXCT20} investigated the HPO of transfer learners and classifiers, pointing out that the HPO of both learners can effectively enhance the predictive performance of CPDP. Although these methods effectively enhance performance, they only focus on optimizing parameters for classifiers and transfer learning. This focus may ignore other performance factors, such as feature selection or optimization strategy.

%Regarding optimization strategy, there has been a wealth of fundamental and methodological research for CPDP. In particular, they mainly tackle single-objective, single-level problems. In contrast, the research on multi-objective or bilevel problems has been lukewarm in the literature. Canfora et al.~\cite{CanforaLPOPP13,CanforaLPOPP15} proposed a multi-objective strategy for CPDP that maximizes detection effectiveness while minimizing the associated lines of code costs. Their findings show that this approach is more cost-effective than single-objective methods. In~\cite{RyuB16}, they introduce novel multi-objective learning techniques for CPDP that effectively handle class imbalance by employing the harmony search algorithm to optimize multiple objectives. Kanwar et al.~\cite{KanwarAS23} introduce a multi-objective RF algorithm combined with a data resampling technique for CPDP to minimize false alarms and maximize detection probability while addressing class imbalance issues. Although these algorithms are effective, they did not consider the internal nested hierarchical issues brought by the model parameters. In terms of structure, Li et al.~\cite{LiXCT20} construct CPDP considering the impact of parameters and construct it as a single-objective bilevel optimization model. This tool demonstrates improved performance in terms of AUC. However, it only considers a single optimization objective and does not consider other potential influencing factors, which may limit the model's generalizability and diversity.

In CPDP, finding the optimal ML pipeline is also crucial beyond hyperparameter settings. Although many methods have been compared in CPDP, there is no consensus on the best one~\cite{HosseiniTG19}. This lack of consensus highlights the need for a more systematic approach, such as AutoML, to identify optimal ML pipelines. AutoML streamlines the process of selecting and optimizing ML models. It automates tasks such as feature engineering, model selection, and HPO, making it easier to develop high-performing models with minimal human intervention. There are numerous AutoML frameworks in the ML domain. For example, \texttt{Auto-WEKA}~\cite{ThorntonHHL13} began the trend by automating model selection and hyperparameter tuning. Then, \texttt{Auto-sklearn}~\cite{FeurerKESBH15} built upon this by leveraging scikit-learn's algorithms. \texttt{H2O}~\cite{LeDellP20} AutoML focused on enhancing model performance through ensemble methods. \texttt{AutoKeras}~\cite{JinSH19} simplified deep learning with neural architecture search. \texttt{MLjar}~\cite{mljar} provided a user-friendly platform for automated model tuning. Lastly, \texttt{SapientML}~\cite{SahaUM0LHYKP22} emphasized interpretability and fairness in automated modeling. Although AutoML is an effective tool, its application in SE remains limited. For example, Tanaka et al.~\cite{TanakaMY19} investigated the effectiveness of \texttt{Auto-sklearn} in predicting SDs across different releases in $20$ open-source projects. Another study~\cite{CalefatoQLK23} examined the potential of AutoML in data-driven SE, focusing on its ability to automate AI/ML pipeline construction and address skill gaps. These studies highlight the growing interest and potential of AutoML within the SE domain.

%Existing AutoML methods are mainly optimized for specific ML tasks, and the data distribution and feature space of CPDP may be significantly different from these tasks. Therefore, directly applying existing AutoML methods to CPDP may not achieve the expected prediction effect. 

%Current CPDP methods are primarily focused on optimizing single learners for optimal performance. However, ensembles of multiple promising learners are often known to yield better results than individual learners. This approach has been successfully demonstrated in various competitions; for example, He et al.~\cite{HeZRS16} secured first place in the ILSRVC2015 competition by ensembling predictions from multiple learners, and similarly, ensemble methods frequently achieve top rankings in data science competitions like Kaggle~\cite{Hoch15,abs-2009-07701}. Inspired by these successes, modern AutoML systems such as \texttt{Auto-sklearn}, \texttt{Auto-Pytorch}~\cite{ZimmerLH21}, and \texttt{VolcanoML}~\cite{LiSZJLDZY00021} have adopted post-hoc ensemble strategies that leverage all base learners throughout the optimization process, resulting in superior empirical performance compared to single best learners. Despite the effectiveness of these post-hoc ensemble designs in practice, their objectives do not fully align with those of CPDP methods. Therefore, the challenge of simultaneously guiding the search for both performance and diversity in a specific CPDP task remains an open question.

%% file: threats.tex
%!TeX root=main.tex

\section{Threats to Validity}
\label{sec:threats}

In this section, we follow the threat classification schemes for experiment validity described by Cook~\cite{CookCD79}. 

Internal validity threats relate to how the experimental setup might influence the performance, focusing on the parameters used within the study. In our case, the time budget for optimization is a key parameter that could affect the results. To address this, we adopt a total time budget per the recommendations from the state-of-the-practice within the AutoML community~\cite{FeurerKESBH15}. This budget is considered reasonable for the number of runs required. 

External validity concerns relate to the generalizability of our findings to other SE projects. To mitigate such threats, we select $20$ diverse projects that span a wide spectrum of real-world scenarios, as outlined in~\pref{sec:settings}. This ensures that our findings are not limited to a narrow set of conditions or contexts. Each project within our dataset is employed as the target domain data, with the remaining projects in the same dataset serving as source datasets. This cross-validation approach strengthens the argument that our findings apply to a broad range of software projects beyond those included in our study.

Construct validity concerns in our study arise from the experimental uncertainties of the optimization algorithms. Such uncertainties could skew the experimental results, misrepresenting the effectiveness of the evaluated techniques. To mitigate these threats, we execute $31$ runs for each technique to ensure that the results are not an artifact of random chance. Moreover, the comparative analysis of the techniques is bolstered by employing the Scott-Knott test~\cite{MittasA13}, which is further supported by the Wilcoxon rank sum test~\cite{Haynes2013} and the $A_{12}$ effect size~\cite{VarghaD00}. This rigorous statistical framework allows us to assert that when technique A is reported as superior to technique B, it is indeed statistically better with a large effect size.

%% file: conclusion.tex
%!TeX root=main.tex

\section{Conclusion}
\label{sec:conclusion}

CPDP is a crucial task in SE, aiming to predict SDs across different projects. The inherent variability in cross-project data, characterized by feature inconsistency and distributional discrepancies, often leads to a decline in predictive accuracy. To tackle this challenge, we propose an AutoML method, \our, which utilizes meticulously selected feature selection, transfer learning, and classification techniques as its base components, effectively enhancing SDP effectiveness. Given the significance of hyperparameters and the necessity to explore the search space efficiently, we propose a MBLO structure. This structure optimizes the ML pipelines and their corresponding parameters, ensuring a comprehensive search space exploration. In particular, we utilize a multi-objective approach for the upper-level optimization problem to overcome the limitations of single-objective optimization, thereby effectively balancing various performance metrics. To further improve predictive performance, we present an advanced ensemble learning method that integrates ensemble techniques and local search. This approach ensures accuracy and maintains diversity among ensemble members, which is critical for robust predictions. Additionally, we design a MBLO algorithm to solve this complex problem, leveraging the strengths of multi-objective tabu search and TPE. Through extensive experimental comparisons across multiple datasets and scenarios, we demonstrate the effectiveness of our approach in addressing the CPDP problem. Our results showcase the potential of \our~to improve SDP across diverse software projects significantly, highlighting their practical implications in enhancing software quality and reliability.

Based on the tool in this paper, we envisage the following aspects for future endeavors.
\begin{enumerate}
	\item Incorporating the tool into continuous integration and continuous deployment pipelines could provide real-time SDP, thereby enhancing the efficiency and reliability of software development processes.
	\item In the process of HPO, we attempt to optimize all hyperparameters. However, some hyperparameters have a significant impact on performance, while others have a minor impact. In the future, we aim to employ advanced techniques to optimize hyperparameters based on their relative importance to performance selectively.
	\item Enhancing the tool with interpretable and explainable AI features would offer insights into its decision-making process, thereby fostering trust and understanding among users.	
	\item Exploring collaborative learning approaches where multiple projects contribute to a shared model could potentially improve prediction accuracy and reduce the need for extensive retraining.	
	\item While the current tool is designed for CPDP, its underlying principles and methodologies could be adapted and extended to other domains within SE or even other fields where predictive modeling is crucial.
	
\end{enumerate}

%% file: main_tech_report.bbl
\begin{thebibliography}{100}
\providecommand{\url}[1]{#1}
\csname url@samestyle\endcsname
\providecommand{\newblock}{\relax}
\providecommand{\bibinfo}[2]{#2}
\providecommand{\BIBentrySTDinterwordspacing}{\spaceskip=0pt\relax}
\providecommand{\BIBentryALTinterwordstretchfactor}{4}
\providecommand{\BIBentryALTinterwordspacing}{\spaceskip=\fontdimen2\font plus
\BIBentryALTinterwordstretchfactor\fontdimen3\font minus
  \fontdimen4\font\relax}
\providecommand{\BIBforeignlanguage}[2]{{%
\expandafter\ifx\csname l@#1\endcsname\relax
\typeout{** WARNING: IEEEtran.bst: No hyphenation pattern has been}%
\typeout{** loaded for the language `#1'. Using the pattern for}%
\typeout{** the default language instead.}%
\else
\language=\csname l@#1\endcsname
\fi
#2}}
\providecommand{\BIBdecl}{\relax}
\BIBdecl

\bibitem{biedron2024}
S.~R. Biedron, ``Cybercrime in the digital age,'' Ph.D. dissertation,
  University of Oxford, 2024.

\bibitem{QiuLJ18}
S.~Qiu, L.~Lu, and S.~Jiang, ``Multiple-components weights model for
  cross-project software defect prediction,'' \emph{{IET} Softw.}, vol.~12,
  no.~4, pp. 345--355, 2018.

\bibitem{HosseiniTG19}
S.~Hosseini, B.~Turhan, and D.~Gunarathna, ``A systematic literature review and
  meta-analysis on cross project defect prediction,'' \emph{{IEEE} Trans.
  Software Eng.}, vol.~45, no.~2, pp. 111--147, 2019.

\bibitem{HerboldTG18}
S.~Herbold, A.~Trautsch, and J.~Grabowski, ``A comparative study to benchmark
  cross-project defect prediction approaches,'' \emph{{IEEE} Trans. Software
  Eng.}, vol.~44, no.~9, pp. 811--833, 2018.

\bibitem{TongZLXLLW24}
H.~Tong, D.~Zhang, J.~Liu, W.~Xing, L.~Lu, W.~Lu, and Y.~Wu, ``{MASTER:}
  multi-source transfer weighted ensemble learning for multiple sources
  cross-project defect prediction,'' \emph{{IEEE} Trans. Software Eng.},
  vol.~50, no.~5, pp. 1281--1305, 2024.

\bibitem{QuCZJ18}
Y.~Qu, X.~Chen, Y.~Zhao, and X.~Ju, ``Impact of hyper parameter optimization
  for cross-project software defect prediction,'' \emph{International Journal
  of Performability Engineering}, vol.~14, no.~6, p. 1291, 2018.

\bibitem{Ozturk191}
M.~M. {\"{O}}zt{\"{u}}rk, ``The impact of parameter optimization of ensemble
  learning on defect prediction,'' \emph{Comput. Sci. J. Moldova}, vol.~27,
  no.~1, pp. 85--128, 2019.

\bibitem{KwonRB23}
S.~Kwon, D.~Ryu, and J.~Baik, ``An effective approach to improve the
  performance of ecpdp (early cross-project defect prediction) via
  data-transformation and parameter optimization,'' \emph{Softw. Qual. J.},
  vol.~31, no.~4, pp. 1009--1044, 2023.

\bibitem{WanZQLQ24}
X.~Wan, Z.~Z.~F. Qin, X.~Lu, and K.~Qiu, ``Adjusted trust score: A novel
  approach for estimating the trustworthiness of software defect prediction
  models,'' \emph{IEEE Transactions on Reliability}, 2024.

\bibitem{OmondiagbeLM24}
O.~P. Omondiagbe, S.~A. Licorish, and S.~G. MacDonell, ``Improving transfer
  learning for software cross-project defect prediction,'' \emph{Appl.
  Intell.}, vol.~54, no.~7, pp. 5593--5616, 2024.

\bibitem{ChenL21}
\BIBentryALTinterwordspacing
R.~Chen and K.~Li, ``Transfer bayesian optimization for expensive black-box
  optimization in dynamic environment,'' in \emph{2021 {IEEE} International
  Conference on Systems, Man, and Cybernetics, {SMC} 2021, Melbourne,
  Australia, October 17-20, 2021}.\hskip 1em plus 0.5em minus 0.4em\relax
  {IEEE}, 2021, pp. 1374--1379. [Online]. Available:
  \url{https://doi.org/10.1109/SMC52423.2021.9659200}
\BIBentrySTDinterwordspacing

\bibitem{WangL24}
S.~Wang and K.~Li, ``Constrained bayesian optimization under partial
  observations: Balanced improvements and provable convergence,'' in
  \emph{AAAI'24: Proc. of the Thirty-Eighth {AAAI} Conference on Artificial
  Intelligence}.\hskip 1em plus 0.5em minus 0.4em\relax {AAAI} Press, 2024, pp.
  15\,607--15\,615.

\bibitem{ChenL23}
R.~Chen and K.~Li, ``Data-driven evolutionary multi-objective optimization
  based on multiple-gradient descent for disconnected pareto fronts,'' in
  \emph{EMO'23: Proc. of the 12th International Conference on Evolutionary
  Multi-Criterion Optimization}, ser. Lecture Notes in Computer Science, vol.
  13970.\hskip 1em plus 0.5em minus 0.4em\relax Springer, 2023, pp. 56--70.

\bibitem{LiC23}
K.~Li and R.~Chen, ``Batched data-driven evolutionary multiobjective
  optimization based on manifold interpolation,'' \emph{{IEEE} Trans. Evol.
  Comput.}, vol.~27, no.~1, pp. 126--140, 2023.

\bibitem{XuLL24}
J.~Xu, K.~Li, and D.~Li, ``An automated few-shot learning for time-series
  forecasting in smart grid under data scarcity,'' \emph{{IEEE} Trans. Artif.
  Intell.}, vol.~5, no.~6, pp. 2482--2492, 2024.

\bibitem{LiCY24}
K.~Li, R.~Chen, and X.~Yao, ``A data-driven evolutionary transfer optimization
  for expensive problems in dynamic environments,'' \emph{{IEEE} Trans. Evol.
  Comput.}, vol.~28, no.~5, pp. 1396--1411, 2024.

\bibitem{ChenLTL22}
L.~Chen, H.~Liu, K.~C. Tan, and K.~Li, ``Transfer learning-based parallel
  evolutionary algorithm framework for bilevel optimization,'' \emph{{IEEE}
  Trans. Evol. Comput.}, vol.~26, no.~1, pp. 115--129, 2022.

\bibitem{FanLT20}
X.~Fan, K.~Li, and K.~C. Tan, ``Surrogate assisted evolutionary algorithm based
  on transfer learning for dynamic expensive multi-objective optimisation
  problems,'' in \emph{CEC'20: Proc. of 2020 {IEEE} Congress on Evolutionary
  Computation}.\hskip 1em plus 0.5em minus 0.4em\relax {IEEE}, 2020, pp. 1--8.

\bibitem{LiX0WT20}
K.~Li, Z.~Xiang, T.~Chen, S.~Wang, and K.~C. Tan, ``Understanding the automated
  parameter optimization on transfer learning for cross-project defect
  prediction: an empirical study,'' in \emph{{ICSE} '20: 42nd International
  Conference on Software Engineering, Seoul, South Korea, 27 June - 19 July,
  2020}.\hskip 1em plus 0.5em minus 0.4em\relax {ACM}, 2020, pp. 566--577.

\bibitem{FranceschiFSGP18}
L.~Franceschi, P.~Frasconi, S.~Salzo, R.~Grazzi, and M.~Pontil, ``Bilevel
  programming for hyperparameter optimization and meta-learning,'' in
  \emph{Proceedings of the 35th International Conference on Machine Learning,
  {ICML} 2018, Stockholmsm{\"{a}}ssan, Stockholm, Sweden, July 10-15, 2018},
  ser. Proceedings of Machine Learning Research, J.~G. Dy and A.~Krause, Eds.,
  vol.~80.\hskip 1em plus 0.5em minus 0.4em\relax {PMLR}, 2018, pp. 1563--1572.

\bibitem{LiXCT20}
K.~Li, Z.~Xiang, T.~Chen, and K.~C. Tan, ``Bilo-cpdp: Bi-level programming for
  automated model discovery in cross-project defect prediction,'' in \emph{35th
  {IEEE/ACM} International Conference on Automated Software Engineering, {ASE}
  2020, Melbourne, Australia, September 21-25, 2020}.\hskip 1em plus 0.5em
  minus 0.4em\relax {IEEE}, 2020, pp. 573--584.

\bibitem{LiSY18}
Y.~Li, J.~Su, and X.~Yang, ``Multi-objective vs. single-objective approaches
  for software defect prediction,'' in \emph{Proceedings of the 2018 2nd
  International Conference on Management Engineering, Software Engineering and
  Service Sciences}, 2018, pp. 122--127.

\bibitem{NiCWSG19}
C.~Ni, X.~Chen, F.~Wu, Y.~Shen, and Q.~Gu, ``An empirical study on pareto based
  multi-objective feature selection for software defect prediction,'' \emph{J.
  Syst. Softw.}, vol. 152, pp. 215--238, 2019.

\bibitem{CanforaLPOPP15}
G.~Canfora, A.~D. Lucia, M.~D. Penta, R.~Oliveto, A.~Panichella, and
  S.~Panichella, ``Defect prediction as a multiobjective optimization
  problem,'' \emph{Softw. Test. Verification Reliab.}, vol.~25, no.~4, pp.
  426--459, 2015.

\bibitem{CanforaLPOPP13}
------, ``Multi-objective cross-project defect prediction,'' in \emph{Sixth
  {IEEE} International Conference on Software Testing, Verification and
  Validation, {ICST} 2013, Luxembourg, Luxembourg, March 18-22, 2013}.\hskip
  1em plus 0.5em minus 0.4em\relax {IEEE} Computer Society, 2013, pp. 252--261.

\bibitem{RyuB16}
D.~Ryu and J.~Baik, ``Effective multi-objective na{\"{\i}}ve bayes learning for
  cross-project defect prediction,'' \emph{Appl. Soft Comput.}, vol.~49, pp.
  1062--1077, 2016.

\bibitem{KanwarAS23}
S.~Kanwar, L.~K. Awasthi, and V.~Shrivastava, ``Efficient random forest
  algorithm for multi-objective optimization in software defect prediction,''
  \emph{IETE Journal of Research}, pp. 1--13, 2023.

\bibitem{ChenDLTC24}
J.~Chen, J.~Ding, K.~Li, K.~C. Tan, and T.~Chai, ``A knee point driven
  evolutionary algorithm for multiobjective bilevel optimization,''
  \emph{{IEEE} Trans. Cybern.}, vol.~54, no.~7, pp. 4177--4189, 2024.

\bibitem{HosseiniTM18}
S.~Hosseini, B.~Turhan, and M.~M{\"{a}}ntyl{\"{a}}, ``A benchmark study on the
  effectiveness of search-based data selection and feature selection for cross
  project defect prediction,'' \emph{Inf. Softw. Technol.}, vol.~95, pp.
  296--312, 2018.

\bibitem{WZGJ19}
W.~Wen, B.~Zhang, X.~Gu, and X.~Ju, ``An empirical study on combining source
  selection and transfer learning for cross-project defect prediction,'' in
  \emph{2019 IEEE 1st International Workshop on Intelligent Bug Fixing
  (IBF)}.\hskip 1em plus 0.5em minus 0.4em\relax IEEE, 2019, pp. 29--38.

\bibitem{NamPK13}
J.~Nam, S.~J. Pan, and S.~Kim, ``Transfer defect learning,'' in \emph{35th
  International Conference on Software Engineering, {ICSE} '13, San Francisco,
  CA, USA, May 18-26, 2013}.\hskip 1em plus 0.5em minus 0.4em\relax {IEEE}
  Computer Society, 2013, pp. 382--391.

\bibitem{LeiXH20}
T.~Lei, J.~Xue, and W.~Han, ``Cross-project software defect prediction based on
  feature selection and transfer learning,'' in \emph{Machine Learning for
  Cyber Security - Third International Conference, {ML4CS} 2020, Guangzhou,
  China, October 8-10, 2020, Proceedings, Part {III}}, ser. Lecture Notes in
  Computer Science, vol. 12488.\hskip 1em plus 0.5em minus 0.4em\relax
  Springer, 2020, pp. 363--371.

\bibitem{LeiXWNSZ22}
T.~Lei, J.~Xue, Y.~Wang, Z.~Niu, Z.~Shi, and Y.~Zhang, ``Wcm-wtra: A
  cross-project defect prediction method based on feature selection and
  distance-weight transfer learning,'' \emph{Chinese Journal of Electronics},
  vol.~31, no.~2, pp. 354--366, 2022.

\bibitem{LiZJXGR23}
Z.~Li, H.~Zhang, X.~Jing, J.~Xie, M.~Guo, and J.~Ren, ``{DSSDPP:} data
  selection and sampling based domain programming predictor for cross-project
  defect prediction,'' \emph{{IEEE} Trans. Software Eng.}, vol.~49, no.~4, pp.
  1941--1963, 2023.

\bibitem{HeZRS16}
K.~He, X.~Zhang, S.~Ren, and J.~Sun, ``Deep residual learning for image
  recognition,'' in \emph{2016 {IEEE} Conference on Computer Vision and Pattern
  Recognition, {CVPR} 2016, Las Vegas, NV, USA, June 27-30, 2016}.\hskip 1em
  plus 0.5em minus 0.4em\relax {IEEE} Computer Society, 2016, pp. 770--778.

\bibitem{BergstraBBK11}
J.~Bergstra, R.~Bardenet, Y.~Bengio, and B.~K{\'{e}}gl, ``Algorithms for
  hyper-parameter optimization,'' in \emph{Advances in Neural Information
  Processing Systems 24: 25th Annual Conference on Neural Information
  Processing Systems 2011. Proceedings of a meeting held 12-14 December 2011,
  Granada, Spain}, 2011, pp. 2546--2554.

\bibitem{Stackelberg52}
H.~von Stackelberg, \emph{The Theory of the Market Economy}.\hskip 1em plus
  0.5em minus 0.4em\relax New York, NY, USA: Oxford University Press, 1952.

\bibitem{ChenLLT24}
L.~Chen, H.~Liu, K.~Li, and K.~C. Tan, ``Evolutionary bilevel optimization via
  multiobjective transformation-based lower-level search,'' \emph{{IEEE} Trans.
  Evol. Comput.}, vol.~28, no.~3, pp. 733--747, 2024.

\bibitem{AvraamidouP19a}
S.~Avraamidou and E.~N. Pistikopoulos, ``A multi-parametric optimization
  approach for bilevel mixed-integer linear and quadratic programming
  problems,'' \emph{Comput. Chem. Eng.}, vol. 125, pp. 98--113, 2019.

\bibitem{YangSXZ19}
H.~Yang, M.~Shi, Y.~Xia, and P.~Zhang, ``Security research on wireless
  networked control systems subject to jamming attacks,'' \emph{{IEEE} Trans.
  Cybern.}, vol.~49, no.~6, pp. 2022--2031, 2019.

\bibitem{GuanSQLGJ23}
Y.~Guan, Q.~Song, W.~Qi, K.~Li, L.~Guo, and A.~Jamalipour, ``Multidimensional
  resource fragmentation-aware virtual network embedding in {MEC} systems
  interconnected by metro optical networks,'' \emph{IEEE Internet of Things
  Journal}, vol.~10, no.~24, 2023.

\bibitem{QianHWLGGZZW22}
Y.~Qian, S.~Huang, B.~Wang, X.~Ling, X.~Guan, Z.~Gu, S.~Zeng, W.~Zhou, and
  H.~Wang, ``Robust network architecture search via feature distortion
  restraining,'' in \emph{Computer Vision - {ECCV} 2022 - 17th European
  Conference, Tel Aviv, Israel, October 23-27, 2022, Proceedings, Part {V}},
  ser. Lecture Notes in Computer Science, vol. 13665.\hskip 1em plus 0.5em
  minus 0.4em\relax Springer, 2022, pp. 122--138.

\bibitem{CarnereroCanoMSC20}
J.~Carnerero{-}Cano, L.~Mu{\~{n}}oz{-}Gonz{\'{a}}lez, P.~Spencer, and E.~C.
  Lupu, ``Regularisation can mitigate poisoning attacks: {A} novel analysis
  based on multiobjective bilevel optimisation,'' \emph{CoRR}, vol.
  abs/2003.00040, 2020.

\bibitem{YangYHLW23}
D.~Yang, S.~Yi, Q.~He, D.~Liu, and Y.~Wang, ``Railway alignment optimization
  based on multiobjective bi-level programming considering ecological impact,''
  \emph{{IEEE} Trans. Intell. Transp. Syst.}, vol.~24, no.~2, pp. 1712--1726,
  2023.

\bibitem{LiNGY22}
K.~Li, H.~Nie, H.~Gao, and X.~Yao, ``Posterior decision making based on
  decomposition-driven knee point identification,'' \emph{{IEEE} Trans. Evol.
  Comput.}, vol.~26, no.~6, pp. 1409--1423, 2022.

\bibitem{LiZLZL09}
K.~Li, J.~Zheng, M.~Li, C.~Zhou, and H.~Lv, ``A novel algorithm for
  non-dominated hypervolume-based multiobjective optimization,'' in
  \emph{SMC'09: Proc. of the {IEEE} International Conference on Systems, Man
  and Cybernetics}.\hskip 1em plus 0.5em minus 0.4em\relax {IEEE}, 2009, pp.
  5220--5226.

\bibitem{LiKM11}
K.~Li, S.~Kwong, and K.~Man, ``{JGBL} paradigm: a novel strategy to enhance the
  exploration ability of nsga-ii,'' in \emph{GECCO'11: Proc. of 13th Annual
  Genetic and Evolutionary Computation Conference}.\hskip 1em plus 0.5em minus
  0.4em\relax {ACM}, 2011, pp. 99--100.

\bibitem{LiDY18}
K.~Li, K.~Deb, and X.~Yao, ``R-metric: Evaluating the performance of
  preference-based evolutionary multiobjective optimization using reference
  points,'' \emph{{IEEE} Trans. Evol. Comput.}, vol.~22, no.~6, pp. 821--835,
  2018.

\bibitem{GaoNL19}
H.~Gao, H.~Nie, and K.~Li, ``Visualisation of pareto front approximation: {A}
  short survey and empirical comparisons,'' in \emph{CEC'19: Proc. of 2019
  {IEEE} Congress on Evolutionary Computation}.\hskip 1em plus 0.5em minus
  0.4em\relax {IEEE}, 2019, pp. 1750--1757.

\bibitem{LiLL22}
S.~Li, K.~Li, and W.~Li, ``Do we really need to use constraint violation in
  constrained evolutionary multi-objective optimization?'' in \emph{PPSN'22:
  Proc. of the 17th International Conference on Parallel Problem Solving from
  Nature}, ser. Lecture Notes in Computer Science, vol. 13399.\hskip 1em plus
  0.5em minus 0.4em\relax Springer, 2022, pp. 124--137.

\bibitem{ShanL21}
X.~Shan and K.~Li, ``An improved two-archive evolutionary algorithm for
  constrained multi-objective optimization,'' in \emph{EMO'21: Proc. of 11th
  International Conference on Evolutionary Multi-Criterion Optimization}, ser.
  Lecture Notes in Computer Science, vol. 12654.\hskip 1em plus 0.5em minus
  0.4em\relax Springer, 2021, pp. 235--247.

\bibitem{LiCFY19}
K.~Li, R.~Chen, G.~Fu, and X.~Yao, ``Two-archive evolutionary algorithm for
  constrained multiobjective optimization,'' \emph{{IEEE} Trans. Evol.
  Comput.}, vol.~23, no.~2, pp. 303--315, 2019.

\bibitem{LiLLY24}
S.~Li, K.~Li, W.~Li, and M.~Yang, ``Evolutionary alternating direction method
  of multipliers for constrained multi-objective optimization with unknown
  constraints,'' \emph{{IEEE} Trans. Evol. Comput.}, 2024, accepted for
  publication.

\bibitem{CaoKWL12}
J.~Cao, S.~Kwong, R.~Wang, and K.~Li, ``A weighted voting method using minimum
  square error based on extreme learning machine,'' in \emph{ICMLC'12: Proc. of
  2012 International Conference on Machine Learning and Cybernetics}.\hskip 1em
  plus 0.5em minus 0.4em\relax {IEEE}, 2012, pp. 411--414.

\bibitem{LiKD15}
K.~Li, S.~Kwong, and K.~Deb, ``A dual-population paradigm for evolutionary
  multiobjective optimization,'' \emph{Inf. Sci.}, vol. 309, pp. 50--72, 2015.

\bibitem{LiKWTM13}
K.~Li, S.~Kwong, R.~Wang, K.~Tang, and K.~Man, ``Learning paradigm based on
  jumping genes: {A} general framework for enhancing exploration in
  evolutionary multiobjective optimization,'' \emph{Inf. Sci.}, vol. 226, pp.
  1--22, 2013.

\bibitem{LiK14}
K.~Li and S.~Kwong, ``A general framework for evolutionary multiobjective
  optimization via manifold learning,'' \emph{Neurocomputing}, vol. 146, pp.
  65--74, 2014.

\bibitem{RuanLDL20}
X.~Ruan, K.~Li, B.~Derbel, and A.~Liefooghe, ``Surrogate assisted evolutionary
  algorithm for medium scale multi-objective optimisation problems,'' in
  \emph{{GECCO}'20: Proc. of 2020 Genetic and Evolutionary Computation
  Conference}.\hskip 1em plus 0.5em minus 0.4em\relax {ACM}, 2020, pp.
  560--568.

\bibitem{LiLY23}
K.~Li, G.~Lai, and X.~Yao, ``Interactive evolutionary multiobjective
  optimization via learning to rank,'' \emph{{IEEE} Trans. Evol. Comput.},
  vol.~27, no.~4, pp. 749--763, 2023.

\bibitem{LiCSY19}
K.~Li, R.~Chen, D.~A. Savic, and X.~Yao, ``Interactive decomposition
  multiobjective optimization via progressively learned value functions,''
  \emph{{IEEE} Trans. Fuzzy Syst.}, vol.~27, no.~5, pp. 849--860, 2019.

\bibitem{LiZKLW14}
K.~Li, Q.~Zhang, S.~Kwong, M.~Li, and R.~Wang, ``Stable matching-based
  selection in evolutionary multiobjective optimization,'' \emph{{IEEE} Trans.
  Evol. Comput.}, vol.~18, no.~6, pp. 909--923, 2014.

\bibitem{LiLDMY20}
K.~Li, M.~Liao, K.~Deb, G.~Min, and X.~Yao, ``Does preference always help? {A}
  holistic study on preference-based evolutionary multiobjective optimization
  using reference points,'' \emph{{IEEE} Trans. Evol. Comput.}, vol.~24, no.~6,
  pp. 1078--1096, 2020.

\bibitem{LiDZZ17}
K.~Li, K.~Deb, Q.~Zhang, and Q.~Zhang, ``Efficient nondomination level update
  method for steady-state evolutionary multiobjective optimization,''
  \emph{{IEEE} Trans. Cybern.}, vol.~47, no.~9, pp. 2838--2849, 2017.

\bibitem{Li19}
K.~Li, ``Progressive preference learning: Proof-of-principle results in
  {MOEA/D},'' in \emph{EMO'19: Proc. of 10th International Conference on
  Evolutionary Multi-Criterion Optimization}, ser. Lecture Notes in Computer
  Science, vol. 11411.\hskip 1em plus 0.5em minus 0.4em\relax Springer, 2019,
  pp. 631--643.

\bibitem{LiKCLZS12}
K.~Li, S.~Kwong, J.~Cao, M.~Li, J.~Zheng, and R.~Shen, ``Achieving balance
  between proximity and diversity in multi-objective evolutionary algorithm,''
  \emph{Inf. Sci.}, vol. 182, no.~1, pp. 220--242, 2012.

\bibitem{WuKZLWL15}
M.~Wu, S.~Kwong, Q.~Zhang, K.~Li, R.~Wang, and B.~Liu, ``Two-level stable
  matching-based selection in {MOEA/D},'' in \emph{SMC'15: Proc. of 2015 {IEEE}
  International Conference on Systems, Man, and Cybernetics}.\hskip 1em plus
  0.5em minus 0.4em\relax {IEEE}, 2015, pp. 1720--1725.

\bibitem{LiKZD15}
K.~Li, S.~Kwong, Q.~Zhang, and K.~Deb, ``Interrelationship-based selection for
  decomposition multiobjective optimization,'' \emph{{IEEE} Trans. Cybern.},
  vol.~45, no.~10, pp. 2076--2088, 2015.

\bibitem{WuLKZZ17}
M.~Wu, K.~Li, S.~Kwong, Y.~Zhou, and Q.~Zhang, ``Matching-based selection with
  incomplete lists for decomposition multiobjective optimization,''
  \emph{{IEEE} Trans. Evol. Comput.}, vol.~21, no.~4, pp. 554--568, 2017.

\bibitem{Zhou0M22}
S.~Zhou, K.~Li, and G.~Min, ``Adversarial example generation via genetic
  algorithm: a preliminary result,'' in \emph{{GECCO}'22: Genetic and
  Evolutionary Computation Conference}.\hskip 1em plus 0.5em minus 0.4em\relax
  {ACM}, 2022, pp. 469--470.

\bibitem{Williams0M22}
P.~N. Williams, K.~Li, and G.~Min, ``Black-box adversarial attack via
  overlapped shapes,'' in \emph{{GECCO}'22: Genetic and Evolutionary
  Computation Conference}.\hskip 1em plus 0.5em minus 0.4em\relax {ACM}, 2022,
  pp. 467--468.

\bibitem{WilliamsLM23b}
------, ``Evolutionary art attack for black-box adversarial example
  generation,'' \emph{{IEEE} Trans. Evol. Comput.}, 2024, accepted for
  publication.

\bibitem{WilliamsL23b}
P.~N. Williams and K.~Li, ``Camopatch: An evolutionary strategy for generating
  camoflauged adversarial patches,'' in \emph{NeurIPS'23: Proc. of 37th
  Conference on Neural Information Processing Systems}, 2023.

\bibitem{Williams023}
------, ``Black-box sparse adversarial attack via multi-objective optimisation
  {CVPR} proceedings,'' in \emph{CVPR'23: Proc. of 2023 {IEEE/CVF} Conference
  on Computer Vision and Pattern Recognition}.\hskip 1em plus 0.5em minus
  0.4em\relax {IEEE}, 2023, pp. 12\,291--12\,301.

\bibitem{WilliamsLM23a}
P.~N. Williams, K.~Li, and G.~Min, ``Sparse adversarial attack via bi-objective
  optimization,'' in \emph{EMO'23: Proc. of 12th International Conference on
  Evolutionary Multi-Criterion Optimization}, ser. Lecture Notes in Computer
  Science, vol. 13970.\hskip 1em plus 0.5em minus 0.4em\relax Springer, 2023,
  pp. 118--133.

\bibitem{WilliamsLM23}
------, ``A surrogate assisted evolutionary strategy for image approximation by
  density-ratio estimation,'' in \emph{CEC'23: Proc. of 2023 {IEEE} Congress on
  Evolutionary Computation}.\hskip 1em plus 0.5em minus 0.4em\relax {IEEE},
  2023, pp. 1--8.

\bibitem{LiFK11}
K.~Li, {\'{A}}.~Fialho, and S.~Kwong, ``Multi-objective differential evolution
  with adaptive control of parameters and operators,'' in \emph{LION5: Proc. of
  the 5th International Conference on Learning and Intelligent Optimization},
  2011, pp. 473--487.

\bibitem{LiKWCR12}
K.~Li, S.~Kwong, R.~Wang, J.~Cao, and I.~J. Rudas, ``Multi-objective
  differential evolution with self-navigation,'' in \emph{SMC'12: Proc. of 2012
  {IEEE} International Conference on Systems, Man, and Cybernetics}.\hskip 1em
  plus 0.5em minus 0.4em\relax {IEEE}, 2012, pp. 508--513.

\bibitem{LiWKC13}
K.~Li, R.~Wang, S.~Kwong, and J.~Cao, ``Evolving extreme learning machine
  paradigm with adaptive operator selection and parameter control,''
  \emph{International Journal of Uncertainty, Fuzziness and Knowledge-Based
  Systems}, vol.~21, no. supp02, pp. 143--154, 2013.

\bibitem{LiFKZ14}
K.~Li, {\'{A}}.~Fialho, S.~Kwong, and Q.~Zhang, ``Adaptive operator selection
  with bandits for a multiobjective evolutionary algorithm based on
  decomposition,'' \emph{{IEEE} Trans. Evol. Comput.}, vol.~18, no.~1, pp.
  114--130, 2014.

\bibitem{SunL20}
L.~Sun and K.~Li, ``Adaptive operator selection based on dynamic thompson
  sampling for {MOEA/D},'' in \emph{PPNS'20: Proc. of 16th International
  Conference on Parallel Problem Solving from Nature}, ser. Lecture Notes in
  Computer Science, vol. 12270.\hskip 1em plus 0.5em minus 0.4em\relax
  Springer, 2020, pp. 271--284.

\bibitem{CaoKWLLK15}
J.~Cao, S.~Kwong, R.~Wang, X.~Li, K.~Li, and X.~Kong, ``Class-specific soft
  voting based multiple extreme learning machines ensemble,''
  \emph{Neurocomputing}, vol. 149, pp. 275--284, 2015.

\bibitem{LiXT19}
K.~Li, Z.~Xiang, and K.~C. Tan, ``Which surrogate works for empirical
  performance modelling? {A} case study with differential evolution,'' in
  \emph{CEC'19: Proc. of 2019 {IEEE} Congress on Evolutionary
  Computation}.\hskip 1em plus 0.5em minus 0.4em\relax {IEEE}, 2019, pp.
  1988--1995.

\bibitem{WuLKZZ19}
M.~Wu, K.~Li, S.~Kwong, Q.~Zhang, and J.~Zhang, ``Learning to decompose: {A}
  paradigm for decomposition-based multiobjective optimization,'' \emph{{IEEE}
  Trans. Evol. Comput.}, vol.~23, no.~3, pp. 376--390, 2019.

\bibitem{Liu0020}
M.~Liu, K.~Li, and T.~Chen, ``Deepsqli: deep semantic learning for testing
  {SQL} injection,'' in \emph{{ISSTA}'20: 29th {ACM} {SIGSOFT} International
  Symposium on Software Testing and Analysis}.\hskip 1em plus 0.5em minus
  0.4em\relax {ACM}, 2020, pp. 286--297.

\bibitem{ZhouHSL24}
S.~Zhou, M.~Huang, Y.~Sun, and K.~Li, ``Evolutionary multi-objective
  optimization for contextual adversarial example generation,'' \emph{Proc.
  {ACM} Softw. Eng.}, vol.~1, no. {FSE}, pp. 2285--2308, 2024.

\bibitem{LyuLHWYL23}
B.~Lyu, L.~Lu, M.~Hamdi, S.~Wen, Y.~Yang, and K.~Li, ``{MTLP-JR:} multi-task
  learning-based prediction for joint ranking in neural architecture search,''
  \emph{Comput. Electr. Eng.}, vol. 105, p. 108474, 2023.

\bibitem{HuangL23}
M.~Huang and K.~Li, ``Exploring structural similarity in fitness landscapes via
  graph data mining: {A} case study on number partitioning problems,'' in
  \emph{IJCAI'23: Proc. of the Thirty-Second International Joint Conference on
  Artificial Intelligence}.\hskip 1em plus 0.5em minus 0.4em\relax ijcai.org,
  2023, pp. 5595--5603.

\bibitem{LyuYWHL23}
B.~Lyu, Y.~Yang, S.~Wen, T.~Huang, and K.~Li, ``Neural architecture search for
  portrait parsing,'' \emph{{IEEE} Trans. Neural Networks Learn. Syst.},
  vol.~34, no.~3, pp. 1112--1121, 2023.

\bibitem{LyuHYCYLWH23}
B.~Lyu, M.~Hamdi, Y.~Yang, Y.~Cao, Z.~Yan, K.~Li, S.~Wen, and T.~Huang,
  ``Efficient spectral graph convolutional network deployment on memristive
  crossbars,'' \emph{{IEEE} Trans. Emerg. Top. Comput. Intell.}, vol.~7, no.~2,
  pp. 415--425, 2023.

\bibitem{XuLA21}
J.~Xu, K.~Li, and M.~Abusara, ``Multi-objective reinforcement learning based
  multi-microgrid system optimisation problem,'' in \emph{EMO'21: Proc. of 11th
  International Conference on Evolutionary Multi-Criterion Optimization}, ser.
  Lecture Notes in Computer Science, vol. 12654.\hskip 1em plus 0.5em minus
  0.4em\relax Springer, 2021, pp. 684--696.

\bibitem{XuLA021}
J.~Xu, K.~Li, M.~Abusara, and Y.~Zhang, ``Admm-based {OPF} problem against
  cyber attacks in smart grid,'' in \emph{SMC'21: Proc. of 2021 {IEEE}
  International Conference on Systems, Man, and Cybernetics}.\hskip 1em plus
  0.5em minus 0.4em\relax {IEEE}, 2021, pp. 1418--1423.

\bibitem{XuLL24a}
J.~Xu, K.~Li, and D.~Li, ``Multioutput framework for time-series forecasting in
  smart grid meets data scarcity,'' \emph{{IEEE} Trans. Ind. Informatics},
  vol.~20, no.~9, pp. 11\,202--11\,212, 2024.

\bibitem{XuLA22}
J.~Xu, K.~Li, and M.~Abusara, ``Preference based multi-objective reinforcement
  learning for multi-microgrid system optimization problem in smart grid,''
  \emph{Memetic Comput.}, vol.~14, no.~2, pp. 225--235, 2022.

\bibitem{BillingsleyLMMG20}
J.~Billingsley, K.~Li, W.~Miao, G.~Min, and N.~Georgalas, ``Routing-led
  placement of vnfs in arbitrary networks,'' in \emph{CEC'20: Proc. of 2020
  {IEEE} Congress on Evolutionary Computation}.\hskip 1em plus 0.5em minus
  0.4em\relax {IEEE}, 2020, pp. 1--8.

\bibitem{BillingsleyMLMG20}
J.~Billingsley, W.~Miao, K.~Li, G.~Min, and N.~Georgalas, ``Performance
  analysis of {SDN} and {NFV} enabled mobile cloud computing,'' in
  \emph{GLOBECOM'20: Proc. of 2020 {IEEE} Global Communications
  Conference}.\hskip 1em plus 0.5em minus 0.4em\relax {IEEE}, 2020, pp. 1--6.

\bibitem{GuanSQGLJ23}
Y.~Guan, Q.~Song, W.~Qi, L.~Guo, K.~Li, and A.~Jamalipour, ``Multidimensional
  resource fragmentation-aware virtual network embedding for iot applications
  in {MEC} networks,'' \emph{{IEEE} Internet Things J.}, vol.~10, no.~24, pp.
  22\,223--22\,232, 2023.

\bibitem{BillingsleyLMMG21}
J.~Billingsley, K.~Li, W.~Miao, G.~Min, and N.~Georgalas, ``Parallel algorithms
  for the multiobjective virtual network function placement problem,'' in
  \emph{EMO'21: Proc. of the 11th International Conference on Evolutionary
  Multi-Criterion Optimization}, ser. Lecture Notes in Computer Science, vol.
  12654.\hskip 1em plus 0.5em minus 0.4em\relax Springer, 2021, pp. 708--720.

\bibitem{YangL23}
H.~Yang and K.~Li, ``Instoptima: Evolutionary multi-objective instruction
  optimization via large language model-based instruction operators,'' in
  \emph{EMNLP'23: Findings of the Association for Computational
  Linguistics}.\hskip 1em plus 0.5em minus 0.4em\relax Association for
  Computational Linguistics, 2023, pp. 13\,593--13\,602.

\bibitem{HKV2019}
F.~Hutter, L.~Kotthoff, and J.~Vanschoren, Eds., \emph{Automated Machine
  Learning - Methods, Systems, Challenges}, ser. The Springer Series on
  Challenges in Machine Learning.\hskip 1em plus 0.5em minus 0.4em\relax
  Springer, 2019.

\bibitem{ZhouYLCLZQX18}
Y.~Zhou, Y.~Yang, H.~Lu, L.~Chen, Y.~Li, Y.~Zhao, J.~Qian, and B.~Xu, ``How far
  we have progressed in the journey? an examination of cross-project defect
  prediction,'' \emph{{ACM} Trans. Softw. Eng. Methodol.}, vol.~27, no.~1, pp.
  1:1--1:51, 2018.

\bibitem{PedregosaVGMTGBPWDVPCBPD11}
F.~Pedregosa, G.~Varoquaux, A.~Gramfort, V.~Michel, B.~Thirion, O.~Grisel,
  M.~Blondel, P.~Prettenhofer, R.~Weiss, V.~Dubourg, J.~VanderPlas, A.~Passos,
  D.~Cournapeau, M.~Brucher, M.~Perrot, and E.~Duchesnay, ``Scikit-learn:
  Machine learning in python,'' \emph{J. Mach. Learn. Res.}, vol.~12, pp.
  2825--2830, 2011.

\bibitem{LiCWMTTL17}
J.~Li, K.~Cheng, S.~Wang, F.~Morstatter, R.~P. Trevino, J.~Tang, and H.~Liu,
  ``Feature selection: {A} data perspective,'' \emph{{ACM} Comput. Surv.},
  vol.~50, no.~6, pp. 94:1--94:45, 2018.

\bibitem{NiLCGCH17}
C.~Ni, W.~Liu, X.~Chen, Q.~Gu, D.~Chen, and G.~Q. Huang, ``A cluster based
  feature selection method for cross-project software defect prediction,''
  \emph{J. Comput. Sci. Technol.}, vol.~32, no.~6, pp. 1090--1107, 2017.

\bibitem{MuthukrishnanR16}
R.~Muthukrishnan and R.~Rohini, ``Lasso: A feature selection technique in
  predictive modeling for machine learning,'' in \emph{2016 IEEE international
  conference on advances in computer applications (ICACA)}.\hskip 1em plus
  0.5em minus 0.4em\relax IEEE, 2016, pp. 18--20.

\bibitem{GenuerPT10}
R.~Genuer, J.~Poggi, and C.~Tuleau{-}Malot, ``Variable selection using random
  forests,'' \emph{Pattern Recognit. Lett.}, vol.~31, no.~14, pp. 2225--2236,
  2010.

\bibitem{NagappanBZ06}
N.~Nagappan, T.~Ball, and A.~Zeller, ``Mining metrics to predict component
  failures,'' in \emph{28th International Conference on Software Engineering
  {(ICSE} 2006), Shanghai, China, May 20-28, 2006}, L.~J. Osterweil, H.~D.
  Rombach, and M.~L. Soffa, Eds.\hskip 1em plus 0.5em minus 0.4em\relax {ACM},
  2006, pp. 452--461.

\bibitem{HePMY13}
Z.~He, F.~Peters, T.~Menzies, and Y.~Yang, ``Learning from open-source
  projects: An empirical study on defect prediction,'' in \emph{2013 {ACM} /
  {IEEE} International Symposium on Empirical Software Engineering and
  Measurement, Baltimore, Maryland, USA, October 10-11, 2013}.\hskip 1em plus
  0.5em minus 0.4em\relax {IEEE} Computer Society, 2013, pp. 45--54.

\bibitem{Herbold13}
S.~Herbold, ``Training data selection for cross-project defect prediction,'' in
  \emph{9th International Conference on Predictive Models in Software
  Engineering, {PROMISE} '13, Baltimore, MD, USA, October 9, 2013}.\hskip 1em
  plus 0.5em minus 0.4em\relax {ACM}, 2013, pp. 6:1--6:10.

\bibitem{RyuCB16}
D.~Ryu, O.~Choi, and J.~Baik, ``Value-cognitive boosting with a support vector
  machine for cross-project defect prediction,'' \emph{Empir. Softw. Eng.},
  vol.~21, no.~1, pp. 43--71, 2016.

\bibitem{TurhanMBS09}
B.~Turhan, T.~Menzies, A.~B. Bener, and J.~S.~D. Stefano, ``On the relative
  value of cross-company and within-company data for defect prediction,''
  \emph{Empir. Softw. Eng.}, vol.~14, no.~5, pp. 540--578, 2009.

\bibitem{LimsetthoBKHM18}
N.~Limsettho, K.~E. Bennin, J.~W. Keung, H.~Hata, and K.~Matsumoto, ``Cross
  project defect prediction using class distribution estimation and
  oversampling,'' \emph{Inf. Softw. Technol.}, vol. 100, pp. 87--102, 2018.

\bibitem{0001MKZ14}
F.~Zhang, A.~Mockus, I.~Keivanloo, and Y.~Zou, ``Towards building a universal
  defect prediction model,'' in \emph{11th Working Conference on Mining
  Software Repositories, {MSR} 2014, Proceedings, May 31 - June 1, 2014,
  Hyderabad, India}, P.~T. Devanbu, S.~Kim, and M.~Pinzger, Eds.\hskip 1em plus
  0.5em minus 0.4em\relax {ACM}, 2014, pp. 182--191.

\bibitem{RyuJB15}
D.~Ryu, J.~Jang, and J.~Baik, ``A hybrid instance selection using
  nearest-neighbor for cross-project defect prediction,'' \emph{J. Comput. Sci.
  Technol.}, vol.~30, no.~5, pp. 969--980, 2015.

\bibitem{PetersMGZ13}
F.~Peters, T.~Menzies, L.~Gong, and H.~Zhang, ``Balancing privacy and utility
  in cross-company defect prediction,'' \emph{{IEEE} Trans. Software Eng.},
  vol.~39, no.~8, pp. 1054--1068, 2013.

\bibitem{Tantithamthavorn19}
C.~Tantithamthavorn, S.~McIntosh, A.~E. Hassan, and K.~Matsumoto, ``The impact
  of automated parameter optimization on defect prediction models,''
  \emph{{IEEE} Trans. Software Eng.}, vol.~45, no.~7, pp. 683--711, 2019.

\bibitem{Saeed2023}
M.~S. Saeed, ``Role of feature selection in cross project software defect
  prediction-a review,'' \emph{International Journal of Computations,
  Information and Manufacturing (IJCIM)}, vol.~3, no.~2, pp. 37--56, 2023.

\bibitem{TsoumakasPV09}
G.~Tsoumakas, I.~Partalas, and I.~P. Vlahavas, ``An ensemble pruning primer,''
  in \emph{Applications of Supervised and Unsupervised Ensemble Methods}, ser.
  Studies in Computational Intelligence.\hskip 1em plus 0.5em minus 0.4em\relax
  Springer, 2009, vol. 245, pp. 1--13.

\bibitem{ShenL0TZ022}
Y.~Shen, Y.~Lu, Y.~Li, Y.~Tu, W.~Zhang, and B.~Cui, ``Divbo: Diversity-aware
  {CASH} for ensemble learning,'' in \emph{Advances in Neural Information
  Processing Systems 35: Annual Conference on Neural Information Processing
  Systems 2022, NeurIPS 2022, New Orleans, LA, USA, November 28 - December 9,
  2022}, 2022.

\bibitem{Yule90}
G.~U. Yule, ``Vii. on the association of attributes in statistics: with
  illustrations from the material of the childhood society, \&c,''
  \emph{Philosophical Transactions of the Royal Society of London. Series A,
  Containing Papers of a Mathematical or Physical Character}, vol. 194, no.
  252-261, pp. 257--319, 1900.

\bibitem{Wolpert92}
D.~H. Wolpert, ``Stacked generalization,'' \emph{Neural Networks}, vol.~5,
  no.~2, pp. 241--259, 1992.

\bibitem{GloverL97}
F.~W. Glover and M.~Laguna, \emph{Tabu Search}.\hskip 1em plus 0.5em minus
  0.4em\relax Kluwer, 1997.

\bibitem{TanKLY03}
K.~C. Tan, E.~F. Khor, T.~H. Lee, and Y.~J. Yang, ``A tabu-based exploratory
  evolutionary algorithm for multiobjective optimization,'' \emph{Artif.
  Intell. Rev.}, vol.~19, no.~3, pp. 231--260, 2003.

\bibitem{DebA95}
K.~Deb and R.~B. Agrawal, ``Simulated binary crossover for continuous search
  space,'' \emph{Complex Syst.}, vol.~9, no.~2, 1995.

\bibitem{DebG1996}
K.~Deb and M.~Goyal, ``A combined genetic adaptive search (geneas) for
  engineering design,'' \emph{Computer Science and informatics}, vol.~26, pp.
  30--45, 1996.

\bibitem{DebAPM02}
K.~Deb, S.~Agrawal, A.~Pratap, and T.~Meyarivan, ``A fast and elitist
  multiobjective genetic algorithm: {NSGA-II},'' \emph{{IEEE} Trans. Evol.
  Comput.}, vol.~6, no.~2, pp. 182--197, 2002.

\bibitem{KhatriS22}
Y.~Khatri and S.~K. Singh, ``Cross project defect prediction: a comprehensive
  survey with its {SWOT} analysis,'' \emph{Innov. Syst. Softw. Eng.}, vol.~18,
  no.~2, pp. 263--281, 2022.

\bibitem{ShepperdSSM13}
M.~J. Shepperd, Q.~Song, Z.~Sun, and C.~Mair, ``Data quality: Some comments on
  the {NASA} software defect datasets,'' \emph{{IEEE} Trans. Software Eng.},
  vol.~39, no.~9, pp. 1208--1215, 2013.

\bibitem{DAmbrosLR10}
M.~D'Ambros, M.~Lanza, and R.~Robbes, ``An extensive comparison of bug
  prediction approaches,'' in \emph{Proceedings of the 7th International
  Working Conference on Mining Software Repositories, {MSR} 2010 (Co-located
  with ICSE), Cape Town, South Africa, May 2-3, 2010, Proceedings}, 2010, pp.
  31--41.

\bibitem{JureczkoM10}
M.~Jureczko and L.~Madeyski, ``Towards identifying software project clusters
  with regard to defect prediction,'' in \emph{Proceedings of the 6th
  International Conference on Predictive Models in Software Engineering,
  {PROMISE} 2010, Timisoara, Romania, September 12-13, 2010}.\hskip 1em plus
  0.5em minus 0.4em\relax {ACM}, 2010, p.~9.

\bibitem{WuZKC11}
R.~Wu, H.~Zhang, S.~Kim, and S.~Cheung, ``Relink: recovering links between bugs
  and changes,'' in \emph{SIGSOFT/FSE'11 19th {ACM} {SIGSOFT} Symposium on the
  Foundations of Software Engineering {(FSE-19)} and ESEC'11: 13th European
  Software Engineering Conference (ESEC-13), Szeged, Hungary, September 5-9,
  2011}.\hskip 1em plus 0.5em minus 0.4em\relax {ACM}, 2011, pp. 15--25.

\bibitem{CalefatoQLK23}
F.~Calefato, L.~Quaranta, F.~Lanubile, and M.~Kalinowski, ``Assessing the use
  of automl for data-driven software engineering,'' in \emph{{ACM/IEEE}
  International Symposium on Empirical Software Engineering and Measurement,
  {ESEM} 2023, New Orleans, LA, USA, October 26-27, 2023}.\hskip 1em plus 0.5em
  minus 0.4em\relax {IEEE}, 2023, pp. 1--12.

\bibitem{JinSH19}
H.~Jin, Q.~Song, and X.~Hu, ``Auto-keras: An efficient neural architecture
  search system,'' in \emph{Proceedings of the 25th {ACM} {SIGKDD}
  International Conference on Knowledge Discovery {\&} Data Mining, {KDD} 2019,
  Anchorage, AK, USA, August 4-8, 2019}.\hskip 1em plus 0.5em minus 0.4em\relax
  {ACM}, 2019, pp. 1946--1956.

\bibitem{mljar}
A.~P\l{}o\'{n}ska and P.~P\l{}o\'{n}ski, ``Mljar: State-of-the-art automated
  machine learning framework for tabular data. version 0.10.3,'' \L{}apy,
  Poland, 2021.

\bibitem{SahaUM0LHYKP22}
R.~K. Saha, A.~Ura, S.~Mahajan, C.~Zhu, L.~Li, Y.~Hu, H.~Yoshida, S.~Khurshid,
  and M.~R. Prasad, ``{SAPIENTML:} synthesizing machine learning pipelines by
  learning from human-written solutions,'' in \emph{44th {IEEE/ACM} 44th
  International Conference on Software Engineering, {ICSE} 2022, Pittsburgh,
  PA, USA, May 25-27, 2022}.\hskip 1em plus 0.5em minus 0.4em\relax {ACM},
  2022, pp. 1932--1944.

\bibitem{FeurerKESBH15}
M.~Feurer, A.~Klein, K.~Eggensperger, J.~T. Springenberg, M.~Blum, and
  F.~Hutter, ``Efficient and robust automated machine learning,'' in
  \emph{Advances in Neural Information Processing Systems 28: Annual Conference
  on Neural Information Processing Systems 2015, December 7-12, 2015, Montreal,
  Quebec, Canada}, 2015, pp. 2962--2970.

\bibitem{LiDZK15}
K.~Li, K.~Deb, Q.~Zhang, and S.~Kwong, ``An evolutionary many-objective
  optimization algorithm based on dominance and decomposition,'' \emph{{IEEE}
  Trans. Evol. Comput.}, vol.~19, no.~5, pp. 694--716, 2015.

\bibitem{BonabC19}
H.~R. Bonab and F.~Can, ``Less is more: {A} comprehensive framework for the
  number of components of ensemble classifiers,'' \emph{{IEEE} Trans. Neural
  Networks Learn. Syst.}, vol.~30, no.~9, pp. 2735--2745, 2019.

\bibitem{Metz78}
C.~E. Metz, ``Basic principles of roc analysis,'' in \emph{Seminars in nuclear
  medicine}, vol.~8, no.~4.\hskip 1em plus 0.5em minus 0.4em\relax Elsevier,
  1978, pp. 283--298.

\bibitem{HanK2000}
J.~Han and M.~Kamber, \emph{Data Mining: Concepts and Techniques}.\hskip 1em
  plus 0.5em minus 0.4em\relax Morgan Kaufmann, 2000.

\bibitem{BaldiBCAN00}
P.~Baldi, S.~Brunak, Y.~Chauvin, C.~A.~F. Andersen, and H.~Nielsen, ``Assessing
  the accuracy of prediction algorithms for classification: an overview,''
  \emph{Bioinform.}, vol.~16, no.~5, pp. 412--424, 2000.

\bibitem{Haynes2013}
W.~Haynes, \emph{Wilcoxon Rank Sum Test}.\hskip 1em plus 0.5em minus
  0.4em\relax New York, NY: Springer New York, 2013, pp. 2354--2355.

\bibitem{ArcuriB11}
A.~Arcuri and L.~C. Briand, ``A practical guide for using statistical tests to
  assess randomized algorithms in software engineering,'' in \emph{Proceedings
  of the 33rd International Conference on Software Engineering, {ICSE} 2011,
  Waikiki, Honolulu , HI, USA, May 21-28, 2011}.\hskip 1em plus 0.5em minus
  0.4em\relax {ACM}, 2011, pp. 1--10.

\bibitem{MittasA13}
N.~Mittas and L.~Angelis, ``Ranking and clustering software cost estimation
  models through a multiple comparisons algorithm,'' \emph{{IEEE} Trans.
  Software Eng.}, vol.~39, no.~4, pp. 537--551, 2013.

\bibitem{VarghaD00}
A.~Vargha and H.~D. Delaney, ``A critique and improvement of the cl common
  language effect size statistics of mcgraw and wong,'' \emph{J. Educ. Behav.
  Stat.}, vol.~25, no.~2, pp. 101--132, 2000.

\bibitem{Ozturk19}
M.~M. {\"O}zt{\"u}rk, ``Comparing hyperparameter optimization in cross-and
  within-project defect prediction: A case study,'' \emph{Arabian Journal for
  Science and Engineering}, vol.~44, pp. 3515--3530, 2019.

\bibitem{ThorntonHHL13}
C.~Thornton, F.~Hutter, H.~H. Hoos, and K.~Leyton{-}Brown, ``Auto-weka:
  combined selection and hyperparameter optimization of classification
  algorithms,'' in \emph{The 19th {ACM} {SIGKDD} International Conference on
  Knowledge Discovery and Data Mining, {KDD} 2013, Chicago, IL, USA, August
  11-14, 2013}.\hskip 1em plus 0.5em minus 0.4em\relax {ACM}, 2013, pp.
  847--855.

\bibitem{LeDellP20}
E.~LeDell and S.~Poirier, ``H2o automl: Scalable automatic machine learning,''
  in \emph{Proceedings of the AutoML Workshop at ICML}, vol. 2020.\hskip 1em
  plus 0.5em minus 0.4em\relax ICML San Diego, CA, USA, 2020.

\bibitem{TanakaMY19}
K.~Tanaka, A.~Monden, and Z.~Y{\"{u}}cel, ``Prediction of software defects
  using automated machine learning,'' in \emph{20th {IEEE/ACIS} International
  Conference on Software Engineering, Artificial Intelligence, Networking and
  Parallel/Distributed Computing, {SNPD} 2019, Toyama, Japan, July 8-11,
  2019}.\hskip 1em plus 0.5em minus 0.4em\relax {IEEE}, 2019, pp. 490--494.

\bibitem{CookCD79}
T.~D. Cook, D.~T. Campbell, and A.~Day, \emph{Quasi-experimentation: Design \&
  analysis issues for field settings}.\hskip 1em plus 0.5em minus 0.4em\relax
  Boston: Houghton Mifflin, 1979.

\end{thebibliography}
